\documentclass[manuscript]{acmart}

\usepackage{etoolbox}
\providetoggle{eaamo}
\settoggle{eaamo}{true}

\providetoggle{facct}
\settoggle{facct}{true}

\providetoggle{arxiv}
\settoggle{arxiv}{false}

\iftoggle{arxiv}{
    \usepackage{ext/jmlr2e_preprint}
    \usepackage{algorithm,algpseudocode}
    \ShortHeadings{Algorithmic Censoring in Dynamic Learning Systems}{Chien, Roberts, Ustun}
    \jmlrheading{1}{2023}{}{}{}{Jennifer Chien, Margaret Roberts and Berk Ustun}
    \firstpageno{1}
}

\usepackage[utf8]{inputenc} 
\usepackage[T1]{fontenc} 
\usepackage{amsfonts,bbm,bm,courier,microtype}

\usepackage{enumitem,todonotes,comment}

\iftoggle{arxiv}{
\setcitestyle{numbers,comma,square,sort}
}

\usepackage{array,booktabs,tabularx,multirow,colortbl,hhline}
\usepackage{makecell}

\usepackage{graphicx,xcolor,caption,placeins,wrapfig}

\usepackage{mathtools,nicefrac,eqnarray}

\usepackage{algorithm,algpseudocode}

\usepackage[font=small,labelfont=bf,width=\textwidth]{caption}    
\usepackage{hyperref}
\hypersetup{%
  colorlinks=true,
  urlcolor=blue,%
  linkcolor=blue,%
  citecolor=blue,%
  linkbordercolor=blue,
  pdfborderstyle={/S/U/W 1}
 }

\usepackage[capitalise]{cleveref}
\crefformat{section}{\S#2#1#3} 
\crefformat{subsection}{\S#2#1#3}
\crefformat{subsubsection}{\S#2#1#3}

\newcolumntype{H}{>{\setbox0=\hbox\bgroup}c<{\egroup}@{}}
\newcommand{\cell}[2]{\setlength{\tabcolsep}{0pt}\begin{tabular}{#1}#2 \end{tabular}}

\newtheorem{definition}{Definition}

\setitemize{leftmargin=1em,parsep=2pt,label=\raisebox{0.25ex}{\tiny$\bullet$}}
\setlist[enumerate]{leftmargin=*, label= {\arabic*.}, itemsep=0.5em}
\newlist{thmlist}{enumerate}{1}
\setlist[thmlist]{leftmargin=*,label=\raisebox{0.25ex}{\tiny$\bullet$}, topsep=0.2em,itemsep=2pt}

\newcommand{\textheader}[1]{{\textbf{#1}}}
\newcommand{\textmethod}[1]{{\small\textsf{#1}}}
\newcommand{\textds}[1]{{\scriptsize\texttt{#1}}}
\newcommand{\textfn}[1]{{\textit{#1}}}

\newcommand{\metriclabels}[0]{}

\DeclareMathOperator*{\argmin}{argmin}
\newcommand{\indic}[1]{\mathbbm{1}[#1]}
\newcommand{\prob}[1]{\textnormal{Pr}\left(#1\right)}

\newcommand{\R}{\mathbb{R}}

\newcommand{\xb}{\bm{x}}
\newcommand{\X}{\mathcal{X}}
\newcommand{\Y}{\mathcal{Y}}

\newcommand{\clf}[1]{\hat{h}^{#1}}

\newcommand{\Hset}[0]{\mathcal{H}}

\newcommand{\data}[1]{D^{#1}}
\newcommand{\n}[1]{n^{#1}}

\newcommand{\nnew}[1]{n_{\textrm{new}}^{#1}}

\newcommand{\ab}[0]{\bm{a}}

\newcommand{\st}{\textnormal{s.t.}}

\begin{document}

\iftoggle{eaamo}{
    \title{Algorithmic Censoring in Dynamic Learning Systems}
    \author{Jennifer Chien}
    \email{jjchien@eng.ucsd.edu}
    \affiliation{\institution{UC San Diego}\country{USA}}
    \author{Margaret Roberts}
    \email{meroberts@ucsd.edu}
    \affiliation{\institution{UC San Diego}\country{USA}}
    \author{Berk Ustun}
    \email{berk@ucsd.edu}
    \affiliation{\institution{UC San Diego}\country{USA}}
}

\iftoggle{arxiv}{    
\title{Algorithmic Censoring in Dynamic Learning Systems}
\author{%
\name Jennifer Chien \email jjchien@eng.ucsd.edu\\ \addr University of California, San Diego 
    \AND 
    \name Margaret Roberts\email meroberts@ucsd.edu\\
    \addr University of California, San Diego 
    \AND
    \name Berk Ustun\email berk@ucsd.edu\\
    \addr University of California, San Diego
    }%
}

\begin{abstract}
Dynamic learning systems subject to selective labeling exhibit \emph{censoring}, i.e. persistent negative predictions assigned to one or more subgroups of points. In applications like consumer finance, this results in groups of applicants that are persistently denied and thus never enter into the training data. In this work, we formalize censoring, demonstrate how it can arise, and highlight difficulties in detection. We consider safeguards against censoring -- recourse and randomized-exploration -- both of which ensure we collect labels for points that would otherwise go unobserved. The resulting techniques allow examples from censored groups to enter into the training data and correct the model. Our results highlight the otherwise unmeasured harms of censoring and demonstrate the effectiveness of mitigation strategies across a range of data generating processes. 
\end{abstract}

\iftoggle{arxiv}{
\vspace{0.25em}
}

\maketitle

\section{Introduction}
\label{Sec::Introduction}

Machine learning (ML) models are often used in dynamic learning systems that combine training, prediction, and data collection. In applications like consumer finance, hiring, data-driven discovery, and disease screening, these systems are subject to \emph{selective labeling}, meaning we only collect labels for instances that are assigned a positive prediction~\citep[][]{lakkaraju2017selective}. 

Dynamic learning systems that are subject to selective labeling exhibit \emph{censoring} -- i.e., they may fail to assign a positive prediction to one or more subgroups of points
\footnote{This usage of censoring differs from the historical use, i.e. where unacceptable parts are suppressed. Our usage of censoring relates to that used in survival analysis where, e.g., we may not observe data for a population of interest due to dropout/death.}. 
Such subgroups are repeatedly assigned negative predictions despite model updates, thus never entering into the training data. This makes it difficult for the system to correct mistakes or adapt to new patterns.

Censoring is a blind spot of dynamic learning systems -- one that is not necessarily good or bad. Consider a model for loan approval that censors a sub-population of applicants (e.g., individuals without credit history). Censoring may be justified for an applicant who would have ultimately defaulted on their loan.  Conversely, it may be unjustified if the applicant would have ultimately repaid their loan. The underlying issue is in the \emph{ambiguity}: one cannot distinguish between these two populations because we will never approve the applicant and thus never observe their true label.

The potential harms of censoring are magnified through feedback loops. In dynamic learning systems, predictions influence current decisions and data collected to train future models. Consider a consumer who lacks credit history applies for a loan. Their credit report is input into a model that outputs a probability of repayment and makes a subsequent decision. For this consumer, the lack of credit history results in a low predicted probability and subsequent rejection. This results in two costs for the consumer: 1) they do not receive the loan, 2) they are unable to build credit history to strengthen their loan application. This puts them at risk for becoming a ``credit invisible'', i.e. an applicant who is locked out of the credit system due to their inability demonstrate creditworthiness~\citep[][]{kozodoi2021credit}. In the United States, this ``catch 22'' affects approximately 26 million consumers~\citep[][]{blattner2021costly,brevoort_grimm_kambara_2015,nocredithistory}. This is just one example of an individual perpetually denied consideration of a social system through algorithms alone. Censoring can lead to issues across various machine learning applications:
\begin{itemize}[label={}, leftmargin=0pt]
    
    \item \emph{Hiring}: Companies use models to screen resumes and make decisions about which candidates to interview ~\citep[][]{AI-resume-screen}. These systems exhibit censoring as certain subgroups of applicants will never be interviewed nor hired, therefore never entering into the training data. The resulting models may only hire employees similar to those historically hired. In practice, this can hurt both businesses and individuals by excluding qualified applicants~\citep[][]{li2020hiring, raghavan2020mitigating, engler-hiring, engler-hirevue}. 
    
    \item \emph{Disease Screening}: Medical practitioners may train disease detection models on pre-existing datasets~\citep[][]{sengupta2019machine, zhao2022biomedical}. In dermatology, models are used to predict which lesions should be subject to further inspection~\citep[][]{soenksen2021using}. A model that systematically misclassifies a subset of melanoma-prone lesions may put the corresponding individuals at a higher risk of cancer and/or death. Beyond a consistently incorrect negative prediction, they may also never entering the training dataset to correct the model.
    
    \item \emph{Data-Driven Discovery}: In materials science, drug discovery, and chemistry, scientists obtain recommendations for promising experiments based on models trained on small labelled datasets~\citep[][]{raccuglia2016machine, bergen2019machine, pollice2021data}. In drug discovery, for example, scientists may fit a model to predict if a given chemical compound can act as a antibiotic. Compounds that receive higher predicted probabilities are then slated for a confirmatory experiment. In such settings, models may consistently miss a promising subset of molecules that are dissimilar to historical positive examples. In turn, censored molecules will never be assigned high probabilities and never tested in confirmatory experiments. In this case, censoring could lead to less efficient discovery processes or missed discoveries.
    
\end{itemize}

\paragraph{Contributions} The main contributions of this work include:
\begin{enumerate}[leftmargin=*]

\item We define and formalize censoring, describe why it is difficult to detect, and how it may lead to harms.

\item We demonstrate how censoring can arise and catalog four such mechanisms: sample selection bias, heterogeneity in data quality, operational changes, and distributional shifts.

\item We describe general-purpose strategies to safeguard against censoring, characterizing their costs and benefits for different stakeholders.

\item We present a comprehensive study to benchmark general-purpose approaches to resolve and safeguard against censoring. Our study evaluates each technique on dynamic settings that vary in terms of the complexity of the data distributions, propensity for strategic manipulation, and the ability to target the censored group. We conclude that recourse is best in settings with causal features, as it provides a directed exploratory approach to censoring-group validation. We propose a combination of recourse and randomization as a solution for unknown causal settings.

\end{enumerate}

\paragraph{Related Work}

We explore the implications of selective labeling and censoring in dynamic learning systems (DLSs). Prior work studies the implications of selective labeling in a static setting~\citep[][]{lakkaraju2017selective, blum2019recovering, de2018learning, coston2021characterizing}. We consider this effect in dynamic environments~\citep[][]{lum2016predict,liu2018delayed,kallus2018residual,gu2019understanding, damour2020fairness,hu2018short, jiang2021learning} that involve multiple rounds of data collection, training, and prediction~\citep[][]{ensign2018runaway, de2018learning, hashimoto2018fairness, jiang2021learning}. Our work shows that these systems may exhibit feedback loops that lead to persistent denial of individuals. We demonstrate the difficulty in detecting this effect and identify its impact to model-owners as well as individuals. Prior work explores leveraging unlabelled data an intervention at model training~\citep[][]{lee2003learning, elkan2008learning, hu2021predictive, rateike2022don, aka2021measuring}, whereas we examine post-training safeguards: randomization and recourse. 

The traditional approach to resolve censoring in the reinforcement learning literature is to explore~\citep[][]{ensign2017decision,KilGomSchMuaVal20, chen2020strategic, kulkarni2020social, brown2020performative, Dai2020LabelBL, krauth2020offline, mansour2021bayesian, erdelyi2021randomized, simchowitz2021exploration, immorlica2018incentivizing, wei2020optimal, chandak2020reinforcement} -- i.e., to build dynamic learning systems that collect data that would inform future decisions. Exploration may be ill-suited for certain applications that exhibit censoring because they involve randomization. In effect, a randomized policy may be unethical in consumer-facing applications like hiring, lending, and disease-screening, as individuals who are approved at random may be subject to harms such as job insecurity, poor credit, and unnecessary medical tests~\citep[][]{goldstein2018ethical, royall1991ethics, baunach1980random, colli2014ethical}. More broadly, these techniques may also be ill-suited because exploratory policies are only guaranteed to optimize population measures of long-term utility or regret. These measures may not be responsive to censoring because they are too broad to capture the impact of a small group of individuals. 

Recourse, i.e. when providing individuals with actionable changes to obtain a desired predicted outcome, is another solution to safeguard against censoring~\citep[][]{ustun2019actionable,venkatasubramanian2020philosophical,karimi2020algorithmic,karimi2020survey,von2020fairness}. Work in strategic classification shows that recourse can promote gaming when provided on non-causal features, resulting in a change in the predicted outcome without a corresponding change in the true outcome. Causal recourse actions -- i.e., actions that change both the predicted outcome and true outcome -- however, may lead to improvement~\citep[][]{bechavod2020causal, bechavod2021gaming, shavit2020learning, chen2019learning, milli2019social, levanon2021strategic}. Our work complements this by exploring additional failure modes, such as that due to model shift~\citep[see e.g.,][]{upadhyay2021towards,rawal2020can},  retraining~\citep[][]{ross2021learning}, and by considering recourse with explicit guarantees, or a combination of recourse and exploration.

The most relevant work to ours is that of~\citet[][]{KilGomSchMuaVal20}, which studies a similar setting through the lens of reinforcement learning. Their approach casts models as ``approval policies" in a finite-horizon Markov decision process and seeks to find a model that maximizes utility for a model-owner (e.g., a lender) under a group fairness constraint. Although their work does not study censoring, shows that deterministic policies will achieve sub-optimal regret due to insufficient exploration. This is one of the potential consequences of censoring, as models may assign incorrect predictions to individuals who would have positively impacted utility.

\section{Framework}
\label{Sec::Preliminaries}
In this section, we formalize the notion of censoring in dynamic learning systems.
\newcommand{\m}[1]{m^{#1}}
\newcommand{\xt}[1]{x}
\newcommand{\yt}[1]{y}
\newcommand{\xbt}[1]{\xb}
\newcommand{\YC}{\Y^\text{S}}
\newcommand{\clfp}[1]{\hat{p}^{#1}}
\newcommand{\clfd}[2]{\hat{y}^{#1}_{#2}}
\newcommand{\phat}[1]{\hat{p_{#1}}}
\newcommand{\neval}[1]{n_\textnormal{eval}^{#1}}

\subsection{Preliminaries}

We consider a sequential decision-making process over $t = 1,\ldots, T$ with $T \geq 2$ periods. We initialize the process with $\n{1}$ initial training examples denoted $\data{1} := (\xb_i, y_i)_{1}^{\n{1}}$. Each example consists of feature vector $\xb_i = [x_{i,1}, \ldots, x_{i,d}] \in \R^{d}$ and label ${y_i \in \{-1, 1\}}$ where $y = 1$ denotes a target class (e.g., repaid loan within 2 years). We denote the true probability of the target outcome as $p^t: \X \to \prob{Y=1~|~X=\xb_i}$ and the predicted probability from the model as $\clfp{t}: \X \to \prob{Y=1~|~X=\xb_i}$. Each period applies the following three steps:
\begin{enumerate}[leftmargin=0.25in]
    
    \item \emph{Fit}: We use the dataset $\data{t}$ to fit a probabilistic classifier  $\clf{t}{}: \X \to [0,1]$ via standard empirical risk minimization: $$\clfp{t}{} \in \argmin_{h \in \Hset} \hat{R}(h; \data{t})$$ where $\hat{R}(h; \data{t})$ denotes the empirical risk (e.g., negative log-likelihood) and $\Hset$ denotes the model class. We assume that  model owners convert the probability predictions from $\clf{t}{}$ into predictions by a threshold $\rho \in [0,1]$ so that $\clfd{}{\rho}(\xb_i) = 1$ if and only if $\clfp{}(\xb_i) > \rho]$.
    
    \item \emph{Predict}: We use the model $\clfd{t}{}$ to assign predictions to all incoming examples in period $t$: $\neval{t} := n_\textnormal{new}^{t} + n_\textnormal{ret}^{t}$. We use $\nnew{t}$ to denote new points that are sampled i.i.d. from an unknown data distribution, and $n_\textnormal{ret}^{t}$ to denote instances assigned negative predictions in previous periods $t' < t$.

    \item \emph{Collect}: We append labelled examples to the training dataset. Given that the system exhibits selective labeling, we observe $\xb_i$ for all $\neval{t}$ points at prediction time, and observe labels for instances assigned the prediction $\clfd{t}{}(\xb_i) = 1$. We add $\n{t+1}$ examples to the training data, where $$\n{t+1} := \left|\data{t+1}\right| = \sum_{k = 1}^{t} \sum_{i = 1}^{\neval{k}} \indic{\clfd{k}{}(\xb_i) = 1} = \n{t} + \sum_{i = 1}^{\neval{t}} \indic{\clfd{t}{}(\xb_i) = 1}$$. We note that the observed label is probabilistic and therefore not all observed data has the desired outcome.
    
\end{enumerate}

\subsection{Censoring}
\label{Sec::Censoring}

\begin{definition}[Censoring]
A dynamic learning system censors instances with features $\xb$ if it fails to observe a label $y(\xb)$ over the entirety of deployment lifetime without intervention. In other words, all models predict ${\clfd{t}{}(\xb) = -1}$ for all future periods $t \in [t,\ldots,T]$. We denote $c^t(\xb) = -1$ if the system censors instances with features $\xb$ from periods $t$ onwards and $c^t(\xb) = 1$  if the system does not censor instances with features $\xb$ from period $t$ onwards.
\end{definition}

Censoring characterizes a blind spot of DLSs, one that does not necessitate a failure in performance. Consider a lending system that censors applicant $i$. By definition, this means that the system will assign negative predictions to any instance with features $\xb_i$ in perpetuity. This phenomenon may be optimal if the predicted and true label are in agreement subject to some tolerance $\tau$: $\prob{\clfd{}{}(\xb_i) = y(\xb_i)} > \tau$. Conversely, the system performs suboptimally if the predictions are not in agreement -- i.e., $\prob{\clfd{}{}(\xb_i)= y(\xb_i)} < \tau$. In a system exhibiting censoring, the system's optimality is unverified and unknown, as we never observe the outcome label for the censored group. 

Given this blind-spot, censoring causes problems in DLS when: (1) the system \emph{presumes} that the label for the point is negative even though it may be positive, and (2) the system does not \emph{validate} this assumption nor correct this mistake. This behavior is particularly dangerous in various settings: i.e., those with high label noise, where the model was initialized on a small training set, or where the relationship between $\xb_i$ and $y_i$ drifts over time. We examine such settings in \cref{Sec::Causes}, where we induce censoring on immutable covariate $z=1$ where $c_i = z_i = 1$ for systems with censoring with no intervention. We use $z=1$ to calculate metrics on the initially censored group to measure the unobserved harms and efficacy of resolving censoring when safeguards are deployed.

\section{Mechanisms that Induce Censoring}
\label{Sec::Causes}

In this section, we describe four mechanisms that lead to censoring: sample selection bias, heterogeneity in data quality, operational changes, and distributional shifts, although these are not exhaustive. 

\paragraph{Preliminaries}

For each mechanism, we provide an empirical example and report the observed and true AUC to demonstrate difficulty in detection. The true AUC is calculated on an unbiased sample from the underlying data distribution while the observed AUC is calculated on the training data collected over time.  

For clarity, the group of censored points have covariate $z = 1$ censored points using the \emph{censoring indicator} $z_i := \clfd{t}{}(\xb_i)=-1$ for $t \in \{t',\ldots,T\}$. Note that ${z_i = 1 \Leftrightarrow z_i' = 1}$ for any points $i$ and $i'$ such that ${\xb_i = \xb_{i'}}$ -- i.e., when a system has censored a point, it censors all other points with the same features. In practice, censoring may affect groups defined by multiple subsets of features, further exacerbating difficulty of detection. For example, a lending system could censor applicants below the age of 35 without a credit card -- i.e., ${\xb \in x_{i,j} \land x_{i,j'}}$ where ${x_{i,j} = 1[\textfn{age} \leq 35]}$ and ${x_{i,j'} = 1[\textfn{has\_credit\_card} = \textfn{FALSE}]}$. Each mechanism can cause censoring in deployed systems, be it individually or simultaneously. 

\begin{figure}[htbp]
    \centering
    \resizebox{0.7\linewidth}{!}{
    \begin{tabular}{rcc}
    \cell{l}{\\\bf Outcome} & \cell{c}{\textbf{Population}\\$z \in \{-1,1\}$} & \cell{c}{\bf{Censored Group}\\$z=1$}  \\ 
    \cmidrule(lr){1-1} \cmidrule(lr){2-2} \cmidrule(lr){3-3}
    \cell{c}{$y \in \{-1,1\}$} &
    \cell{c}{\includegraphics[trim={0 0 1cm 1cm}, clip, width=0.3\linewidth]{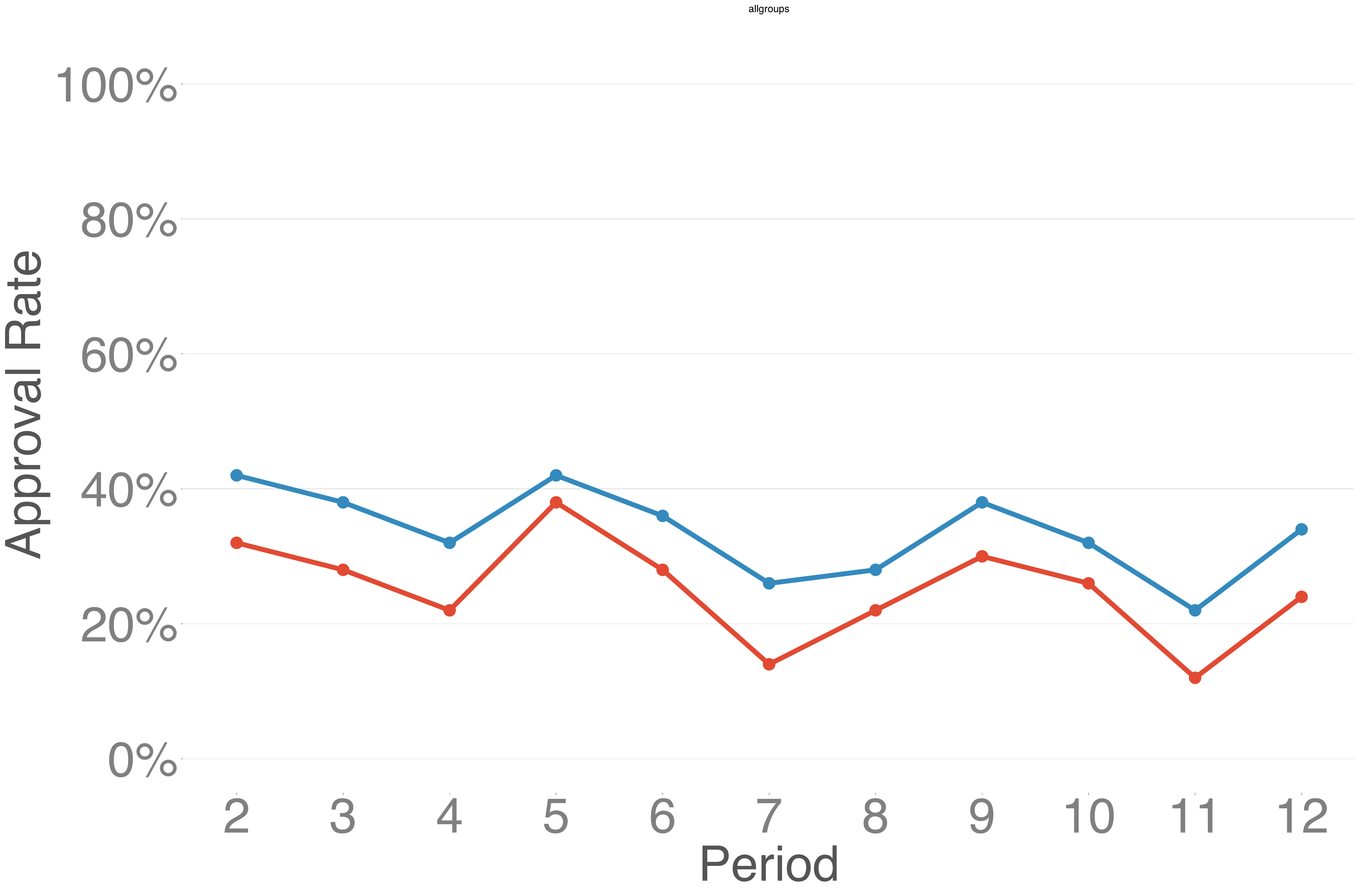}} & 
    \cell{c}{\includegraphics[trim={0 0 1cm 1cm}, clip, width=0.3\linewidth]{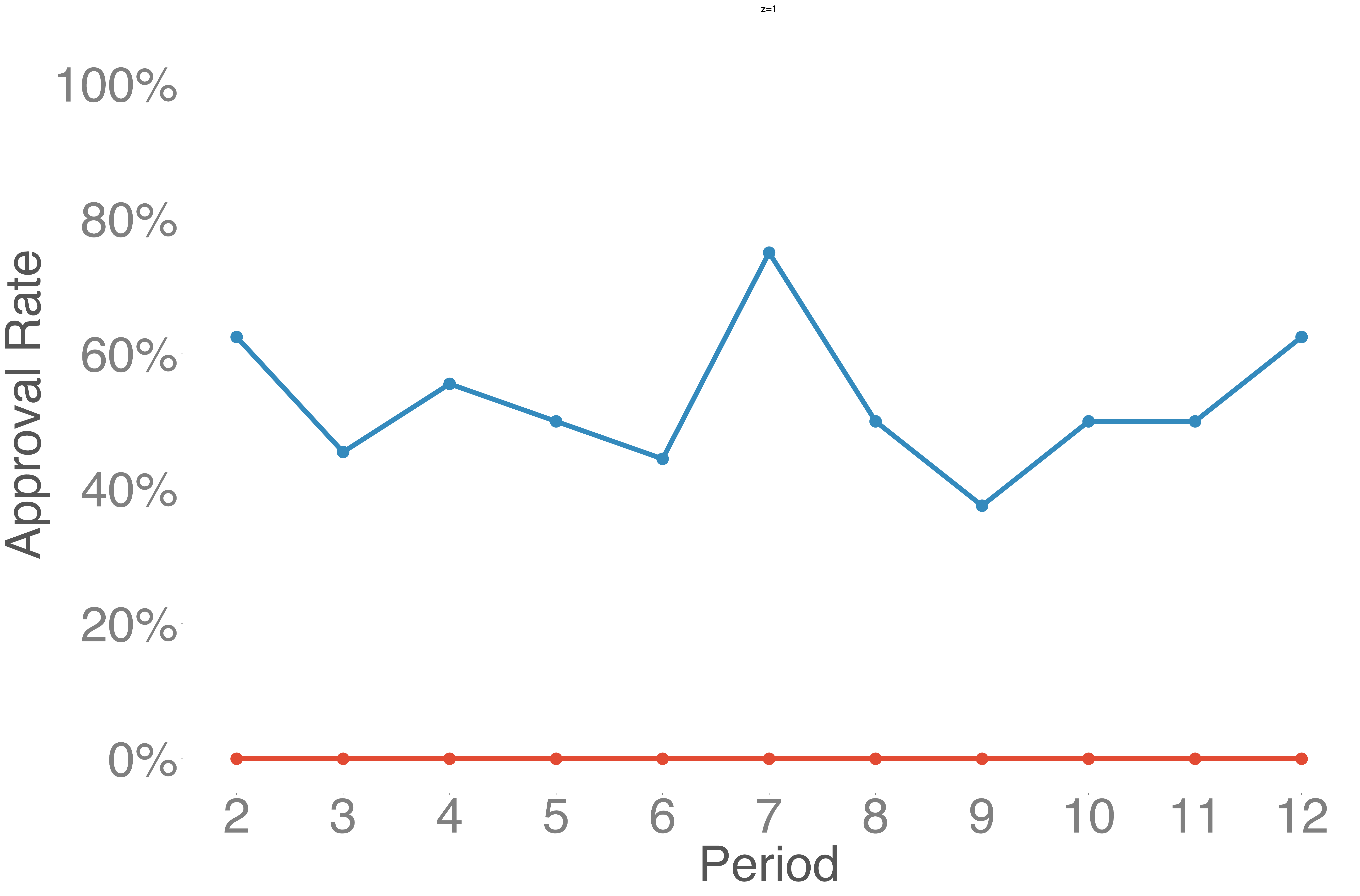}} \\ 
    \cell{c}{$y = 1$} &
    \cell{c}{\includegraphics[trim={0 0 1cm 1cm}, clip, width=0.3\linewidth]{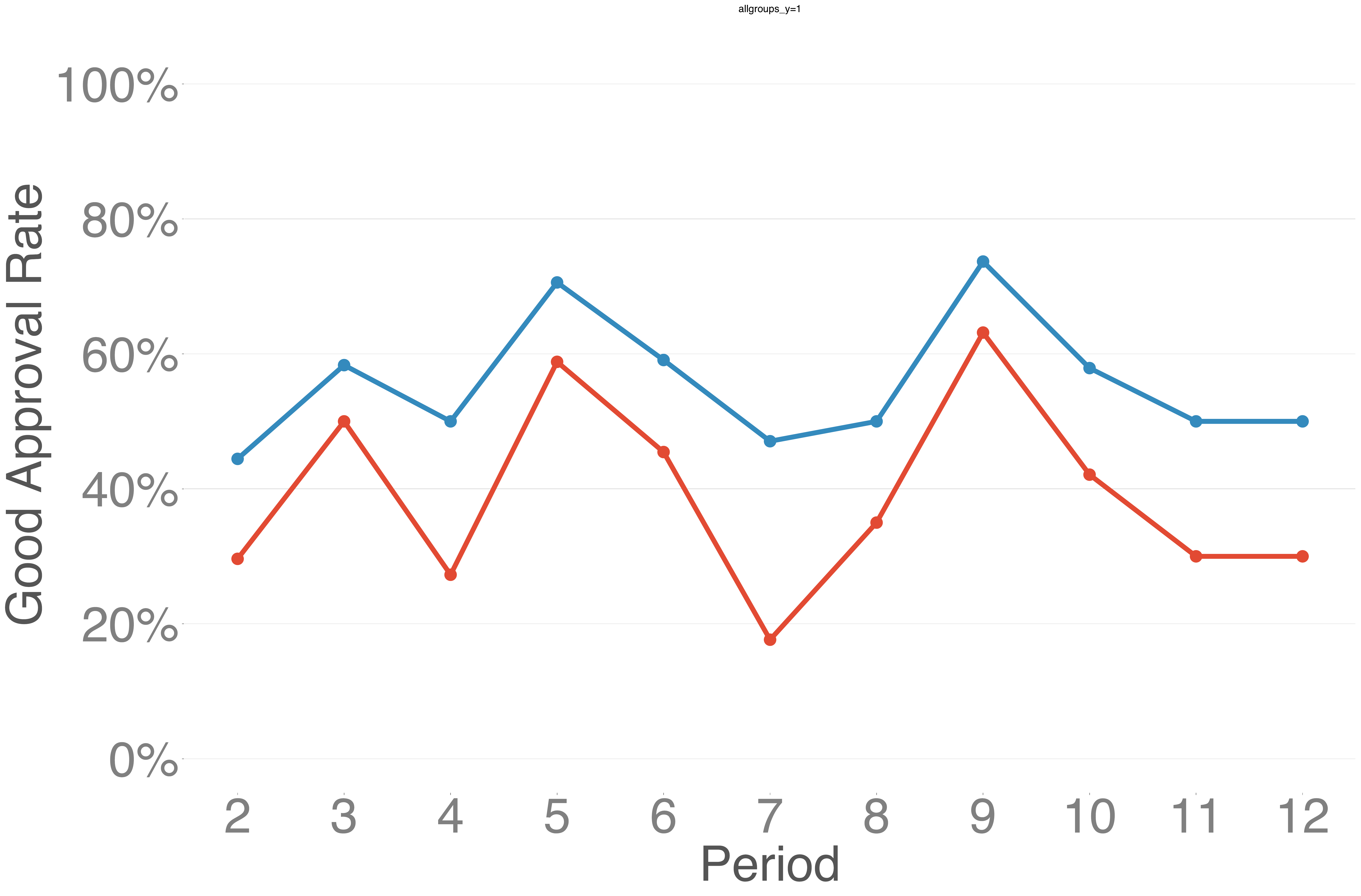}} & 
    \cell{c}{\includegraphics[trim={0 0 1cm 1cm}, clip, width=0.3\linewidth]{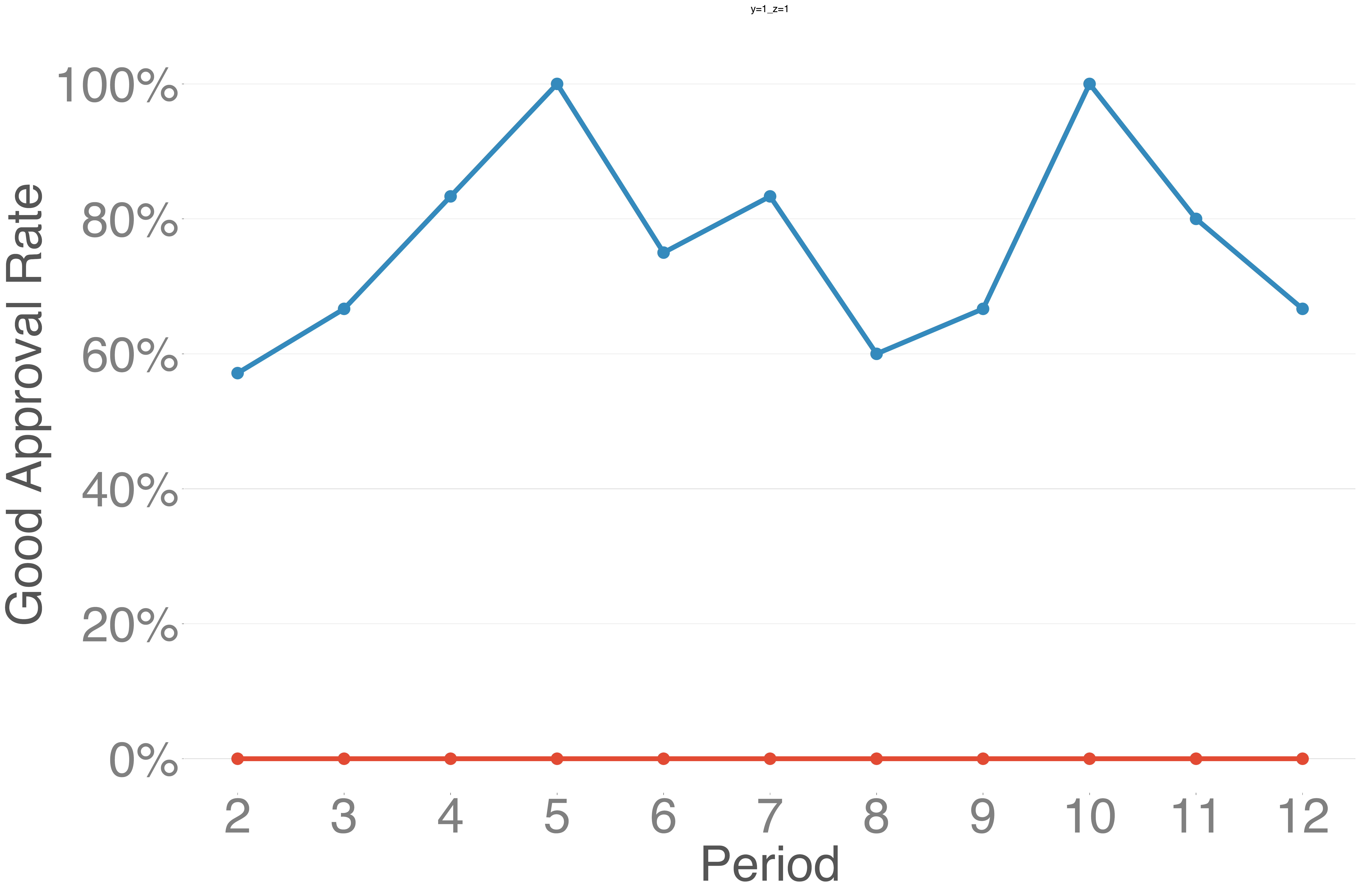}} \\
    \end{tabular}
    }
    \caption{Unrepresentative training data leads to censoring. We train models by sampling from \textds{causal} data generating process in \cref{Table::DAGs}. We show a system initialized with a baseline unbiased sample (blue) and an unrepresentative sample where we have no data for individuals with features $z_i = 1$ and outcome $y_i = 1$ (red). As shown, the model trained on the unrepresentative sample censors all $z_i = 1$ points. We demonstrate that censoring is difficult to detect via standard performance metrics -- the training AUC at the last period is similar: 0.811 (unrepresentative training data) and 0.795 (representative training data), while the true AUC, is much worse for unrepresentative training data (0.660) than representative training data (0.756). }
    \label{Fig::UnrepresentativeTraining}
\end{figure}

\paragraph{Sample Selection Bias}
Censoring can arise when a model is trained on data unrepresentative of the deployment population. This occurs when model-owners use readily available data, such as in already deployed products. This can be favorable to model-owners due to the convenience and accessibility of readily available data and the prohibitive costs of collecting new data (time, money, personnel). These costs can be particularly discouraging when attempting to estimate the requisite sample size for unknown or low occurrence subgroups. For instance, historical roots of lending in the US include institutionalised ``redlining'', disfavoring applicants from neighborhoods that were ``within such a low price or rent range as to attract an undesirable element,'' which often meant near predominantly Black neighborhoods~\citep{rice2013discriminatory, greer2012race, national2012banks}. Black people and people of color more generally were deemed high risk, refused loans, and never entered into the training data, leaving their true repayment status unrealized. This is used to perform data-driven risk prediction today~\cite{dobbie2021measuring, bhutta2022much}.

We demonstrate how this leads to censoring in a system with an initial model trained on an unrepresentative training dataset in Figure \ref{Fig::UnrepresentativeTraining}. Here, $\data{1}$ is biased as it is missing $y=1$, $z=1$ points, leading it to mistakenly predict all $z=1$ points to be negative. As a result, the approval rate for all $z=1$ points is $0$ for the entirety of the system's lifetime, despite making up 15\% of the population. This pattern persists unresolved over multiple rounds of data collection, retraining, and deployment and is not reflected in the AUC.

\paragraph{Heterogeneity in Data Quality}
Censoring can also arise from heterogeneity in the data distribution. For labels, noise can arise from intrinsic ambiguity due to human annotation~\citep[][]{khanal2021does, yan2014learning, veit2017learning, chen2017learning, wei2021learning}. For features, noise can arise from intrinsic difficulty in measurement~\cite[][]{zhang2021learning}. In lending in the US, for example, models use features that are measured with higher rates of noise for historically marginalized groups~\citep[e.g., credit history][]{blattner2021costly, kelley2021anti}. These individuals are referred to as ``credit invisibles''\citep[][]{brevoort_grimm_kambara_2015} and are assigned uncertain predictions. Such predictions leads to perpetual denial and censoring: such denied individuals cannot produce labeled data and therefore increase their credit history due to the original prediction.

We demonstrate how censoring can arise via heterogeneity in label noise in \cref{Fig::noisylabelsex2}. We consider a system where examples have some probability of exhibiting the desired true outcome and compare two systems: one with label noise and one without. In the system without label noise, a Bayes optimal classifier correctly approves all $\xb=1$ points. However, the system with label noise incorrectly assigns a negative prediction to all points with $z=1, \xb=1$. Although we only show a single round of decision-making, this behavior will be repeated at each round of deployment.

\newcommand{\pflip}[0]{p_\textrm{flip}}
\begin{table}[htbp]
    \centering
    \resizebox{0.75\textwidth}{!}{
    \begin{tabular}{*{12}{c}}
    \multicolumn{4}{c}{} & \multicolumn{3}{c}{No Label Noise} & \multicolumn{5}{c}{With Label Noise}\\
    \cmidrule(lr){5-7} \cmidrule(lr){8-12}
    $n$ & $x$ & $z$ & \cell{c}{$\prob{y=1}$} & \multicolumn{2}{c}{\cell{c}{Observed\\Label ($y$)}} & \cell{l}{Model\\Prediction} & \multicolumn{2}{c}{\cell{c}{$\pflip{}(x, z, y)$}} & \multicolumn{2}{c}{\cell{c}{Observed\\Label ($y$)}} & \cell{l}{Model\\Prediction} \\
    \multicolumn{4}{c}{} & -1 & 1 & & -1 & 1 & -1 & 1 & ($\hat{y}$)\\
    \cmidrule{1-12}
    100 & 0 & -1 & 0 & 100 & 0 & 0 & 0 & 0 & 100 & 0 & -1\\
    100 & 0 & 1 & 0.4 & 60 & 40 & 0 & 0 & 0.4 & 76 & 24 & -1\\
    100 & 1 & -1 & 1 & 0 & 100 & 1 & 0 & 0 & 0 & 100 & 1\\
    100 & 1 & 1 & 0.7 & 30 & 70 & 1 & 0 & 0.4 & 58 & 42 & -1\\
    \end{tabular}
    }
    \caption{Example showing how an initial model trained with noisy labels and a probabilistic true outcome can lead to censoring. We train models on dataset with $n=400$ examples from groups $z \in \{-1, 1\}$. We assume that the observed labels in the training data $y$ may differ from the predicted value due to noise. As shown, a Bayes optimal model incorrectly predicts $\hat{y}(x=1, z=1)=-1$, therefore censoring such points. The observed AUC (only for positive predicted points) for the systems with no label noise and label noise are the same (0.500), while the true AUC is 0.826 and 0.738 respectively.}
    \label{Fig::noisylabelsex2}
\end{table}

\begin{figure}[htbp]
    \centering
    \vspace{-10pt}
    \resizebox{0.7\linewidth}{!}{
    \begin{tabular}{rcc}
    \cell{l}{\\\bf Outcome} & \cell{c}{\textbf{Population}\\$z \in \{0,1\}$} & \cell{c}{\bf{Censored Group}\\$z=1$}  \\ 
    \cmidrule(lr){1-1} \cmidrule(lr){2-2} \cmidrule(lr){3-3}
    \cell{c}{$y \in \{-1,1\}$} &
    \cell{c}{\includegraphics[trim={0 0 1cm 1cm}, clip, width=0.3\linewidth]{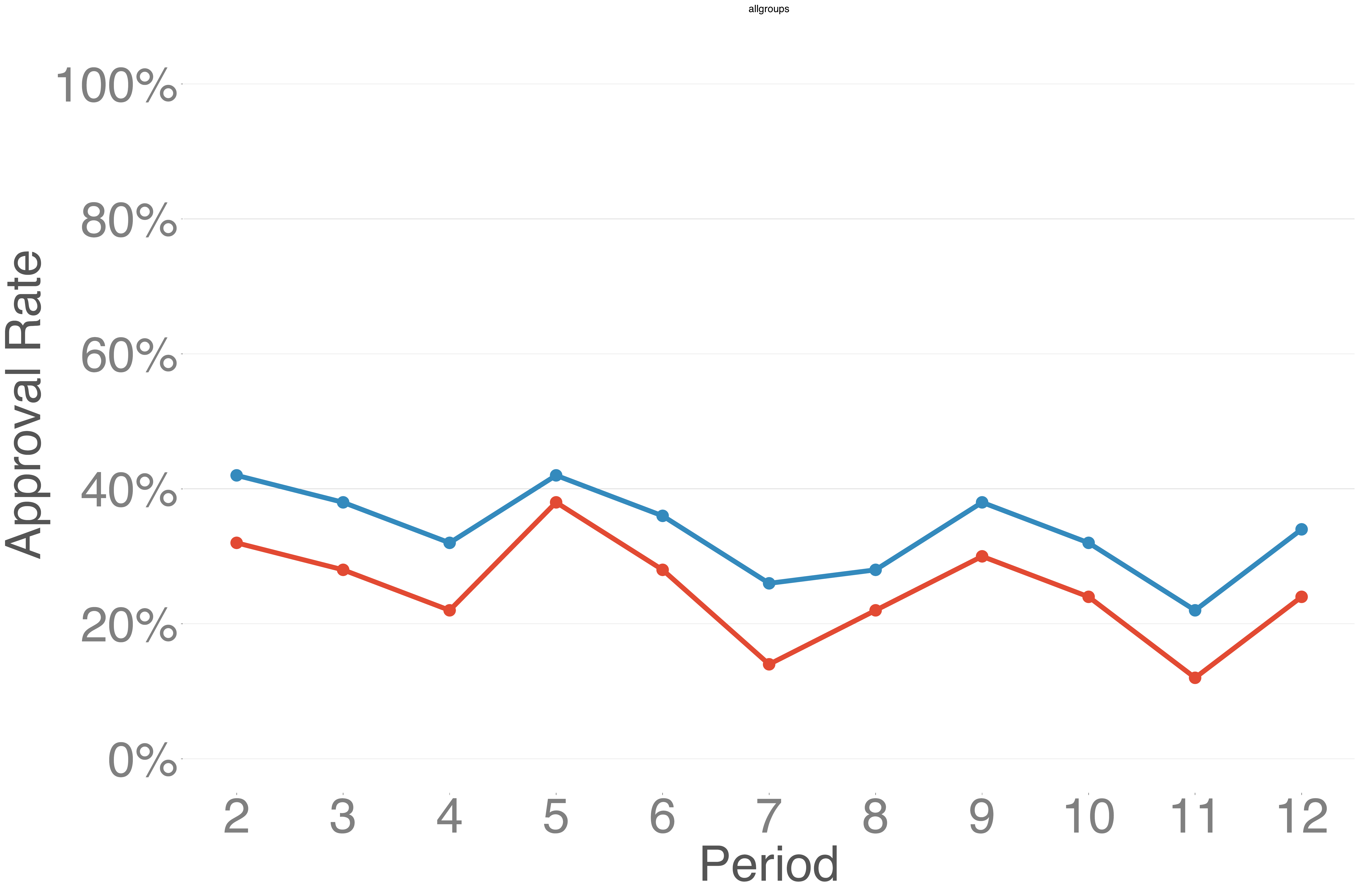}} &
    \cell{c}{\includegraphics[trim={0 0 1cm 1cm}, clip, width=0.3\linewidth]{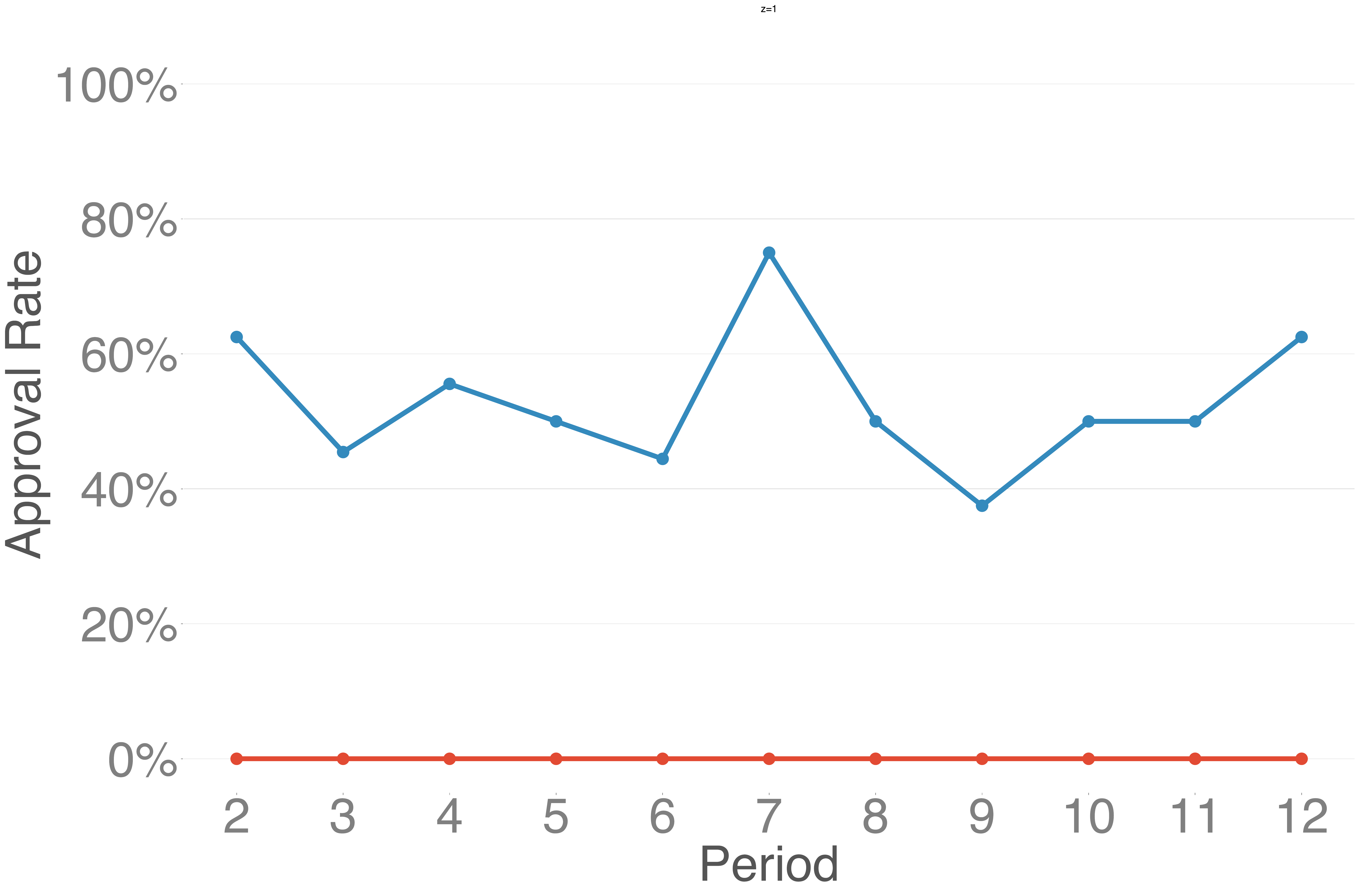}} \\ 
    \cell{c}{$y = 1$} &
    \cell{c}{\includegraphics[trim={0 0 1cm 1cm}, clip, width=0.3\linewidth]{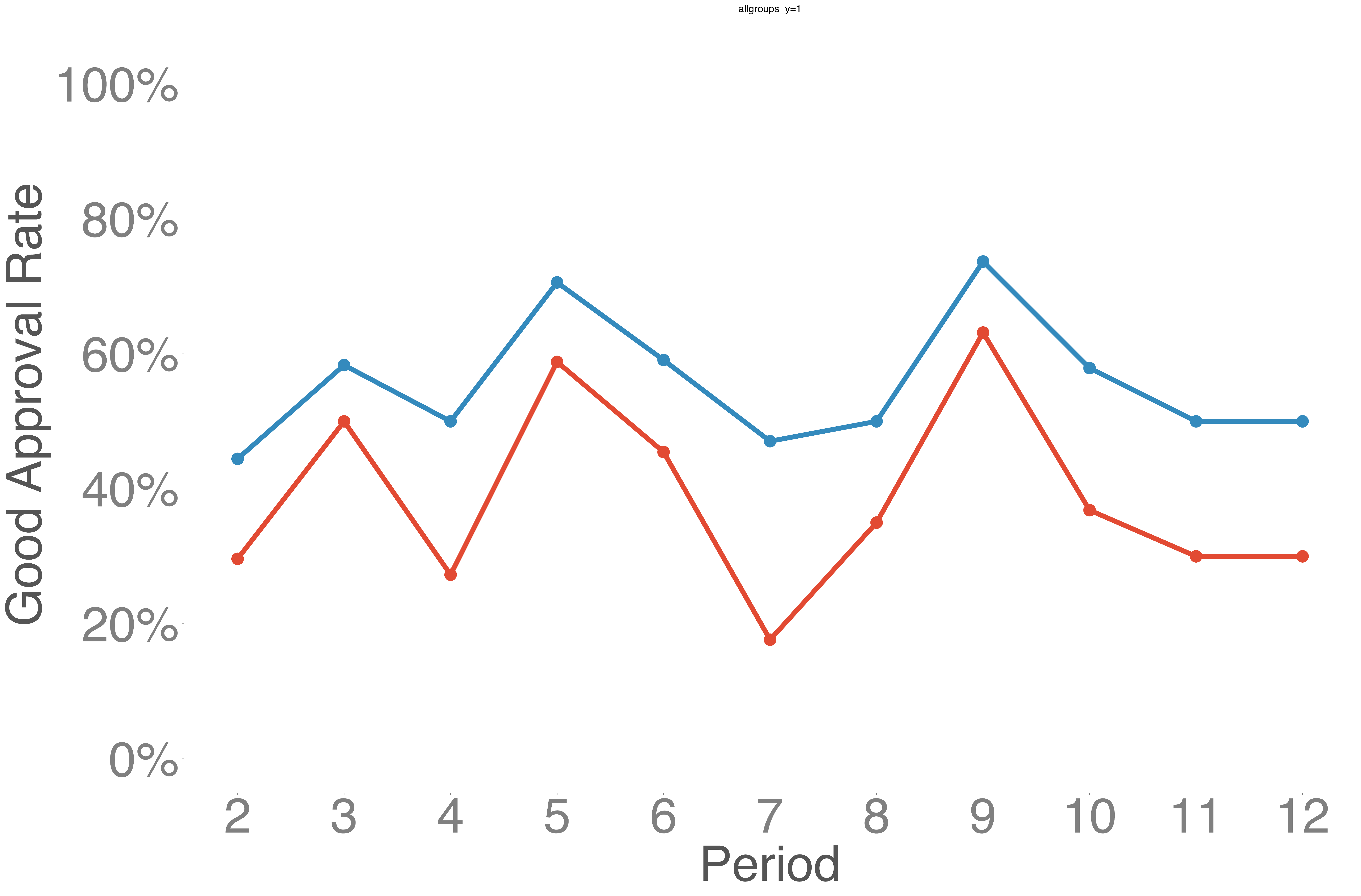}} & 
    \cell{c}{\includegraphics[trim={0 0 1cm 1cm}, clip, width=0.3\linewidth]{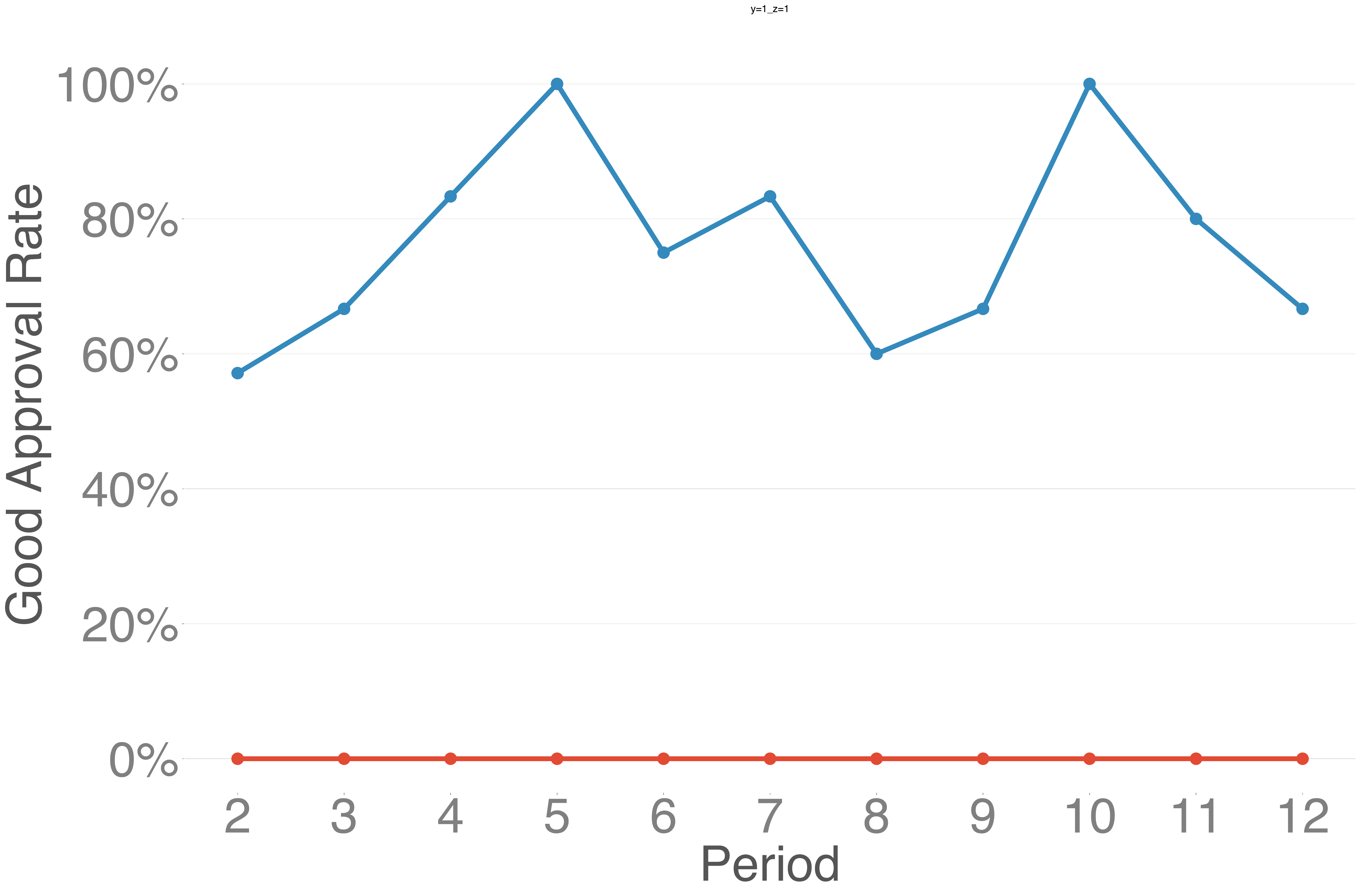}}
    \end{tabular}
    }
    \caption{Operational changes can induce censoring. We consider the \textmethod{causal} setup in \cref{Table::DAGs}. We show a system initialized on an baseline sample (blue) and a system initialized on a sample acquired from a one time adjustment to the $b$ in the first period (red). The operational change system exhibits censoring on all $z=1$ points despite similar overall approval rates. In this system, the observed AUC is 0.790 with operational changes and 0.774 in the system without operational changes. However, the true AUC is 0.638 and 0.756, respectively. 
    }
    \label{Fig::operational changes}
\end{figure}

\paragraph{Operational Changes}
\label{Sec::operationalchangesandfeatureshifts}

Censoring can also arise from operational changes -- i.e. when a model-owner adjusts the approval criteria of a pretrained model. For instance, in lending, if a model-owner implements a more conservative approval policy due to economic hardship or external regulation~\citep[][]{ECOA_2021, kumar2022equalizing, presscfpb, FHA, Akinwumi_FairLending_2022, US_bluprint_ai_BoR_2022, UK_AI_regulation_2022}, they may increase the requisite predicted repayment probability for approval. As a result, some subgroups may be excluded by the initial model, resulting in downstream propagation of these biases. Although such subgroups may have limited representation in the initial training data, the bias incurred from a single round of deployment may perpetuate the subgroup denial bias for all subsequent models even through retraining, censoring such groups in perpetuity.

We demonstrate how operational changes can lead to censoring in \cref{Fig::operational changes}. Here, we consider a stylized example where we make a one-time adjustment on the initial model $b$. We observe that after the initial $b$ change, all models denies $Z=1$ individuals for all rounds of deployment.

\begin{figure}[htbp]
\centering

    \resizebox{0.7\linewidth}{!}{
    \begin{tabular}{rcc}
    \cell{l}{\\\bf Outcome} & \cell{c}{\textbf{Population}\\$z \in \{-1,1\}$} & \cell{c}{\bf{Censored Group}\\$z=1$}  \\ 
    \cmidrule(lr){1-1} \cmidrule(lr){2-2} \cmidrule(lr){3-3}
    \\
    \cell{c}{$y \in \{-1,1\}$} &
    \cell{c}{\includegraphics[trim={0 0 1cm 1cm}, clip, width=0.3\linewidth]{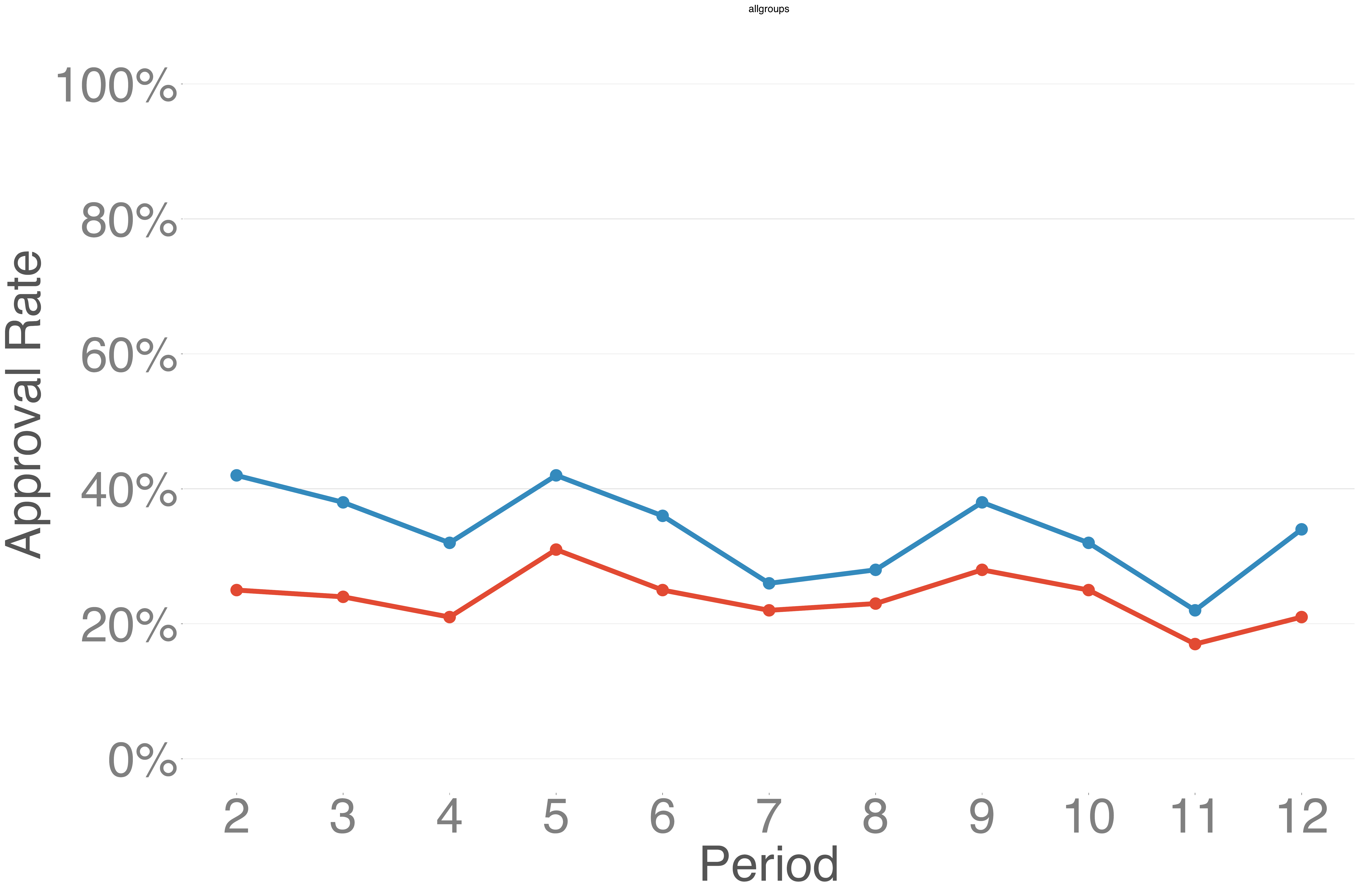}} & 
    \cell{c}{\includegraphics[trim={0 0 1cm 1cm}, clip, width=0.3\linewidth]{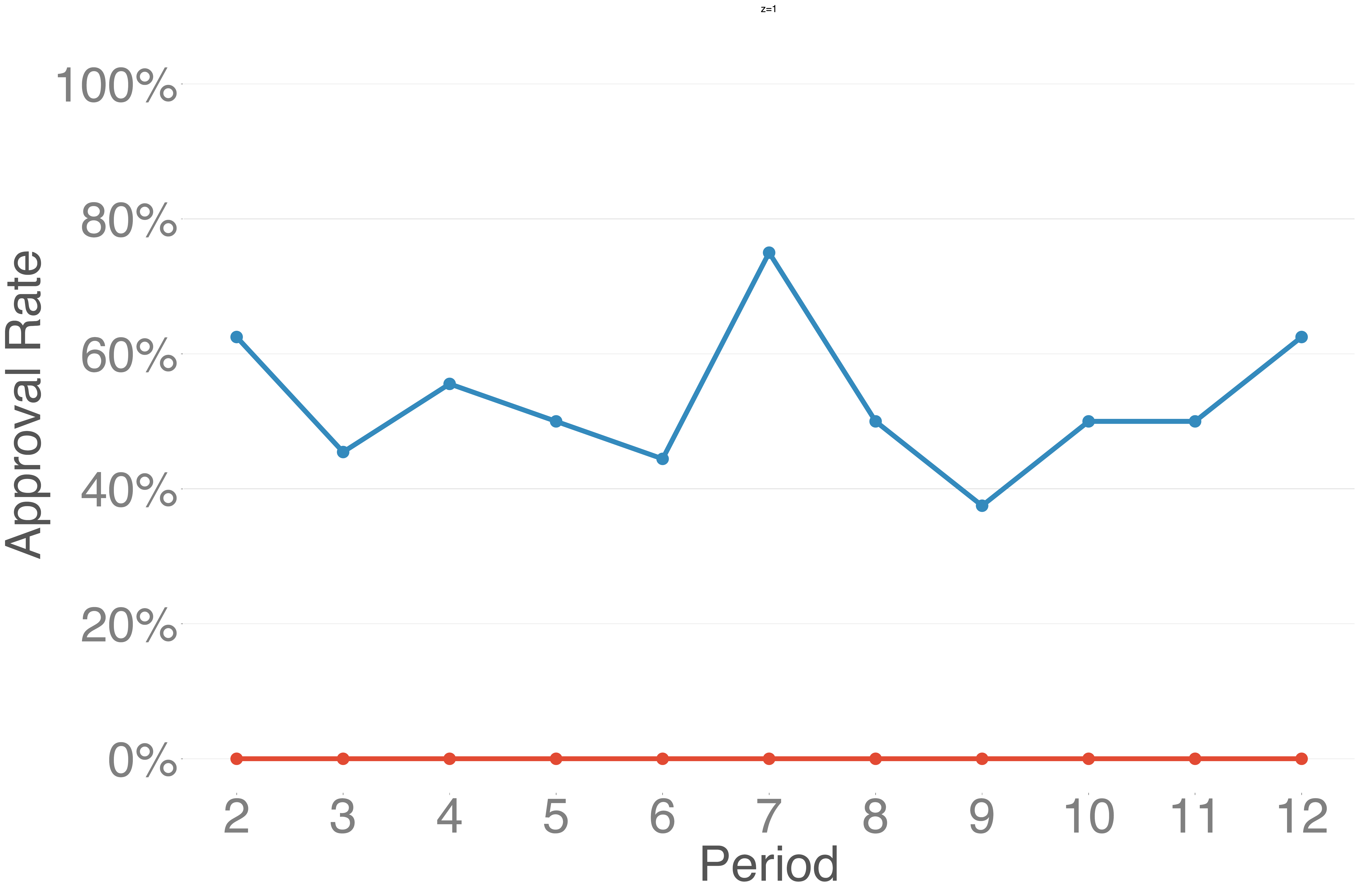}} \\ 
    \cell{c}{$y = 1$} &
    \cell{c}{\includegraphics[trim={0 0 1cm 1cm}, clip, width=0.3\linewidth]{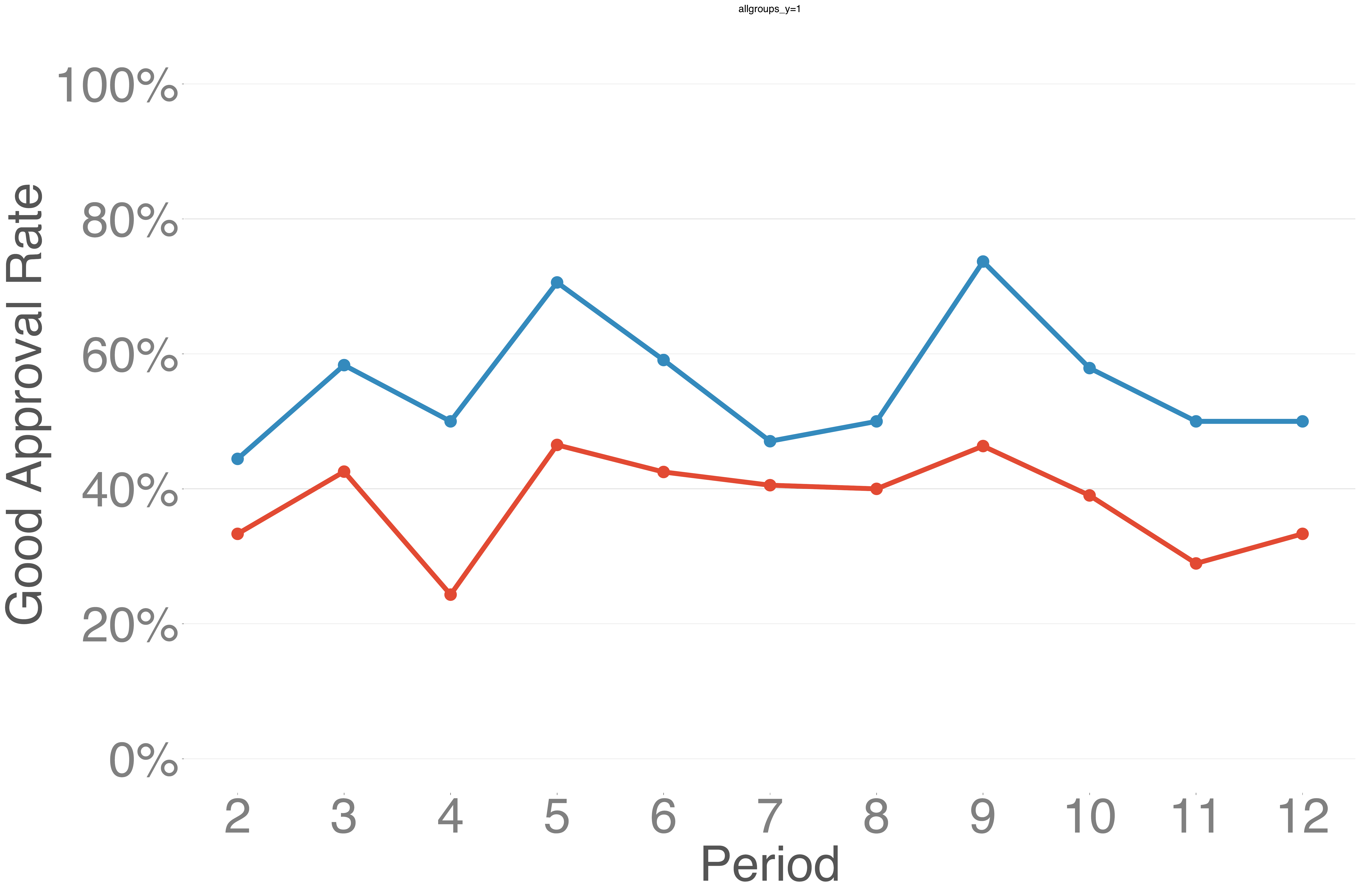}} & 
    \cell{c}{\includegraphics[trim={0 0 1cm 1cm}, clip, width=0.3\linewidth]{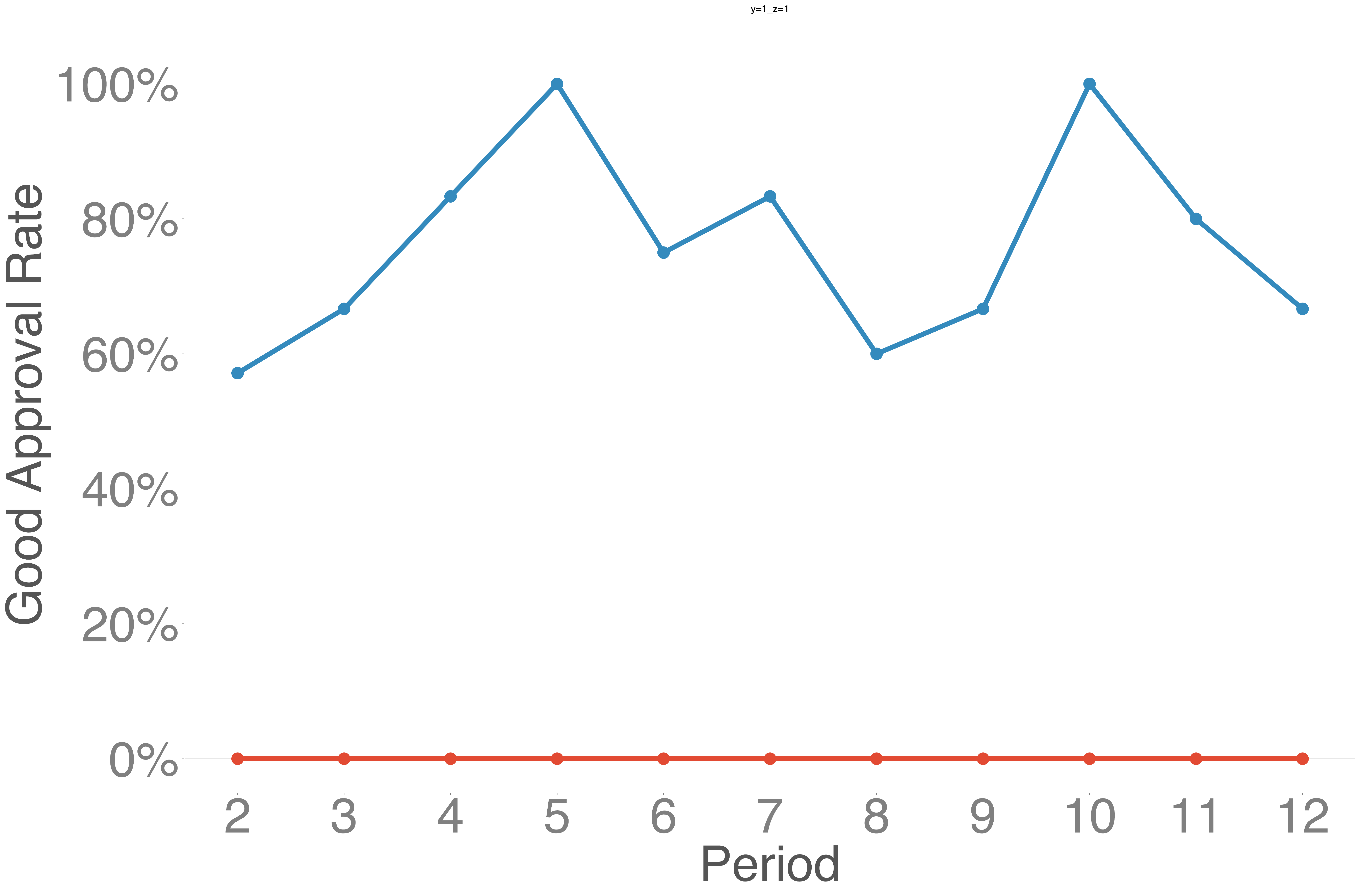}}
    \end{tabular}
    }
    \caption{Distributional shifts can induce censoring. We consider the \textmethod{causal} setup specified in \cref{Table::DAGs}. We show a system initialized with an $x_1$ feature shift for $z=1$ in initial training data -- specifically, before: $x_1\sim \textrm{Normal}(0, 1)$, after $x_1 \sim \textrm{Normal}(-5, 1)$. As shown, feature shift induces censoring of all $z=1$, being persistently denied for all periods. Again, this is difficult to detect as the observed AUC is 0.787 in for the system with feature shift and 0.756 for the system without, even though the true AUC is actually much worse for the system with feature shift (0.614) than the one without (0.756).
    }
    \label{Fig::shifts}
\end{figure}

\paragraph{Distributional Shifts}
Censoring can arise due to distributional shifts -- i.e., changes in the relationship between the features and/or labels. Feature shift can occur in many ways, we will enumerate three. 

First, the underlying feature distribution can change between training and deployment~\citep[i.e., covariate shift][]{quinonero2009dataset}. Such changes could result from broader shifts in the population. In lending, for instance, the distribution of applicants that are home-owners may change as a result of the true underlying data distribution changing, such as if homes became less affordable.This can also be a result of a shift in the observed data distribution, such as if home ownership status stops being collected, but is still included in the model. These can result in incorrect predicted risk and subsequent persistent denial of qualified applicants.

Second, feature shift can occur for a subgroup of the population ~\citep{shimodaira2000improving}. In lending, the distribution of applicants that are homeowners may change due to required features becoming opt-in instead. For applicants that might have previously had verified home-ownership, they may choose not to disclose such information, resulting in incorrect predicted risk and subsequent persistent denial of credit-worthy applicants. 

Third, features present in training that are hard to measure in deployment can cause feature shift. In lending, suppose legislation making home-ownership verification a expensive and lengthy process is passed, making it generally avoided by lenders. This may again result in incorrect predicted risk and subsequent denial towards qualified applicants. We demonstrate feature shift induced censoring for the $z=1$ population in Figure~\ref{Fig::shifts}.

\section{Mitigation}
\label{Sec::Mitigation}
Having seen some of the ways in which censoring may arise, we consider some general-purpose safeguards: randomization and recourse. We provide definitions, scope, technical challenges, and limitations. We demonstrate that these safeguards resolve censoring through empirical experiments in Section \ref{Sec::Experiments}.

\subsection{Exploration}
One of the most studied approaches to exploration is randomized or stochastic decision rules. Randomized exploration allows for a proportion of negative-predicted points to enter into the training dataset, allowing model-owners to validate the model for censored subgroups. If a model assigns incorrect predictions to a subgroup, these examples can inform future models. 

\paragraph{Setup} 
We describe methods for randomized exploration: uniform, which randomly selects instances for points assigned negative prediction; and inverse propensity score weighting (IPW), which weighs the probability of selection inversely proportional to the assigned probability $\clfp{}{}$. This assigns a greater exploration likelihood for points unlikely to be observed through model shift from retraining alone.

\begin{figure}[htbp]
    \centering
    \resizebox{0.7\linewidth}{!}{
    \begin{tabular}{rcc}
    \cell{l}{\\\bf Outcome} & \cell{c}{\textbf{Population}\\$z \in \{-1,1\}$} & \cell{c}{\bf{Censored Group}\\$z=1$}  \\ 
    \cmidrule(lr){1-1} \cmidrule(lr){2-2} \cmidrule(lr){3-3} \\
    \cell{c}{$y \in \{-1,1\}$} &
    \cell{c}{\includegraphics[trim={0cm 8cm 0cm 7cm}, clip, width=0.25\linewidth]{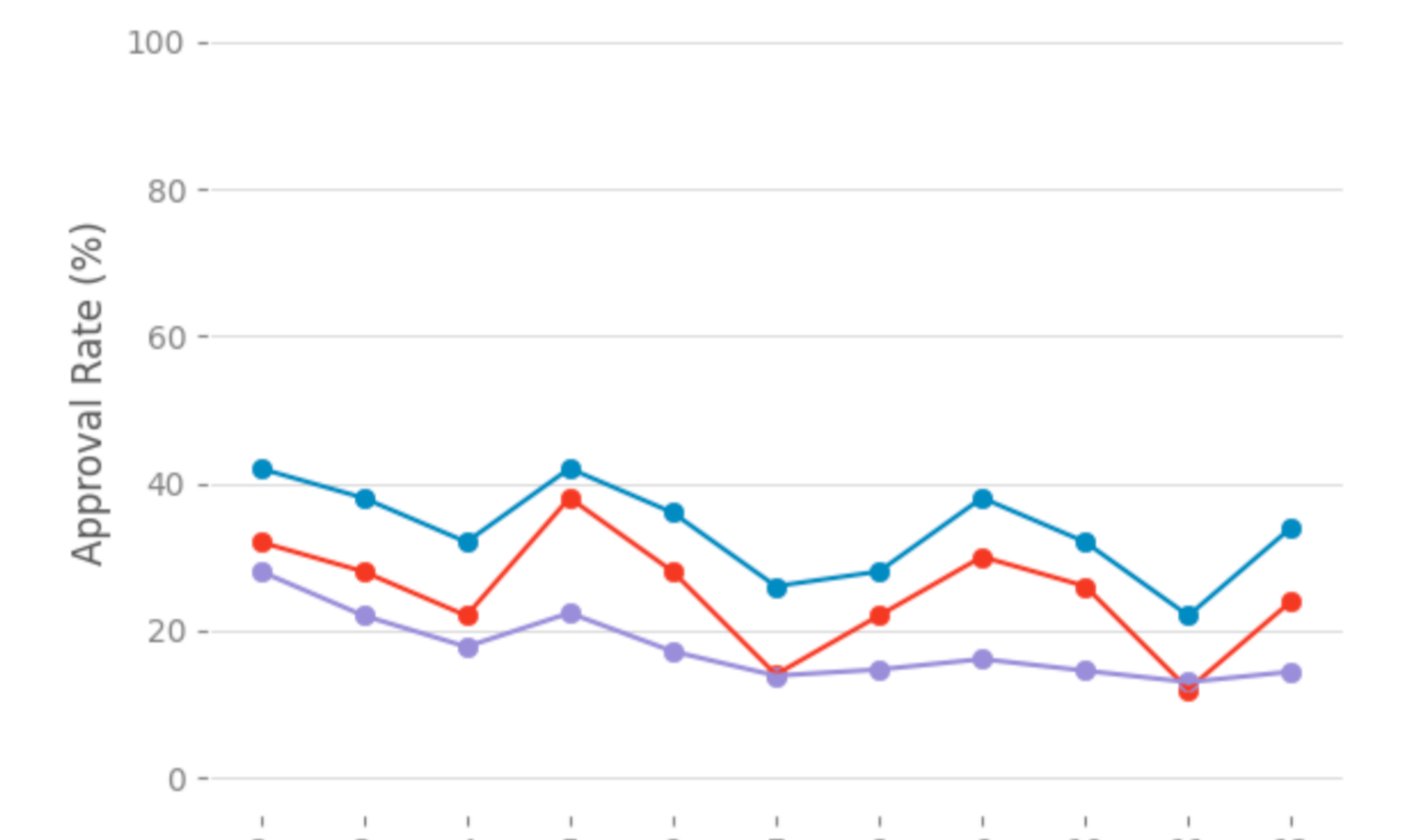}} & 
    \cell{c}{\includegraphics[trim={0cm 8cm 0cm 7cm}, clip, width=0.25\linewidth]{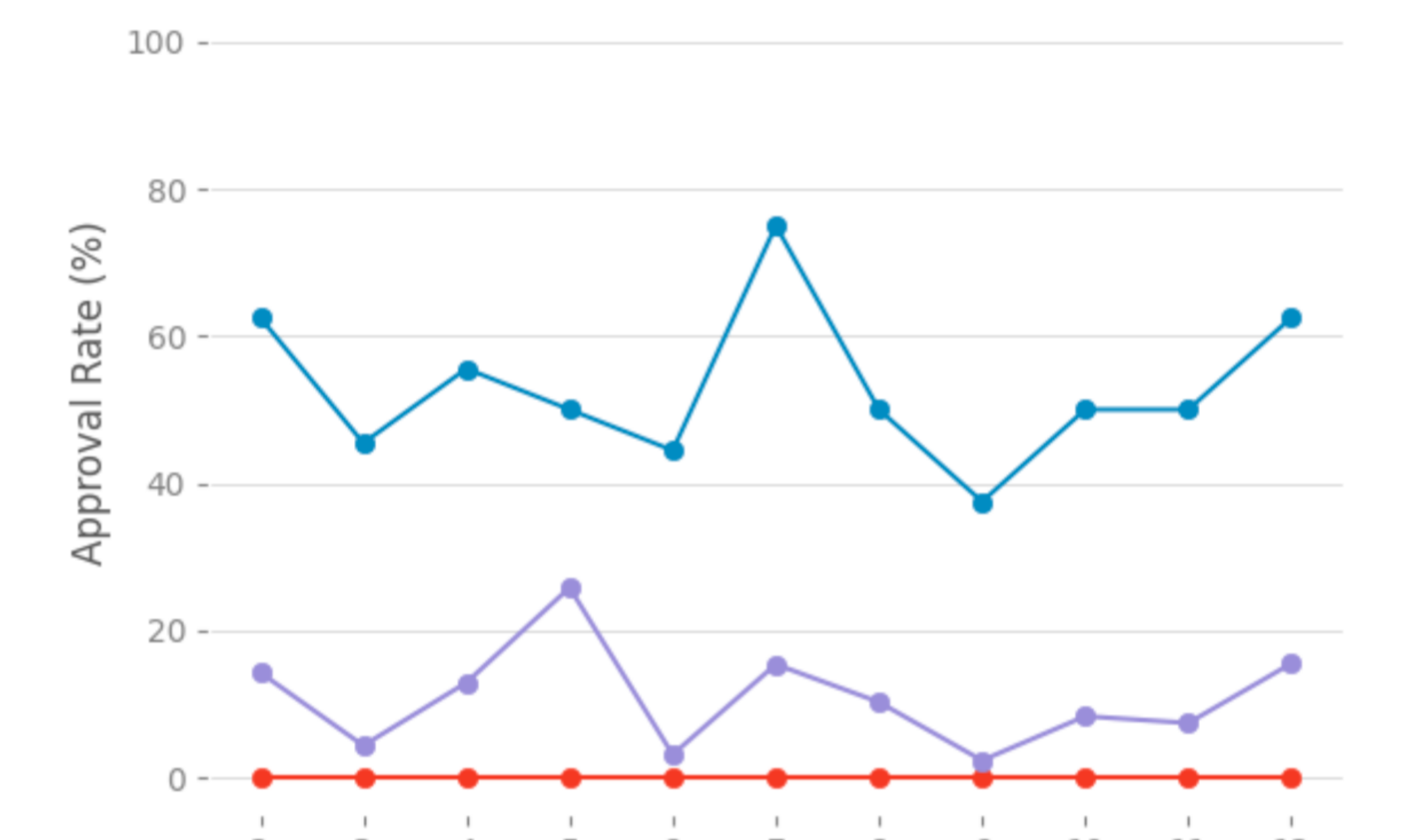}} \\ 
    \cell{c}{$y = 1$} &
    \cell{c}{\includegraphics[trim={0cm 6.5cm 0cm 7cm}, clip, width=0.25\linewidth]{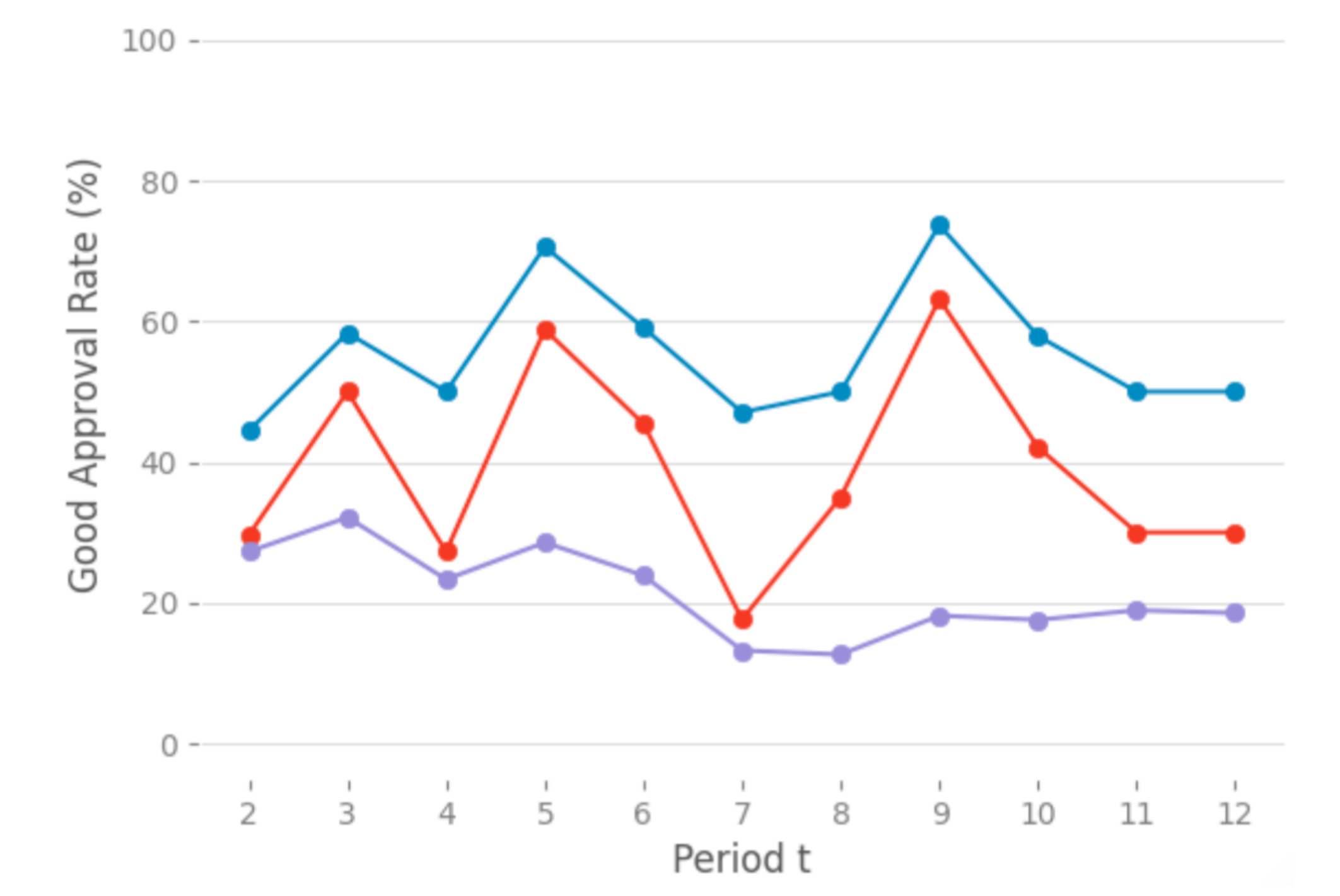}} & 
    \cell{c}{\includegraphics[trim={0cm 6.5cm 0cm 7cm}, clip, width=0.25\linewidth]{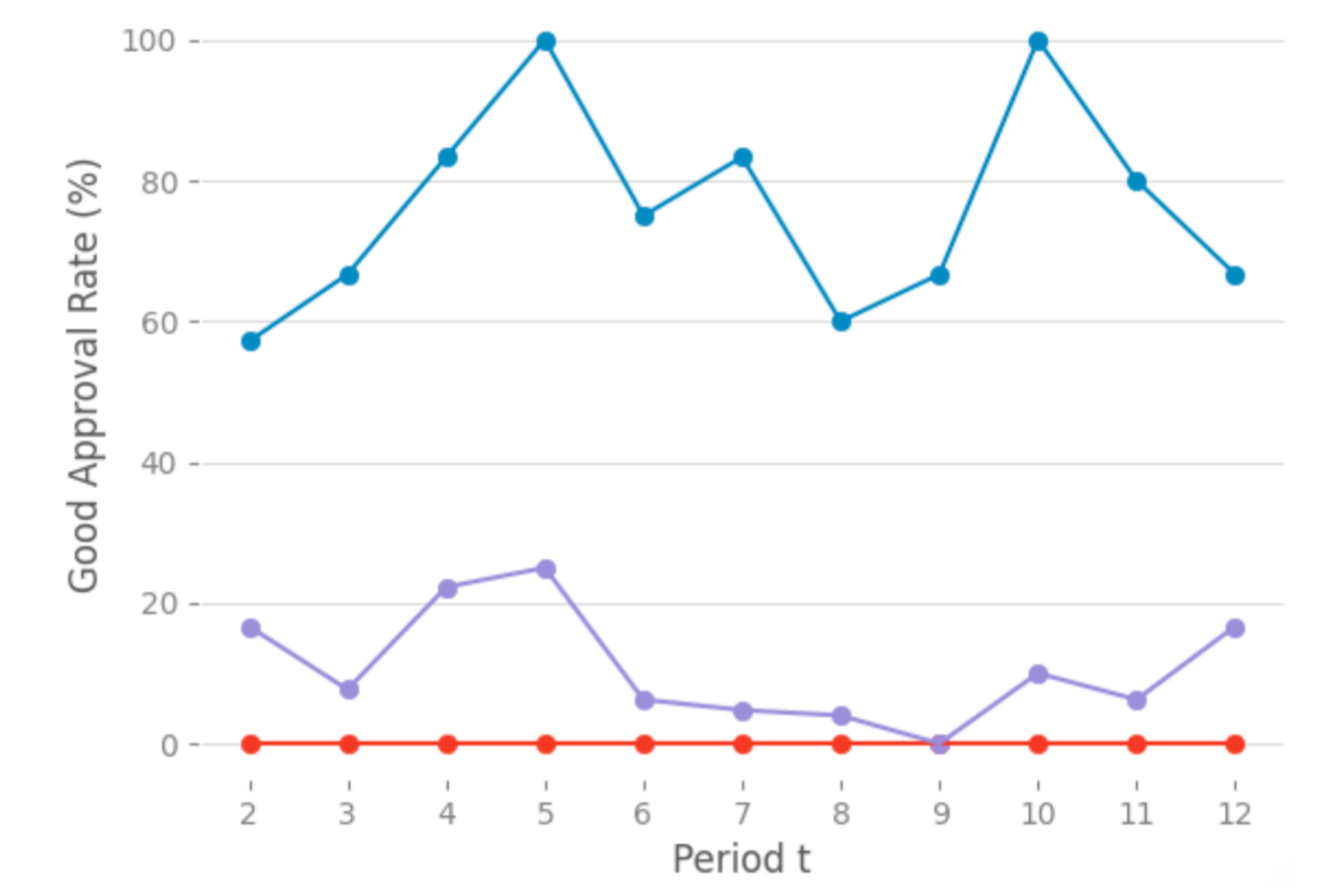}}
    \end{tabular}
    }
    \caption{Toy example where randomization solves censoring. We compare results of three systems under the \textds{causal} setup: a system without censoring (blue), a system with censoring (red), and a system with censoring that randomly approves some denied applicants (purple). The randomization allows censored applicants to enter into the training data and correct the model.}
    \label{Fig::ExplorationProofOfConcept}
\end{figure}

Formally, uniform randomization approves some $\alpha$-proportion of points assigned negative predictions to collect labels. Points assigned negative predictions are approved with equal probability: $$\prob{\xb_i \in \data{t+1} ~|~ \clfd{}{t}(\xb_i) = -1}  = \alpha.$$ \cref{Fig::ExplorationProofOfConcept} shows an example of how randomization can resolve censoring for a true positive subgroup. Without randomization, the $Z=1$ subgroup are perpetually assigned negative predictions throughout the system lifetime. Randomization approves a subset of points assigned $\clfd{}{t} = -1$, providing an opportunity for the model to learn that the $Z=1$ group is $\Y=1$. This helps protect against censoring when model-owners lack knowledge of exactly which consumer segment is being censored.

While this approach is individually fair (i.e., across experimental units), it may require many rounds of randomization to observe examples from a censored group, especially when the group consitutes a small part of the overall population. Therefore we also consider randomization with inverse propensity score weighting (IPW), which assigns a greater approval likelihood to points with lower predicted probabilities $\clf{}(\xb_i)$, as such points are the least likely to be assigned a positive prediction from noise alone or model drift due to retraining. Formally:
\begin{align*}
\prob{\xb_i \in \data{t+1} ~|~ \clfd{}{t}(\xb_i) = -1} = \frac{1}{\clfp{t}(\xb_i)} \sum_{\substack{\xb_j \in \data{t+1},\\\clfd{}{t}(\xb_i) = -1}} \clfp{t}{}(\xb_j)
\end{align*}

\paragraph{Scope}
Randomization may not be suitable for all applications due to ethical concerns. For instance, in lending, random exploration may approve loans for individuals likely to default. While individuals have agency over whether to accept the loan, being offered a loan might express confidence that individuals can repay it. This might make individuals more likely to accept the loan, resulting in negative consequences such as crippling their credit scores and preventing future loan qualification~\citep[][]{havard2005lend, squires2017fight}. IPW, in particular, may exacerbate these harms, as individuals with low-confidence predictions are more likely to be approved and thus will disproportionately bear the burdens of exploration.

\paragraph{Technical Challenges}
Randomization may be difficult to guarantee. In lending, for example, applicants may submit applications to various banks in addition to reducing their likelihood of getting a loan (in the case of IPW) to increase their chances of approval. In addition, randomization may also make models difficult to explain. In applications where consumers are entitled to information on why they were denied, stochastic classifiers are harder to justify to any one individual who was and was not randomly explored upon~\citep[][]{ECOA}.

\paragraph{Limitations}
Though randomization can resolve censoring, it can be costly, particularly when false negatives are a small proportion of all negative predictions. For instance, if the censored group is only 1\% of the denied population and uniform randomization is employed at $\alpha=0.01$. Then in expectation, the system will have to employ randomization for 100 periods before they collect a single data point from the censored group. However, the collection of this point alone will not resolve censoring, as the system will have to gather enough examples of those from the censored group to predict $\hat{y}=1$. So the system will likely have to employ randomization for much more than 100 periods before correcting on such a censored group. This is extremely time and resource costly for both model-owners and points.

Randomization also raises equity concerns in regards to differential approval rates for ``identical'' individuals -- i.e. those with the same features. The legality of such methods are context dependent. In lending, this may be perceived as a fair classifier, but in recidivism prediction, this may violate individuals' rights to equal treatment in the 14th amendment of the US Constitution~\citep[][]{fdic_2021, baunach1980random, colli2014ethical}.

\subsection{Recourse}
\begin{figure}[htbp]
    \centering
    \resizebox{0.7\linewidth}{!}{
    \begin{tabular}{rcc}
    \cell{l}{\\\bf Outcome} & \cell{c}{\textbf{Population}\\$z \in \{-1,1\}$} & \cell{c}{\bf{Censored Group}\\$z=1$}  \\ 
    \cmidrule(lr){1-1} \cmidrule(lr){2-2} \cmidrule(lr){3-3} \\
    \cell{c}{$y \in \{-1,1\}$} &
    \cell{c}{\includegraphics[trim={0cm 8cm 0cm 7cm}, clip, width=0.25\linewidth]{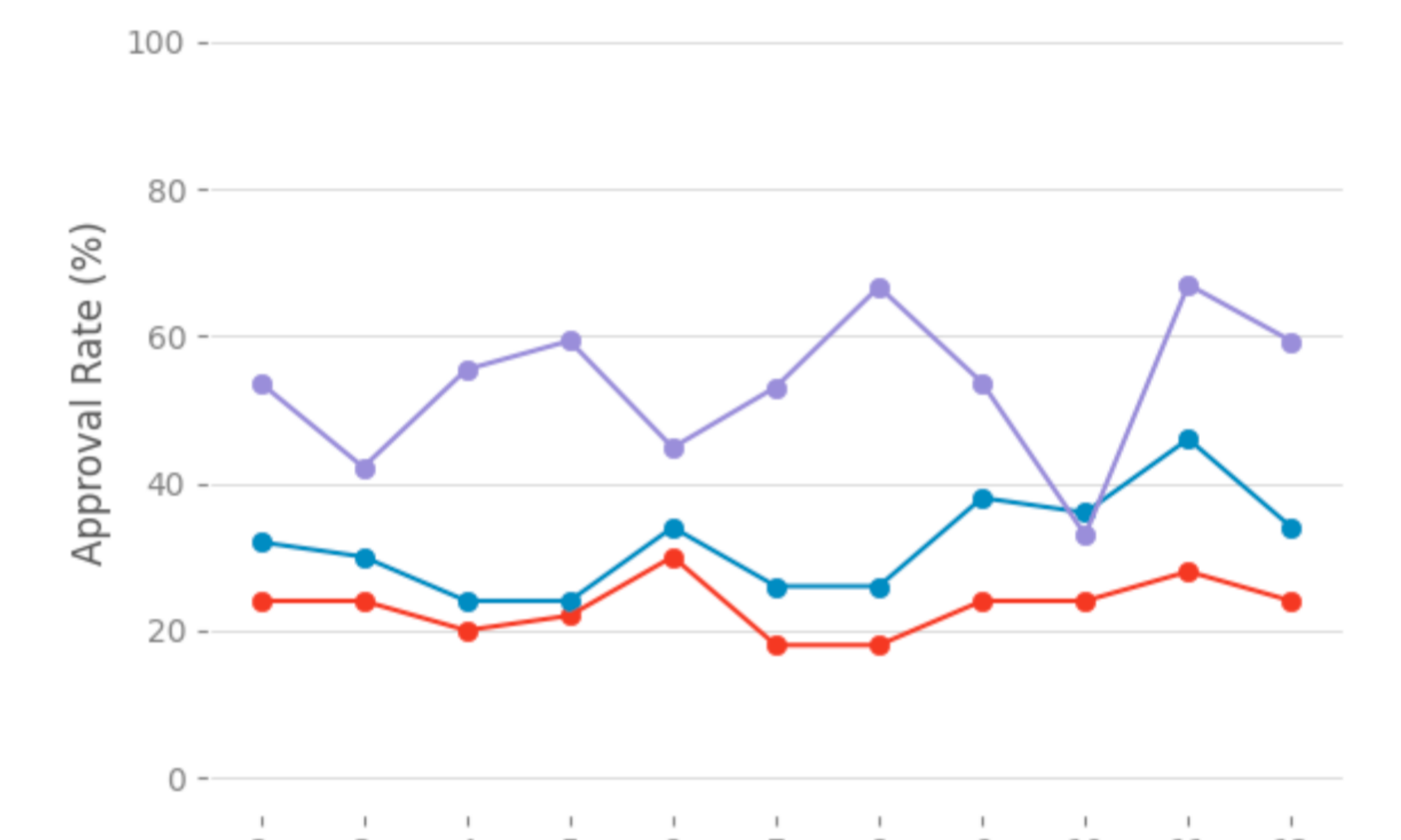}} & 
    \cell{c}{\includegraphics[trim={0cm 8cm 0cm 7cm}, clip, width=0.25\linewidth]{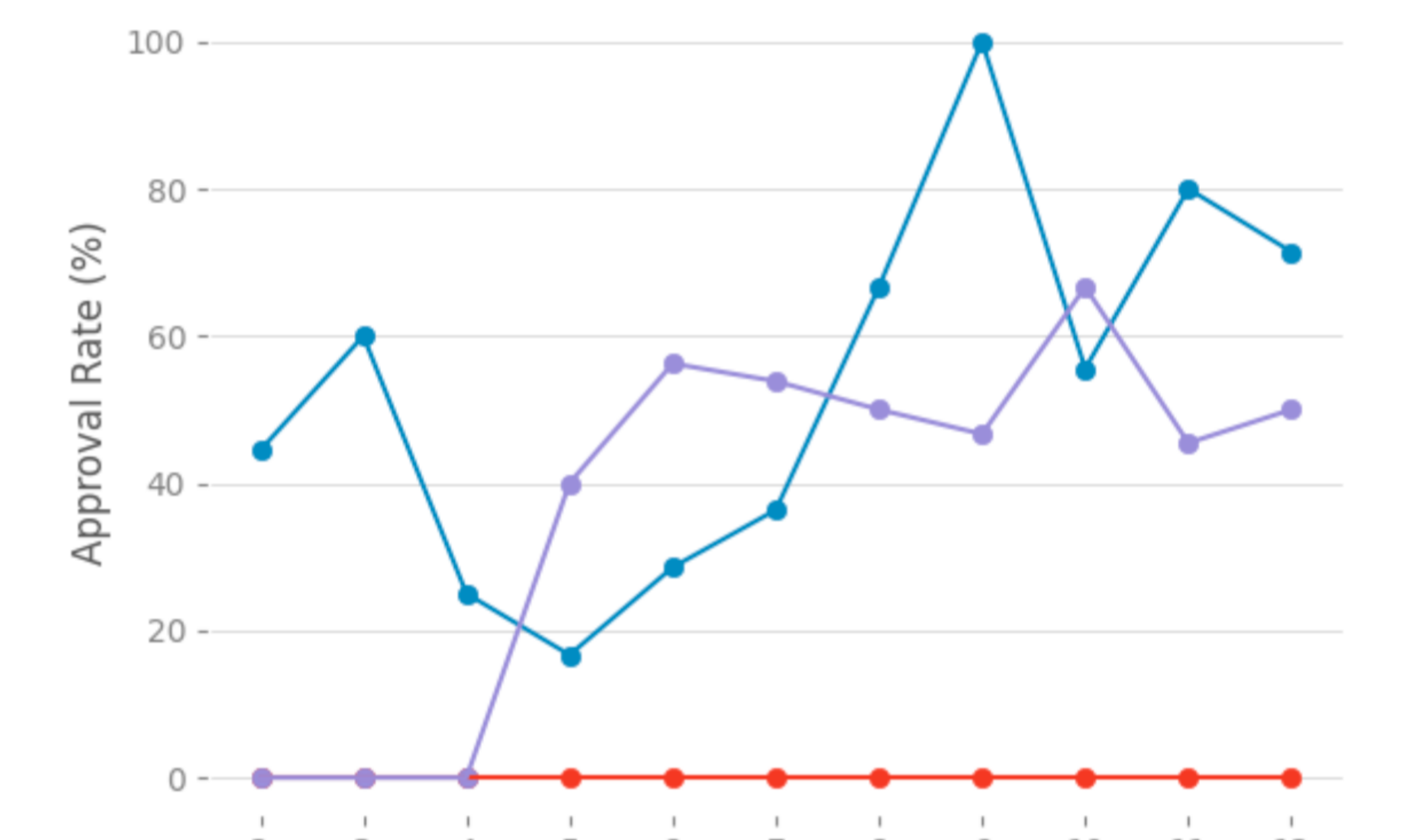}} \\ 
    \cell{c}{$y = 1$} &
    \cell{c}{\includegraphics[trim={0cm 6.5cm 0cm 7cm}, clip, width=0.25\linewidth]{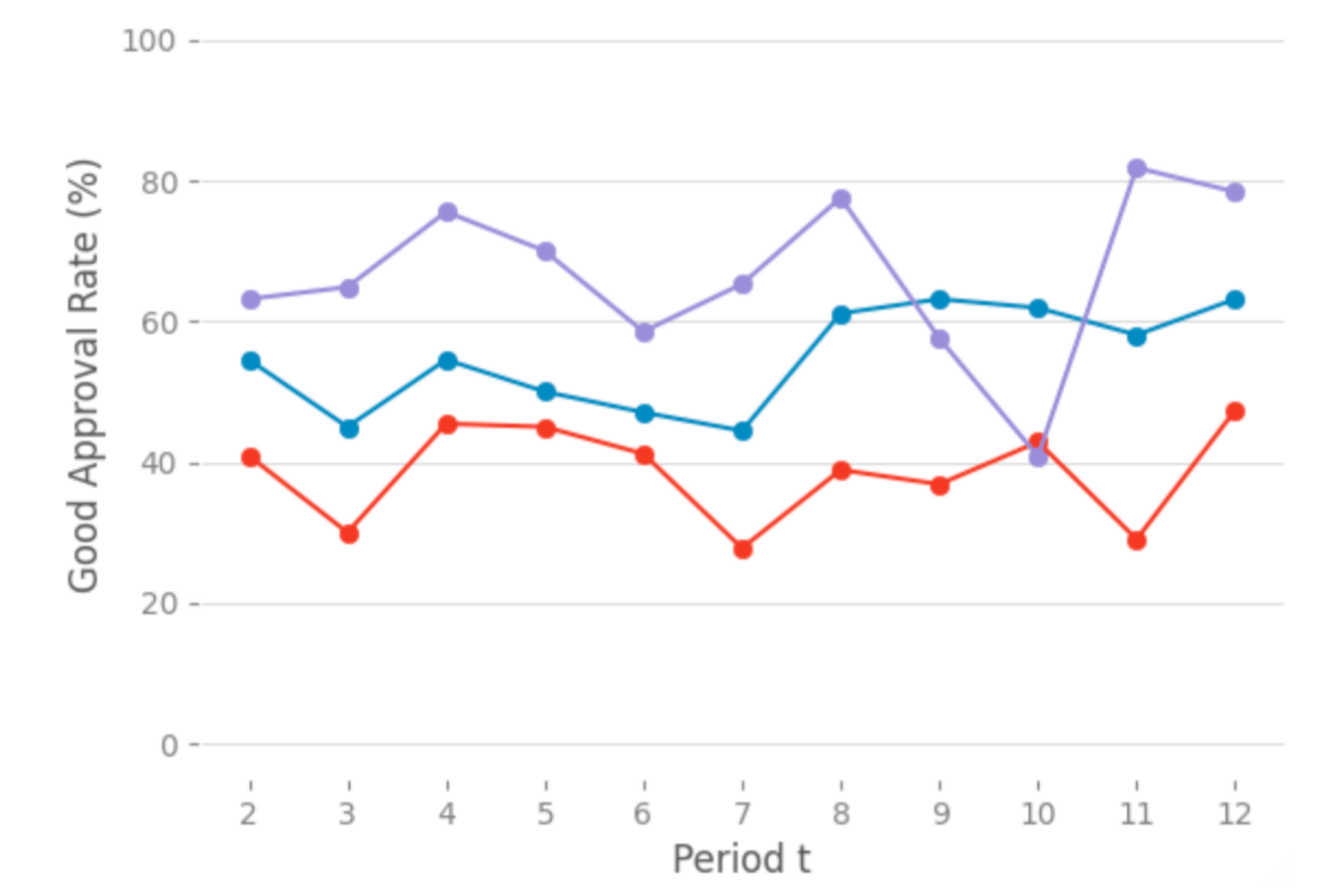}} & 
    \cell{c}{\includegraphics[trim={0cm 6.5cm 0cm 7cm}, clip, width=0.25\linewidth]{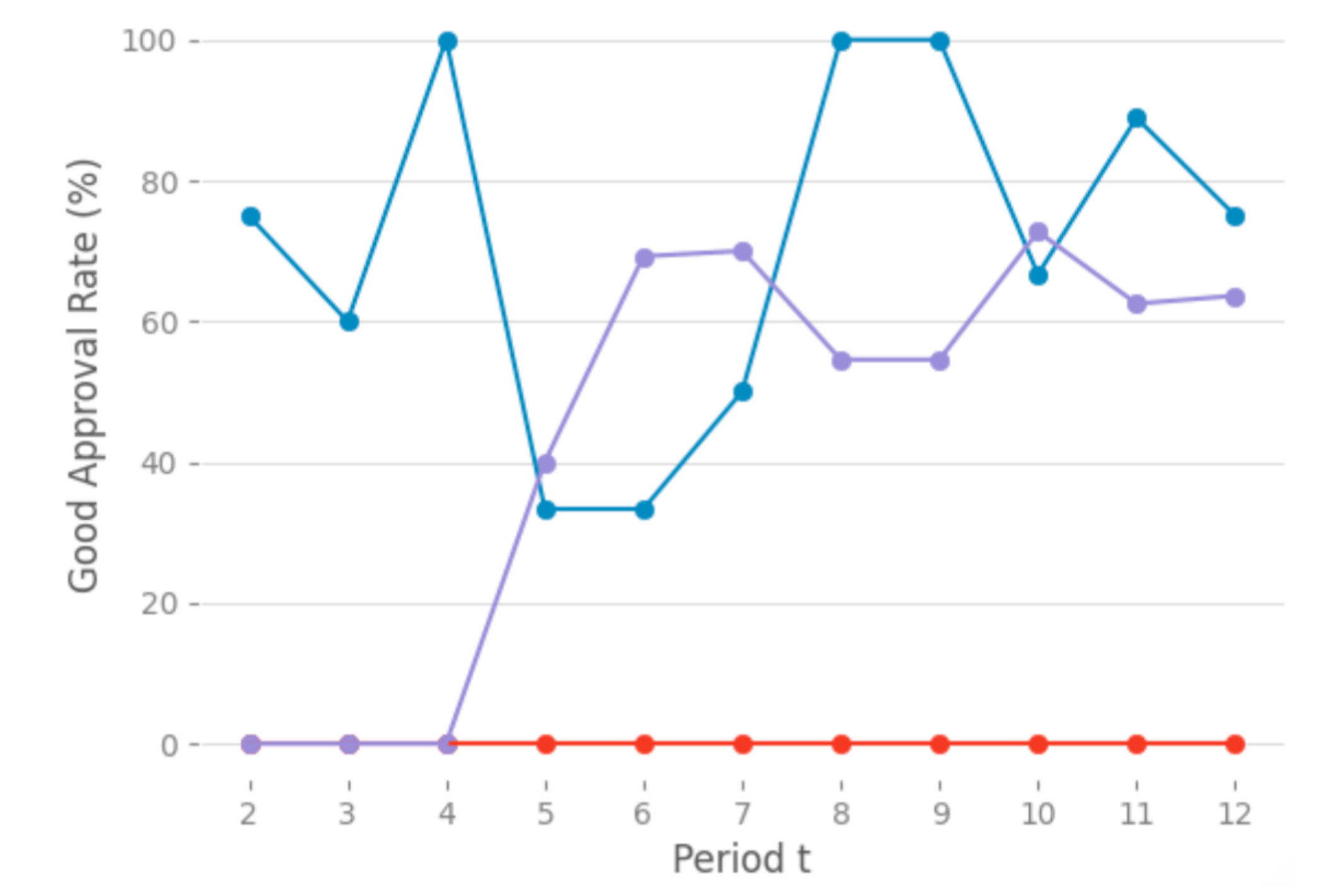}}
    \end{tabular}
    }
    \caption{Toy example where recourse resolves censoring. We compare the results of three systems under the \textds{causal} setup: a system without censoring (blue), a system with censoring (red), and a system with censoring that provides recourse to denied applicants (purple). The recourse allows censored applicants to enter into the training data and correct the model.}
    \label{Fig::RecourseProofOfConcept}
\end{figure}

An alternative safeguard against censoring is recourse, i.e., providing individualized feature changes that result in the desired prediction~\citep{ustun2019actionable}. Recourse resolves some of the ethical challenges posed by randomization because it gives experimental units agency over their predictions and provides a common set of rules. When applied to causal features, it can lead to improvement of the prediction pool, as actions increase the predicted and true probability of the outcome, making this exploratory method lower-risk. We demonstrate how recourse can mitigate censoring in Figure~\ref{Fig::RecourseProofOfConcept}.

\paragraph{Setup}
Formally, recourse provides points with \emph{actions}, i.e. feature changes, that when added to the original features, result in the desired predicted outcome. Actions that are real-world feasible and actionable can be solved via the following optimization:
\begin{align*}
\begin{split}
\min \quad & \textrm{cost}(\ab_i; \xb_i) \\
\st \quad & \clfd{}{}(\xb_i+\ab_i) = 1 \\
&\ab_i \in A(\xb_i)
\end{split}
\label{Eq::RecourseProblem}
\end{align*}
Here, we solve for a set of features changes $\ab_i$ such that when they are added to a point $i$'s original features $\xb_i$, result in the desired predicted outcome. Actions $\ab$ are restricted by real-world constraints such as directionality (i.e., can only increase education), mutability (i.e., savings balance), and feasibility/realism (i.e., a savings balance increase of \$500 not \$100,000).

\paragraph{Scope}
Providing recourse may be unethical in settings where you cannot guarantee its availability for the entire population. This may arise from recourse infeasibility, i.e. there are not available feature changes that result in the desired predicted outcome, stemming from immutable features or feature domain limitations. Recourse may additionally be unethical to deploy on non-causal features, as feature changes may be ineffective and unnecessarily costly.

\paragraph{Technical Challenges}
Recourse is not a stand-alone solution. It does not provide guarantees and requires cooperation between model-owners and users.
Specifically, recourse provides actions calculated under the \textit{current model}, not robust to model or distributional changes\citep[][]{upadhyay2021towards}. In addition, while causal recourse actions have the potential to provide improvement\citep[][]{miller2020strategic}, true causal model is rarely known, meaning non-causal recourse actions have the potential to enable gaming. As a result, we also explore recourse with guarantees in Section \ref{Sec::Experiments}.

\paragraph{Limitations}
Cost of recourse estimation is an open area of research. As such, costs may be difficult to estimate and/or heterogeneously distributed across the population. In addition, equivalent actions in the eyes of the model may vary in the cost across individuals. This raises additional questions on how many equivalent actions should be provided.

\section{Experiments}
\label{Sec::Experiments}

In this section, we evaluate the two proposed strategies to safeguard against censoring. For clarity, we use language specific to the lending application, referring to `points that receive a negative prediction' as `denied applicants', and `points that received a negative prediction and are subject to a new prediction in the next period' as `reapplicants'. However, we emphasize that censoring is salient to many dynamic learning system domains. 

\subsection{Setup}
\label{Sec::ExperimentalSetup}
We compare seven mitigation strategies against two baselines: \textmethod{None} and \textmethod{NoCensoring}. These baselines reflect systems where censoring does and does not occur, respectively. All remaining mitigation strategies are employed with systems initializes with censoring induced via sample selection bias defined in Section \ref{Sec::Causes}. We further detail each mitigation strategy in \cref{Table::TreatmentOverview}.

\begin{table}[h!]
\centering
\resizebox{\textwidth}{!}{%
\scriptsize
    \begin{tabular}{lHlp{0.5\textwidth}}
    \textheader{Name} &
    \textheader{Censoring} &
    \textheader{Label Collection Policy} &
    \textheader{Description} \\ \midrule
    
    \cell{l}{Censoring} & \cell{l}{Yes} & \cell{l}{$\clfd{}{\rho}(\xb_i) = 1$} & This system suffers from censoring and demonstrates the difficulty of detection of censoring. This system quantifies the potential harms censoring that would otherwise be immeasurable in real-world system and serves as a worst-case scenario baseline to evaluate the impact of other policies. \\
    \midrule
    \cell{l}{NoCensoring} & \cell{l}{No} & \cell{l}{$\clfd{}{\rho}(\xb_i) = 1$} & This system is not subject to censoring and represents the best-case scenario. This system serves as a baseline of optimal performance to evaluate the impact of other policies. \\ 
    \midrule
    \cell{l}{Random} & \cell{l}{Yes} & \cell{l}{$\clfd{}{\rho}(\xb_i) = 1 \cup \alpha|\clfd{}{\rho}(\xb_i) = -1|$} & We use a semi-logistic policy to collect labels from two groups: 1) those with the predicted probability of $Y=1$ above the operating threshold of 0.5, and 2) a random sample of 1\% of points below the operating threshold that would otherwise be unobserved. \\
    \midrule
    \cell{l}{IPW} & \cell{l}{Yes} & \cell{l}{$\clfd{}{\rho}(\xb_i) = 1 \cup \alpha|\clfd{}{\rho}(\xb_i) = -1|$} & We use a semi-logistic policy to collect labels from two groups: 1) those with the predicted probability of $Y=1$ above the operating threshold of 0.5, and 2) through inverse probability weighting, 1\% of otherwise unobserved true outcomes. This means we prioritize exploring on points farther from the decision boundary, as such points are very less likely to be observed through retraining alone. \\
    \midrule
    \cell{l}{Rec} & \cell{l}{Yes} & \cell{l}{$\clfd{t}{\rho}(\xb_i) = 1$} & We observe labels for points in two groups: 1) those above the operating threshold, and 2) those that have made feature changes according to previously deployed models that are above the operating threshold. We provide no guarantees of feature change robustness to future model retraining, meaning some points may undergo multiple feature changes before their true outcome is observed. \\
    \midrule
    \cell{l}{Rec\_Guarantee} & \cell{l}{Yes} & \cell{l}{$\clfd{t}{\rho}(\xb_i) = 1 \cup \clfd{t-1}{\rho}(\xb_i) = -1$} & We observe the labels for points from the current period with predicted probabilities above the operating threshold and unobserved points from the previous period after having undergone recourse generated feature changes. Since all points enact recourse actions before the following period, we only need to provide guarantees that account for one period (see \cref{Limitations}). This allows a special track for evaluation that ensures that such points are guaranteed approval.
\end{tabular}}
\caption{Overview of baselines and mitigation strategies. We consider systems where we train a standard logistic regression model with $\ell_2$ regularization to the training dataset aggregated over time. We assign positive predictions when the predicted probability exceeds 50\%. Strategies that employ randomization use an exploration rate of $\alpha=0.01$. $t$ is used to denote the current period.}
\label{Table::TreatmentOverview}
\end{table}

\paragraph{Data Generating Processes}

We evaluate each strategy on eight \emph{data-generating processes} (DGPs) shown in \cref{Table::DAGs}. Each DGP is specified by a \emph{directional acyclic graph} (DAG), which defines the joint distribution of features and outcomes and allows us to evaluate the effect of interventions on the outcome $Y$. We consider different DGPs that vary in causality and complexity to study the effects of censoring and evaluate the effectiveness of mitigation strategies.

\paragraph{Simulation}

We initialize the system with a classifier $\clfd{1}{}$ that exhibits censoring by training on a sample selection biased data. We define the censored group as all $z=1$ individuals and drop all positive examples -- i.e., $\{ i~|~(z_i, y_i) = (1, 1)\}$. We simulate each system for $T = 12$ periods. At period $t$, we fit a logistic regression model with $\ell_2$ regularization $\clfd{t}{}$ using $\data{t}$. We then use $\clfd{t}{}$ to assign predictions to $\nnew{t}=100$ new applicants, and $n_\textnormal{ret}^{t}$ reapplicants from earlier periods. The system approves applicants assigned a predicted probability greater than $\rho = 0.5$.

In systems where we provide recourse (\textmethod{Rec}, \textmethod{Rec\_Guarantee}), we generate recourse actions to all denied applicants using the software of \citep[][]{ustun2019actionable}. We assume all applicants execute these actions before reapplying the following period. In systems with randomization(\textmethod{Random}, \textmethod{IPW}), we approve $\alpha=0.01$ denied applicants. We collect examples for $(\xb_i, y_i)$ for all approved applicants and add them to the training data for the following period $\data{t+1}$.

\paragraph{Evaluation}

We evaluate each mitigation strategy on a variety of metrics defined in \cref{Table::MetricOverview}. These metrics are designed to capture the interests of model-owners and decision subjects. The metrics demonstrate the challenges in detection -- i.e., by reporting the metrics for the observed and true (unbiased) populations, as well as for the censored and non-censored group ($z=1$). We run each simulation ten times and report summary statistics across replicates.

\subsection{Results}
\label{Sec::ExperimentalResults}
We summarize our results in \cref{Table::ExperimentOverview}. First, we demonstrate that censoring is not resolved through retraining alone and that it is difficult to detect. Second, we compare trade-offs among safeguards, varying across feature causality. Finally, we provide high-level recommendations across all settings, in particular for unknown causal settings.

\newcommand{\datacell}[4]{\cell{l}{\textds{#1}\\{#2}\\{#3}\\{#4}}}

\renewcommand{\metriclabels}[0]{\cell{r}{%
Observed AUC\\
True AUC\\
Gain/Loss\\%
\% Approved \\
\% Approved | $Z=1$\\
$\mathbb{E}$[Reapplications]\\
Net Improvement\\
}}

\begin{figure}[htbp]
\centering
\resizebox{\linewidth}{!}{
\begin{tabular}{lrccccccccc}
 & & 
 \multicolumn{2}{c}{\textsc{Baselines}} & 
\multicolumn{2}{c}{\textsc{Exploration}} & 
\multicolumn{2}{c}{\textsc{Recourse}} & 
\multicolumn{3}{c}{\textsc{Combined}} \\ 
\cmidrule(l{3pt}r{3pt}){3-4} 
\cmidrule(l{3pt}r{3pt}){5-6} 
\cmidrule(l{3pt}r{3pt}){7-8} 
\cmidrule(l{3pt}r{3pt}){9-11}

\textbf{Data Generating Process} & 
\cell{r}{\textbf{Metrics}} & 
\textsf{Censoring} & 
\textsf{NoCensoring} & 
\textsf{Random} & 
\textsf{IPW} & 
\textsf{Rec} & 
\textsf{Guarantee} & 
\cell{l}{\textsf{Rec} + \\\textsf{Random}} & 
\cell{l}{\textsf{Guarantee} + \\\textsf{IPW}} & 
\cell{l}{\textsf{Rec} + \\\textsf{IPW}}

\\ \toprule

\datacell{causal}{$y = x_1 + x_2 + z - 0.5$}{$\hat{y} = x_1 + x_2 + z + b$}{$\prob{Z=1} = 15\%$} & \metriclabels{} & 
\cell{r}{0.795\\0.671\\
1.0/1.0\\ 
24.2\%\\\cell{l}{\cellcolor{yellow}0.0\%}\\
-\\-} & 
\cell{r}{0.779\\0.777\\
1.3/1.0\\
33.8\%\\\cell{l}{\cellcolor{yellow}58.6\%}\\
-\\-} & 
\cell{r}{0.783\\0.738\\
1.2/1.2\\
18.6\%\\14.6\%\\
\cell{l}{\cellcolor{lime}3.59}\\-} & 
\cell{r}{0.79\\0.744\\
1.2/1.2\\
17.1\%\\\cell{l}{\cellcolor{lime}21.3\%}\\
3.31\\-} & 
\cell{r}{0.767\\0.727\\
1.6/1.3\\
49.5\%\\\cell{l}{\cellcolor{brown!65}41.6\%}\\
\cell{l}{\cellcolor{lime}1.21}\\\cell{l}{\cellcolor{lime}21.7\%}} & 
\cell{r}{0.766\\0.733\\
1.4/1.2\\
40.2\%\\38.2\%\\
1.00\\26.0\%\\} & 
\cell{r}{0.767\\0.738\\
1.6/1.3\\
54.0\%\\\cell{l}{\cellcolor{blue!25}53.0\%}\\
1.17\\22.1\%\\} &
\cell{r}{0.767\\0.736\\
1.4/1.2\\
40.0\%\\38.0\%\\
1.00\\25.8\%\\} & 
\cell{r}{0.767\\0.738\\
1.6/1.3\\
49.5\%\\48.7\%\\
1.24\\21.1\%\\} \\ 

\midrule

\datacell{causal\_blind}{$y = x_1 + x_2$}{$\hat{y} = x_1 + x_2 + z + b$}{$\prob{Z=1}=15\%$} & \metriclabels{} & 
\cell{r}{0.817\\0.729\\
1.0/1.0\\
29.6\%\\0.2\%\\
-\\-} & 
\cell{r}{0.789\\0.782\\
1.2/1.0\\
35.2\%\\33.5\%\\
-\\-} & 
\cell{r}{0.806\\0.757\\
1.1/1.2\\
20.3\%\\12.3\%\\
\cell{l}{\cellcolor{blue!25}3.56}\\-} & 
\cell{r}{0.816\\0.733\\
1.0/1.0\\
9.6\%\\1.2\%\\
\cell{l}{\cellcolor{blue!25}3.36}\\-} & 
\cell{r}{0.789\\0.767\\
1.5/1.3\\
51.9\%\\47.3\%\\
\cell{l}{\cellcolor{blue!25}1.21}\\23.1\%\\} &
\cell{r}{0.787\\0.763\\
1.3/1.2\\
40\%\\43.2\%\\
1.00\\28.0\%} & 
\cell{r}{0.789\\0.766\\
1.5/1.3\\
55.9\%\\51.2\%\\
\cell{l}{\cellcolor{blue!25}1.17}\\24.1\%\\} &
\cell{r}{0.787\\0.764\\
1.2/1.2\\
39.7\%\\42.8\%\\
1.00\\27.8\%\\} &
\cell{r}{0.789\\0.763\\
1.5/1.3\\
53.2\%\\48.6\%\\
\cell{l}{\cellcolor{blue!25}1.17}\\23.8\%} \\ 

\midrule

\datacell{causal\_linked}{$y = x_1 + x_2 + z - 0.5$}{$\hat{y} = x_1 + x_2 + z + b$}{$\prob{Z=1}=15\%$} & \metriclabels{} & 
\cell{r}{0.837\\0.766\\
1.0/1.0\\
30.8\%\\2.7\%\\
-\\-} & 
\cell{r}{0.826\\0.836\\
1.3/1.0\\
39.1\%\\56.7\%\\
-\\-} & 
\cell{r}{0.834\\0.809\\
1.1/1.2\\
21.3\%\\16.6\%\\
3.58\\-} & 
\cell{r}{0.841\\0.811\\
1.1/1.2\\
21.1\%\\22.6\%\\
3.36\\-} & 
\cell{r}{0.811\\0.780\\
1.2/1.2\\
44.8\%\\19.0\%\\
1.28\\17.9\%\\} &
\cell{r}{0.807\\0.771\\
1.0/1.1\\
28.6\%\\11.1\%\\
1.00\\22.7\%\\} &
\cell{r}{0.812\\0.801\\
1.3/1.2\\
49.6\%\\37.4\%\\
1.27\\18.2\%\\} &
\cell{r}{0.807\\0.772\\
1.0/1.1\\
28.5\%\\11.3\%\\
1.00\\22.7\%\\} &
\cell{r}{0.813\\0.778\\
1.2/1.2\\
44.7\%\\20.5\%\\
1.29\\17.6\%\\} \\ 
   
\midrule

\datacell{causal\_equal}{$y = x_1 + x_2 + z - 1.5$}{$\hat{y} = x_1 + x_2 + z + b$}{$\prob{Z=1}=15\%$} & \metriclabels{} & 
\cell{r}{0.792\\0.707\\
1.0/1.0\\
28.3\%\\0.0\%\\
-\\-} & 
\cell{r}{0.764\\0.774\\
1.2/1.0\\
33.9\%\\33.7\%\\
-\\-} & 
\cell{r}{0.783\\0.742\\
1.1/1.2\\
20\%\\12.1\%\\
3.63\\-} & 
\cell{r}{0.791\\0.712\\
1.0/1.0\\
9.1\%\\1.2\%\\
3.56\\-} & 
\cell{r}{0.761\\0.726\\
1.4/1.3\\
48.7\%\\21.3\%\\
1.21\\21.4\%\\} &
\cell{r}{0.756\\0.726\\
1.2/1.2\\
37.1\%\\20.5\%\\
1.00\\26.1\%\\} &
\cell{r}{0.768\\0.753\\
1.5/1.3\\
54.7\%\\43.3\%\\
1.17\\24.0\%\\} &
\cell{r}{0.76\\0.73\\
1.2/1.2\\
37.4\%\\24.7\%\\
1.00\\26.3\%\\} &
\cell{r}{0.763\\0.733\\
1.5/1.3\\
50.3\%\\32.7\%\\
1.19\\22.2\%\\} \\ 
   
\midrule

\datacell{mixed\_proxy}{$y = x_1 + x_2 + z - 0.5$}{$\hat{y} = x_1 + x_2^{non} + z + b$}{$\prob{Z=1}=15\%$} & \metriclabels{} & 
\cell{r}{\cell{l}{\cellcolor{pink}0.799}\\\cell{l}{\cellcolor{pink}0.679}\\
1.0/1.0\\
25.5\%\\\cell{l}{\cellcolor{blue!25}0.3\%}\\
-\\-} & 
\cell{r}{0.782\\0.776\\
1.3/1.0\\
35.1\%\\59.8\%\\
-\\-} & 
\cell{r}{0.786\\0.739\\
1.2/1.2\\
19.1\%\\\cell{l}{\cellcolor{orange}14.9\%}\\
440\\-} & 
\cell{l}{0.793\\0.746\\
1.2/1.2\\
17.8\%\\\cell{l}{\cellcolor{orange}22.9\%}\\
\cell{l}{\cellcolor{blue!25}3.27}\\-} & 
\cell{l}{0.767\\0.705\\
1.5/1.3\\
43.9\%\\35.5\%\\
\cell{l}{\cellcolor{blue!25}1.40}\\\cell{l}{\cellcolor{lime}12.4\%}\\} &
\cell{r}{0.762\\0.697\\
1.3/1.3\\
39.8\%\\32.5\%\\
1.00\\17.8\%\\} &
\cell{r}{0.77\\0.728\\
1.5/1.3
\\49\%\\\cell{l}{\cellcolor{blue!25}50.6\%}\\
\cell{l}{\cellcolor{blue!25}1.36}\\12.9\%\\} &
\cell{r}{0.762\\0.703\\
1.3/1.3\\
40\%\\34.8\%\\
1.00\\17.9\%\\} &
\cell{r}{0.768\\0.707\\
1.5/1.3\\
44.9\%\\42.9\%\\
\cell{l}{\cellcolor{blue!25}1.44}\\12.3\%\\} \\ 
   
\midrule

\datacell{mixed\_downstream}{$y = x_1 + z - 0.5$}{$\hat{y} = x_1 + x_2^\textnormal{non} + z + b$}{$\prob{Z=1}=15\%$} & \metriclabels{} & 
\cell{r}{0.813\\0.697\\
1.0/1.0\\
25.9\%\\0.1\%\\
-\\-} & 
\cell{r}{0.799\\0.802\\
1.3/1.0\\
35.5\%\\63.1\%\\
-\\-} & 
\cell{r}{0.802\\0.764\\
1.2/1.2\\
19.5\%\\16\%\\
3.54\\-\\} & 
\cell{r}{0.808\\0.771\\
1.1/1.2\\
18.3\%\\24.1\%\\
3.24\\-} & 
\cell{r}{0.783\\0.747\\
1.5/1.3\\
46\%\\44.5\%\\
1.38\\10.6\%\\} &
\cell{r}{0.775\\0.747\\
1.3/1.3\\
41.1\%\\43.2\%\\
1.00\\15.2\%\\} &
\cell{r}{0.783\\0.757\\
1.5/1.3\\
49.5\%\\54.3\%\\
1.39\\10.4\%\\} &
\cell{r}{0.775\\0.752\\
1.2/1.3\\
40.9\%\\42.7\%\\
1.00\\15.0\%\\} &
\cell{r}{0.785\\0.747\\
1.5/1.3\\
46.6\%\\45.8\%\\
1.39\\10.6\%\\} \\ 
   
\midrule

\datacell{gaming}{$y = x_1 + z - 0.5$}{$\hat{y} = x_1^\textnormal{non} + x_2^\textnormal{non} + z + b$}{$\prob{Z=1}=15\%$} & \metriclabels{} & 
\cell{r}{0.805\\0.697\\
1.0/1.0\\
25.2\%\\\cell{l}{\cellcolor{blue!25}0.4\%}\\
-\\-\\} & 
\cell{r}{0.794\\0.799\\
1.3/1.0\\
35.1\%\\65.5\%\\
-\\-\\} & 
\cell{r}{0.796\\0.763\\
1.2/1.2\\
19.3\%\\17.1\%\\
3.54\\-\\} & 
\cell{r}{0.803\\0.769\\
1.2/1.2\\
18.2\%\\24.5\%\\
3.35\\-\\} & 
\cell{r}{0.761\\0.707\\
1.3/1.3\\
39.1\%\\28.4\%\\
1.68\\3.2\%\\} &
\cell{r}{0.741\\0.703\\
1.2/1.3\\
40.7\%\\34.2\%\\
1.00\\5.0\%\\} &
\cell{r}{0.76\\0.732\\
1.4/1.4\\
45\%\\\cell{l}{\cellcolor{blue!25}45.3\%}\\\
1.53\\3.5\%\\} &
\cell{r}{0.741\\0.711\\
1.2/1.3\\
41.2\%\\39.1\%\\
1.00\\4.7\%\\} &
\cell{r}{0.759\\0.716\\
1.4/1.4\\
41.1\%\\36.8\%\\
1.60\\3.1\%\\} \\ 

\midrule

\datacell{german}{$y = 0.3(-l -d + i + s)$}{\cell{l}{$\hat{y} = g + a + e + l$\\$+ d + i + s + b$}}{} & \cell{r}{%
Observed AUC\\
True AUC\\
Gain/Loss\\%
\% Approved (all)\\
\% Approved | $G=1$\\
$\mathbb{E}$[Reapplications]\\
\% Net Improvement\\
}
&
\cell{r}{0.74\\0.572\\
1.0/1.0\\
16.8\%\\1.4\%\\
-\\-} 
& 
\cell{r}{0.704\\0.678\\
2.0/1.1\\
44.3\%\\48.7\%\\
-\\-\\} & 
\cell{r}{0.989\\0.986\\
7.0/6.7\\
28.1\%\\27.1\%\\
3.56\\-\\} & 
\cell{r}{0.988\\0.986\\
1.9/1.0\\
23.6\%\\22.6\%\\
2.71\\-\\} & 
\cell{r}{0.726\\0.563\\
1.2/1.1\\
21.8\%\\8.4\%\\
1.47\\-32.0\%\\} &
\cell{r}{0.722\\0.572\\
1.2/1.4\\
37.7\%\\36.6\%\\
1.00\\-45.0\%\\} &
\cell{r}{0.711\\0.577\\
1.2/1.2\\
27.3\%\\15.7\%\\
1.40\\-33.1\%\\} &
\cell{r}{0.725\\0.571\\
1.2/1.4\\
38.2\%\\37.2\%\\
1.00\\-44.0\%\\} &
\cell{r}{0.724\\0.561\\
1.2/1.2\\
22.3\%\\9.0\%\\
1.49\\-31.1\%\\}
\end{tabular}
}
\caption{Overview of censoring and mitigation strategies across data generating processes. We highlight results referenced in our remarks: {\color{yellow}Censoring Does Not Resolve Itself Without Intervention}, {\color{pink}On the Difficulties of Detection}, {\color{brown!65}All Mitigation Methods Recover from Censoring}, {\color{lime}Recourse Works Best in Causal Settings}, {\color{orange}IPW for Greater Censored Group Uncertainty}, and {\color{blue!25}Combination Methods: A Solution For Unknown Causal Settings}. These phenomena arise across DGPs. We use $b$ to specify the intercept.}
\label{Table::ExperimentOverview}
\end{figure}

\paragraph{Censoring Does Not Resolve Itself Without Intervention} 
Across all DGPs, using the no safeguard strategy (\textmethod{Censoring}) collects labels for $0\%$ of censored group points, compared to the optimal approval rate of $>58\%$ in \textmethod{NoCensoring}. This is also reflected in the net gain across all setups with \textmethod{NoCensoring}, particularly when compared to the baseline \textmethod{None}.
In addition, all systems exhibit censoring across all periods, indicating dynamic learning systems do not recover from retraining alone. 

\paragraph{On the Difficulties of Detection}
Censoring is difficult to detect using performance metrics alone. Across all DGPs with no safeguards (\textmethod{Censoring}), the observed AUC (only considering the points assigned positive predictions) is greater than the unbiased sample (true AUC). For instance, in the \textds{mixed\_proxy} DGP, the observed AUC is 0.799, while the true AUC is 0.679. Since model-owners only observe the labels for positively predicted points, they do not observe their performance on those assigned negative predictions. This inflates the AUC, providing model-owners a false estimate of performance. In addition, we provide uncertainty metrics: miscalibration area, sharpness score, and root mean squared adversarial group calibration error in \cref{Appendix::Uncertainty Measurement}, observing little differences between observed and true, further emphasizing the difficulty in detecting censoring.

\subsubsection{Trade-offs Between Mitigation Strategies}
Mitigation strategies impose different costs and trade-offs across all stakeholders. These also vary across strategy, degree of guarantee/robustness provided, cost of errors, feature causality, and ease of feature observability. Below we make high-level recommendations for navigating such trade-offs.

\paragraph{All Mitigation Methods Recover from Censoring}
Across all DGPs, all randomization, recourse, and combination methods recover from censoring. All strategies result in the approval of a non-zero approval rate for the censored group. For instance, \textmethod{Rec} approves 41.6\% on average while \textmethod{Censoring} approves 0\% of $z=1$ applicants in the \textds{causal} DGP. Each mitigation strategy provides an avenue for otherwise perpetually unobserved points to enter into the dataset. Once the censored group points are collected into the training dataset, the model is more likely to recover from censoring, even when undetected. This demonstrates how these safeguards serve as a self-correction mechanism, functioning without precursory knowledge of what defines the censored points.

\paragraph{Recourse Works Best in Causal Settings}
In settings with causal model features, recourse is less costly than randomization. For model-owners, the approval rate for $z=1$ points is greater in all recourse treatments than in randomization, indicating a faster rate of recovery. For instance, in \textds{causal} \textmethod{Rec}, we observe 41.6\% compared to 21.3\% approved in \textmethod{IPW}. In addition, the net improvement of recourse actions is greatest in models with more causal features: 21.7\% less likely to result in a negative outcome in \textmethod{Rec} compared to 12.4\% in \textmethod{mixed\_proxy}. This is consistent across all causal setups (i.e., \textds{causal}, \textds{causal\_blind}, \textds{causal\_linked}, \textds{causal\_equal}). Recourse allows for focused, directed exploration for otherwise unobserved points. When recourse can lead to improvement, applicants can change their features to increase both the predicted and actual likelihood of the positive outcome.

For censored points, the expected number of reapplications is lower in causal settings: 1.21 vs 3.59 reapplications in \textmethod{Rec} and \textmethod{Random} in \textds{causal}, meaning that any costs incurred from being reevaluated is smaller in recourse as compared with randomization. In addition, when no recourse guarantees are provided (\textmethod{Rec\_Guarantee}), the proportion of invalid recourse actions is lower in causal settings: 0.203 versus 0.359 in \textds{causal} and \textds{mixed\_proxy}, respectively. This suggests that in settings where the model features are known to be causal on the outcome, recourse is less costly than randomization. In recourse settings, \textmethod{Guarantee} also decreases the expected number of reapplications. While this is beneficial to censored points, it also incurs additional costs to model-owners. 

\begin{figure}[!t]
\centering
\includegraphics[trim={0 1.35cm 0 0cm}, clip, width=0.35\linewidth]{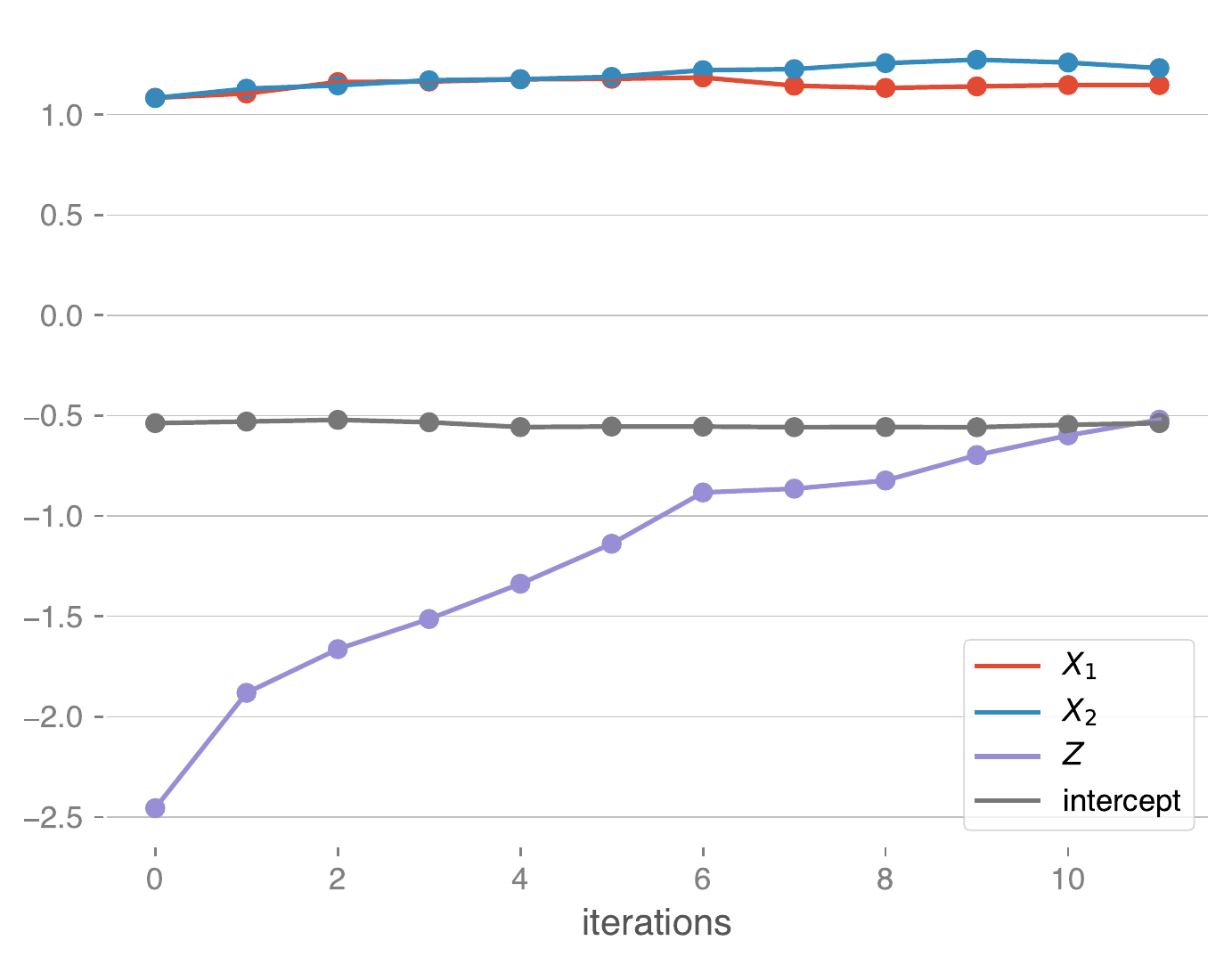}
\includegraphics[trim={0 1.35cm 0 0cm}, clip, width=0.35\linewidth]{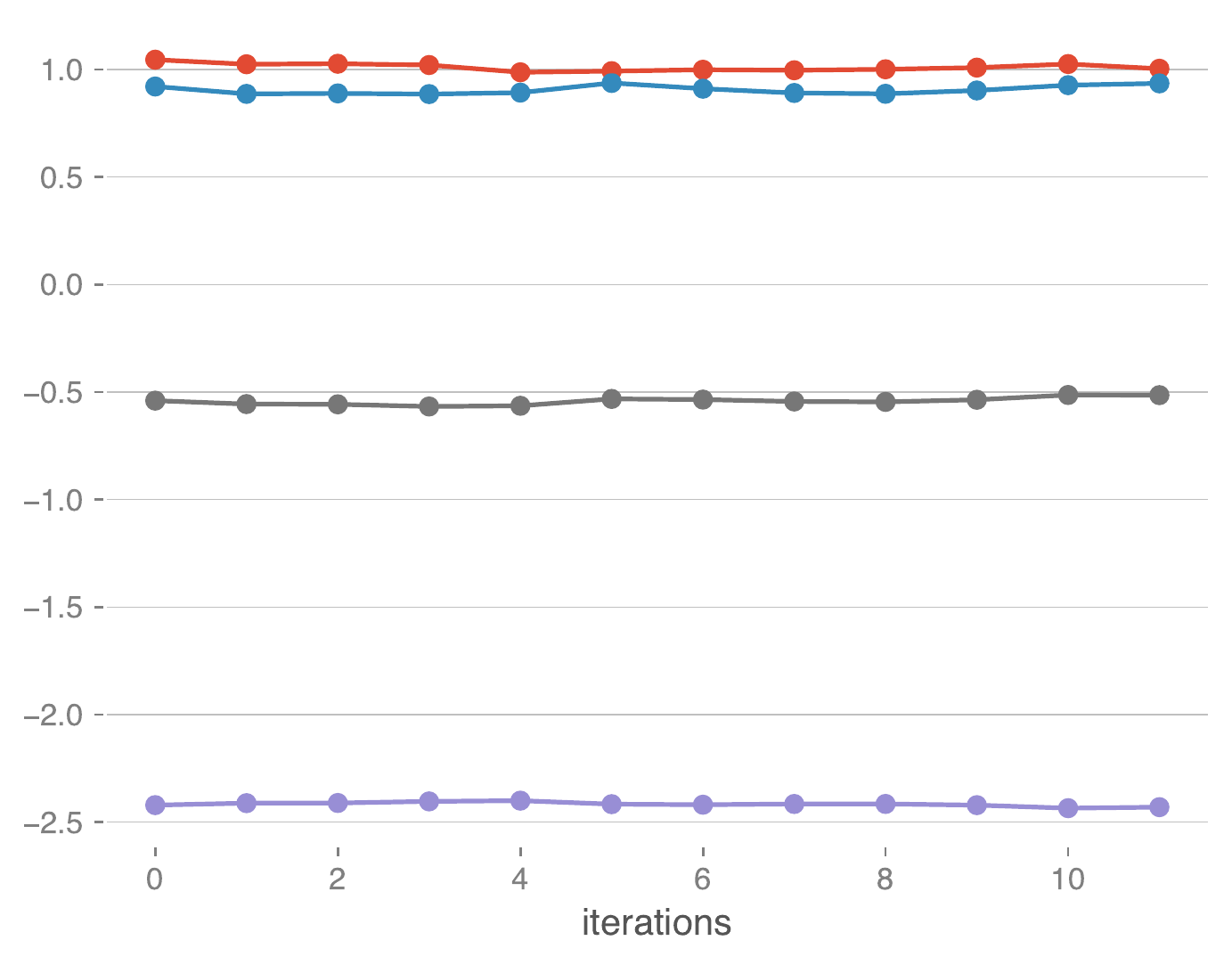} \\
\includegraphics[trim={0 0 0 3cm}, clip, width=0.35\linewidth]{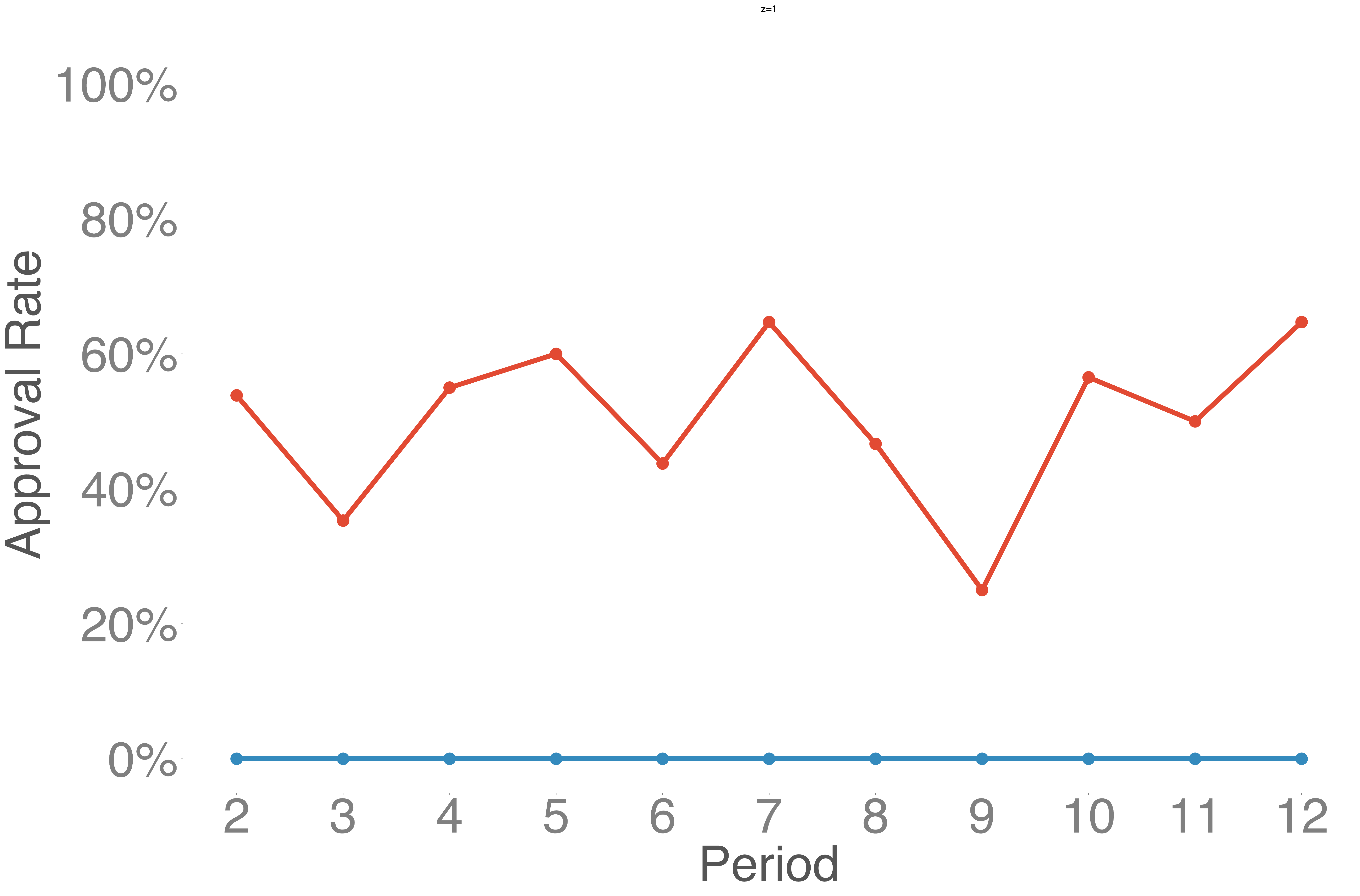}
\includegraphics[trim={0 0 0 3cm}, clip, width=0.35\linewidth]{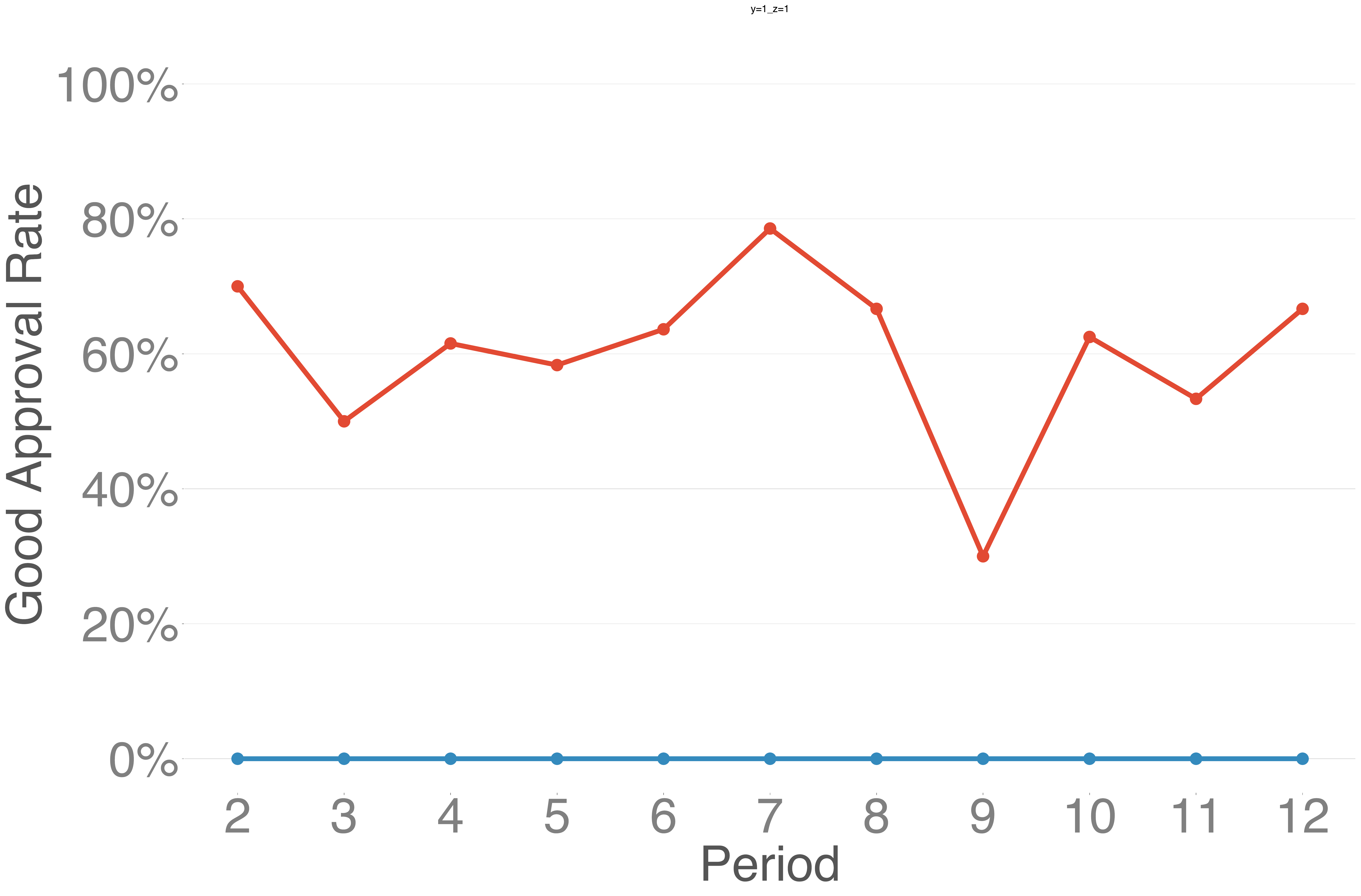}
\caption{\textbf{Recourse cannot solve censoring if it is infeasible.} Here, we consider a systems suffering from censoring under the \textds{Causal} setup. The leftmost plot shows the model coefficients in a setting where recourse is feasible, i.e. $X_1 \in [-2, 1.5]$ allows for recourse actions large enough to overcome the incorrect initial negative coefficient on $z$. This allows $z=1$ applicants to enter into the training dataset and correct the model, as shown by the $z$ coefficient increasing over iterations. In the second plot we show the model coefficients a setting where recourse is infeasible, i.e. $X_1 \in [-2, 1]$. Denied applicants cannot take recourse actions on $x_1$ large enough to be approved. Thus, the system continues suffering from censoring and the coefficient for $z$ stays negative. The third and fourth plots show the approval rates for the $z=1$ and the $z=1, y=1$ subgroups (the latter indicating the \% approved of those wrongfully censored) in of the feasible recourse system (red) and the infeasible recourse system (blue). We observe the approval rate for the censored group is $0\%$ in the infeasible setting and $>0$ for the feasible setting.}
\label{Fig::infeasibleRecourse}
\end{figure}

\paragraph{Failure Modes of Recourse: Infeasibility}
In fully causal settings, recourse may still fail when actions are too costly or outside of the feature range. This is called \emph{infeasible recourse}. This can arise when a model requires a feature change greater than that allowed for the feature to produce a positive prediction. These limitations can be specified by the training data or by real-world directional or bound limitations. We demonstrate how recourse fails to mitigate censoring in \cref{Fig::infeasibleRecourse} due to feature range limitations. Recourse infeasibility is important to consider as it may provide model-owners the illusion of a safeguard without providing the actual benefit.

\paragraph{IPW for Greater Censored Group Uncertainty}
Across all settings \textmethod{IPW} has a higher censored group approval rate, such as 14.9\% in \textmethod{Random}, compared to 22.9\% in \textmethod{IPW} for \textds{mixed\_proxy}.
This is due to the nature of how censoring is induced, i.e. by removing all $z=1, y=1$ examples from the training set. The resulting initial model then had a low $\hat{y}$ for such examples, meaning IPW could more effectively ensure that such examples were included in future training sets, rather than uniform selection across all negatively predicted samples. If instead the censored group was closer to the initial decision boundary (closer to the predicted probability of $\hat{p}=1)$ cutoff), then \textmethod{Random} would have been more effective. Thus, the efficacy of randomized exploration employed may still vary across the nature of the censoring. Settings with greater vulnerability to high inaccuracy censoring (larger disparities in error of predicted probabilities) may achieve more success with \textmethod{IPW}. \textmethod{Random}, however, may be a better catch-all alternative.

\paragraph{Combination Methods: A Solution For Unknown Causal Settings}
In settings with uncertainty about causality and recourse feasibility, model-owners may opt for combination methods, such as \textmethod{Rec+Random}. In both causal and non-causal settings, \textmethod{Rec+Random} has the highest \% Approved for the $z=1$ group. In the \textds{causal} setup, the system approves 53.0\% of censored group applications compared to 0\% in \textmethod{Censoring} and 58.6\% in \textmethod{NoCensoring}. \textds{mixed\_proxy} approves 50.6\%, compared to 0.4\% and 65.5\% of censored group applicants in \textmethod{Censoring} and \textmethod{NoCensoring}. \textds{Gaming}, approves 45.3\% in \textmethod{Rec+Random}, compared to 0.4\% and 65.5\% in \textmethod{None} and \textmethod{NoCensoring}. 
\textmethod{Rec+Random} also consistently has a lower expected number of reapplications than \textmethod{Random}, \textmethod{IPW}, and \textmethod{Rec}. In \textds{causal\_blind}, \textmethod{Rec+Random} has 1.27 expected reapplications, compared to 3.58 in \textmethod{Random}, 3.36 in \textmethod{IPW}, and  1.28 in \textmethod{Rec}. This phenomena still holds in the \textds{gaming} setup: with 1.53 reapplications in \textmethod{Rec+Random}, 3.54 in \textmethod{Random}, 3.35 in \textmethod{IPW}, and 1.68 in \textmethod{Rec}.

\section{Concluding Remarks}
\label{Sec::Discussion}

In this work, we formalize the censoring phenomenon, demonstrate several causes, highlight the difficulty in detection, and propose two families of safeguards for mitigation. We study the prevalence censoring under data generating processes that characterize salient applications -- those that identify potential effects that inflict harm and affect the effectiveness of mitigation. We define and compute metrics that quantify the harms to stakeholders often not considered in DLSs, such as decision-subjects and their respective cumulative harms. Our results show that recourse works best in causal settings, becoming more costly in mixed or non-causal settings. We propose combination methods for unknown causal settings, given that both randomization and recourse may be ethical in such settings.

Our experiments manually induce censoring. This enables us to quantify the harms and costs of censoring that go undetected and unmeasured in deployed systems. In reality, censored groups can be defined across any subset of features, making detection all the more difficult. In particular, in settings where we want to ensure equal access to the associated benefits of the positive predicted outcome, we emphasize the importance of deploying such mitigation methods. Censoring represents a potential harm in which individuals are denied consideration of a social system solely through algorithms alone. These harms should not only be considered, but prioritized.

\paragraph{Limitations} 
\label{Limitations}
We discuss key considerations, design decisions, and simplifying assumptions made in this work.

In our experiments, we assume that the time between periods is fixed and consistent (e.g., 2 years as is the norm in lending). In general, however, this is not always clearly defined. In drug discovery, for example, the next period may only begin after we have collected sufficient samples to re-train our model. In effect, such models are still subject to selective labeling, allowing for censoring to arise. However, variations in period length will impact the rate of recovery from censoring and the potential to resolve changes through recourse or other feature changes.

We recommend carefully considering the degree of model trust or uncertainty before deploying recourse, as it has varying degrees of efficacy. Recourse is particularly complex to operationalize as causal relationships are rarely known, may be heterogeneous across the population, or change with time. Additional barriers to feature selection and may include difficulty in measuring the causal variables, meaning proxy variables may be unavoidable in some cases. Recourse actions taken on proxy or non-causal variables may facilitate unintentional gaming, and significant downstream costs for both model-owners and experimental units. In addition, recourse without guarantees may lead to a loss in model trust, further reducing the likelihood of a recourse-action being taken. 

Finally, recourse does not guarantee that censoring will be fixed, whereas randomized sampling does in expectation. Unlike randomization, which, in the limit, will explore all points, recourse may not explore all negatively predicted domains. Thus, recourse may fail for groups suffering from infeasibility. In addition, because recourse creates incentives for feature change, it may fail to self-correct when recourse actions change the censored feature (or feature set). As a result, we provide results for experiments with recourse guarantees and combination methods as alternative mitigation strategies.

\section*{Acknowledgements}
This work was partially supported by National Science Foundation Grant \#1738411. We thank the following individuals for their thoughts and feedback: Yang Liu, Suresh Venkataburamanian, Jamelle Watson-Daniels, Nenad Tomasev, Michiel Bakker, Neil Mallinar, Avni Kothari, Hailey James, Fatemah Mireshghallah, Marcus Fedarko, and Mary Anne Smart.

\clearpage
\iftoggle{facct}{
\bibliographystyle{ext/ACM-Reference-Format}
}

\bibliography{recourse_dynamics}


\clearpage
\appendix
\onecolumn

\section{Additional Details on Experiments}
\label{Appendix::Experiments}

\subsection{Data Generating Techniques}
\label{Appendix::Data Generation}

In practice, machine learning models in dynamic learning system suffer from the selective labeling problem. This means that we observe at most one potential outcome for each data point, meaning that the remaining outcomes (such as the counterfactual had we made a different prediction) are unknown. The use of synthetic data enables us to generate such counterfactuals and generating metrics measuring the harms of a lack of intervention. We do so by fixing features in the post-recourse feature changes and re-sample downstream variables to generate a new true outcome while holding all upstream variables constant.

\subsection{Data Generating Processes}

In \cref{Table::DAGs}, we specify the DAGs used to simulate true outcomes varying across complexity and causality to the true and predicted outcome. Each node in the DAG is an instance of the TensorFlow Distributions Distribution class. The DAGs are specified as follows:

\begin{table}[htbp]
\centering
\resizebox{\textwidth}{!}{
\begin{tabular}{@{}lc>{\small}>{\scriptsize}l>{\scriptsize}l}
\cell{c}{\textheader{Name}} & 
\cell{c}{\textheader{DAG}} & 
\cell{c}{\textheader{DGP}} &
\cell{c}{\textheader{Description}} \\
\toprule

\textds{causal} &
\cell{c}{\includegraphics[trim={2.2cm 6cm 15.9cm 4cm}, clip, width=0.25\linewidth]{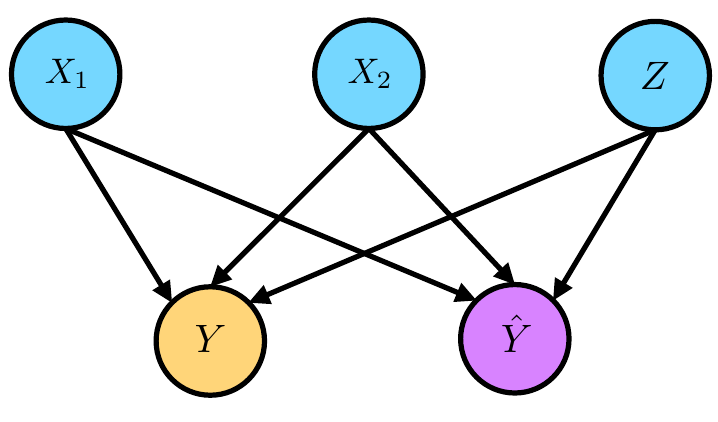}} &
\cell{l}{$X_1 \sim \textnormal{Normal}(0, 1)\in [-2, 1.5]$\\
$X_2 \sim \textnormal{Normal}(0, 1)\in [-2, 1.5]$\
$Z \sim \textnormal{Bernoulli}(0.15)$\\
$Y \sim \textnormal{Bernoulli}(\sigma(X_1 + X_2 + Z -0.5))$} &
\cell{p{1.5in}}{All features are causal and observed}\\ 

\midrule
\textds{causal\_blind} & 
\cell{c}{\includegraphics[trim={2.2cm 6cm 15.9cm 4cm}, clip, width=0.25\linewidth]{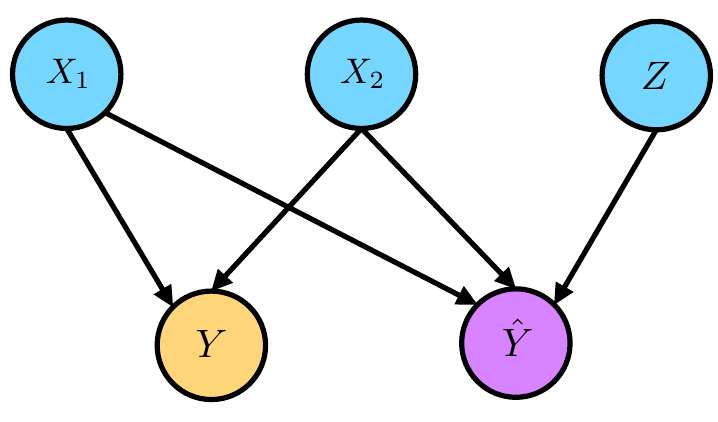}} &
\cell{l}{$X_1 \sim \textnormal{Normal}(0, 1)\in [-2, 2]$\\
$X_2 \sim \textnormal{Normal}(0, 1)\in [-2, 2]$\\
$Z \sim \textnormal{Bernoulli}(0.15)$\\
$Y \sim \textnormal{Bernoulli}(\sigma(X_1 + X_2 -0.5))$}  &
\cell{p{1.5in}}{$Z$ is not causal on $Y$, but is input into the model}\\ 

\midrule
\textds{causal\_linked} &
\cell{c}{\includegraphics[trim={2.1cm 6cm 15.9cm 2cm}, clip, width=0.25\linewidth]{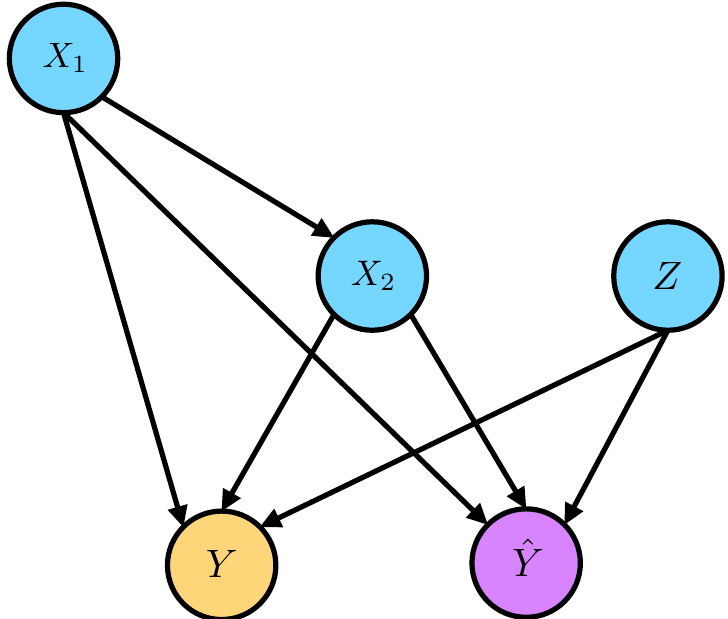}} &
\cell{l}{$X_1 \sim \textnormal{Normal}(0, 1)\in [-2, 1.5]$\\
$X_2 \sim \textnormal{ }X_1$\\
$Z \sim \textnormal{Bernoulli}(0.15)$\\
$Y \sim \textnormal{Bernoulli}(\sigma(X_1 + X_2 + Z -0.5))$} &
\cell{p{1.5in}}{$X_1$ is causal on $X_2$ so that actions on $X_1$ have downstream effects on $X_2$}\\ 

\midrule
\textds{causal\_equal} &
\cell{c}{\includegraphics[trim={2.1cm 6cm 15.9cm 1.9cm}, clip, width=0.25\linewidth]{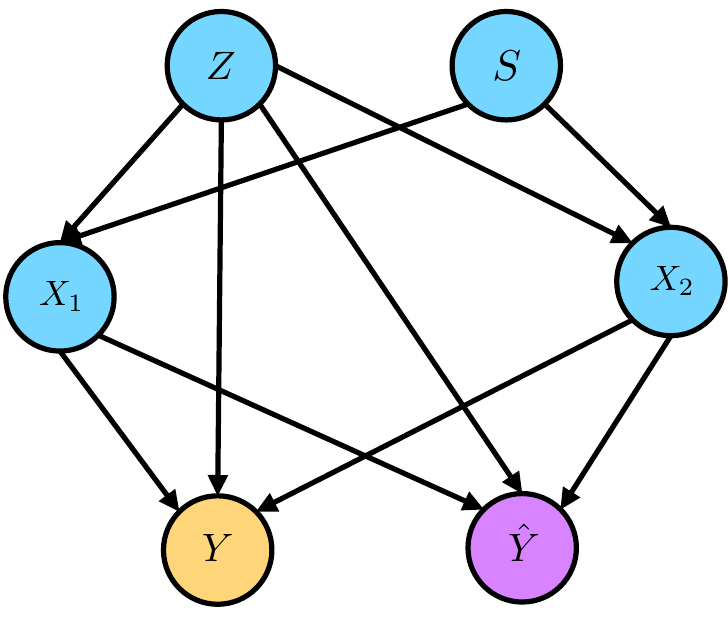}} &
\cell{l}{$X_1 \sim \textnormal{Normal}(\frac{2-z-s}{2}, 1)\in [-2, 2]$\\
$X_2 \sim \textnormal{Normal}(\frac{2-z-s}{2}, 1)\in [-2, 2]$\\
$Z \sim \textnormal{Bernoulli}(0.15)$\\
$S \sim \textnormal{Bernoulli}(0.75)$\\
$Y \sim \textnormal{Bernoulli}(\sigma(X_1 + X_2 + Z -1.5))$} &
\cell{p{1.5in}}{$\prob{Y=1~|~Z=1}$\\\hspace{0.5cm}$= \prob{Y=1~|~Z=0}$}\\

\midrule
\textds{mixed\_proxy} &
\cell{c}{\includegraphics[trim={2.1cm 6cm 15.9cm 1.9cm}, clip, width=0.25\linewidth]{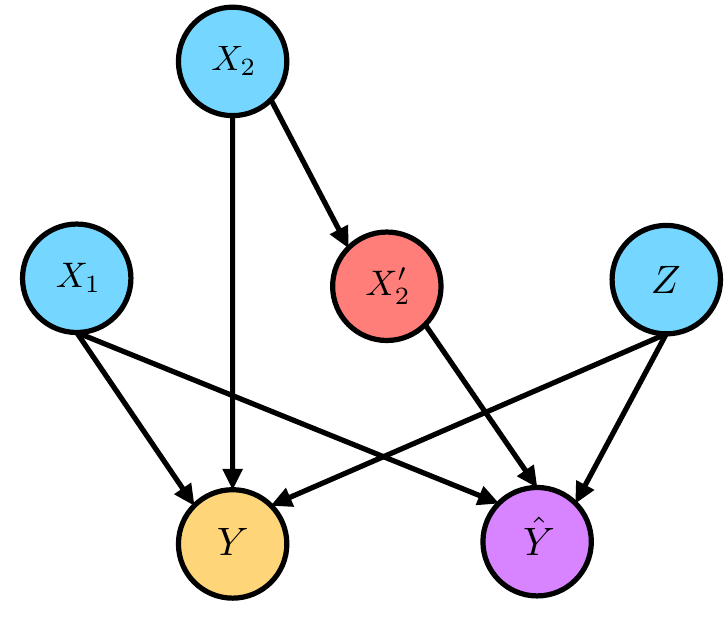}} &
\cell{l}{$X_1 \sim \textnormal{Normal}(0, 1)\in [-2, 1.5]$\\
$X_2 \sim \textnormal{Normal}(0, 1)\in [-2, 1.5]$\\
$X_2' \sim \textnormal{ }X_2$\\
$Z \sim \textnormal{Bernoulli}(0.15)$\\
$Y \sim \textnormal{Bernoulli}(\sigma(X_1 + X_2 + Z -0.5))$} &
\cell{p{1.5in}}{$X_2$ is causal on $Y$ but unobserved. Instead, non-causal variable $X_2'$ is observed and input into model}\\

\midrule
\textds{mixed\_downstream} &
\cell{c}{\includegraphics[trim={2.1cm 4cm 15.9cm 4cm}, clip, width=0.25\linewidth]{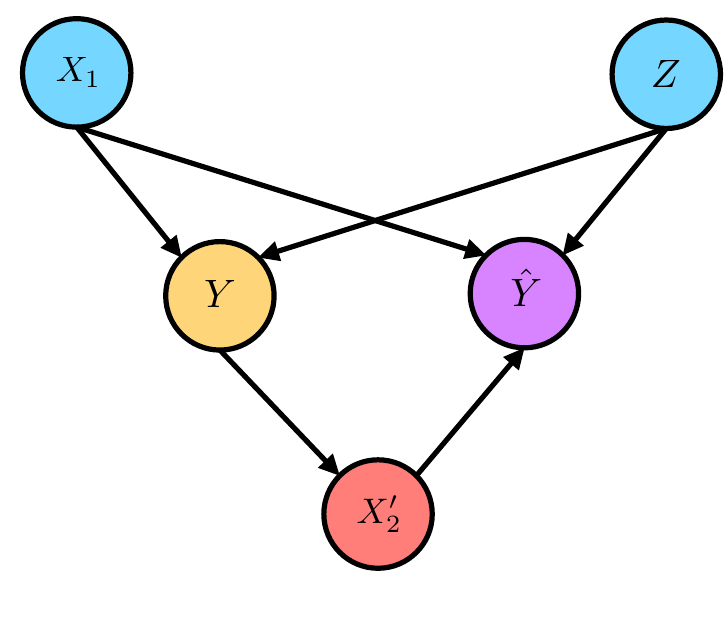}} &
\cell{l}{$X_1 \sim \textnormal{Normal}(0, 1)\in [-2, 1.5]$\\
$X_2' \sim \textnormal{Normal}(y-1, 1)\in [-2, 1.5]$\\
$Z \sim \textnormal{Bernoulli}(0.15)$\\
$Y \sim \textnormal{Bernoulli}(\sigma(X_1 + Z -0.5))$} &
\cell{p{1.5in}}{$X_2'$ is non-causal and downstream of $Y$, but is observed and input into the model}\\

\midrule
\textds{gaming} &
\cell{c}{\includegraphics[trim={0 6cm 15.9cm 1.9cm}, clip, width=0.25\linewidth]{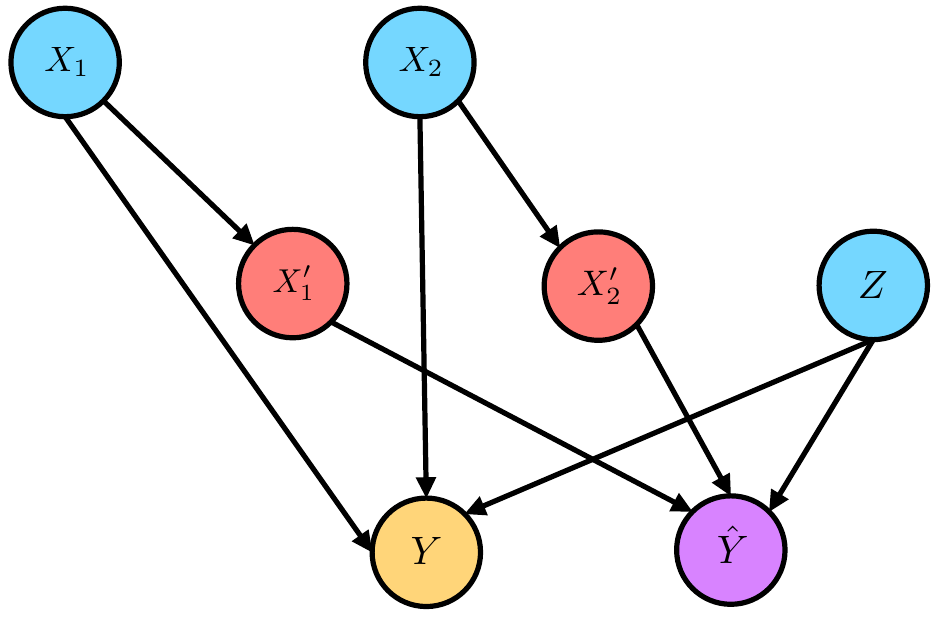}} &
\cell{l}{$X_1 \sim \textnormal{Normal}(0, 1)\in [-2, 1.5]$\\
$X_2 \sim \textnormal{Normal}(0, 1)\in [-2, 1.5]$\\
$X_1' \sim \textnormal{ }X_1$\\
$X_2' \sim \textnormal{ }X_2$\\
$Z \sim \textnormal{Bernoulli}(0.15)$\\
$Y \sim \textnormal{Bernoulli}(\sigma(X_1 + X_2 + Z -0.5))$} &
\cell{p{1.5in}}{$X_1$ and $X_2$ are causal on $Y$. Instead, $X_1'$ and $X_2'$ are observed and non-causally related to $Y$}\\

\midrule
\textds{german} & 
\cell{c}{\includegraphics[trim={2.1cm 6cm 15.3cm 3.5cm}, clip, width=0.25\linewidth]{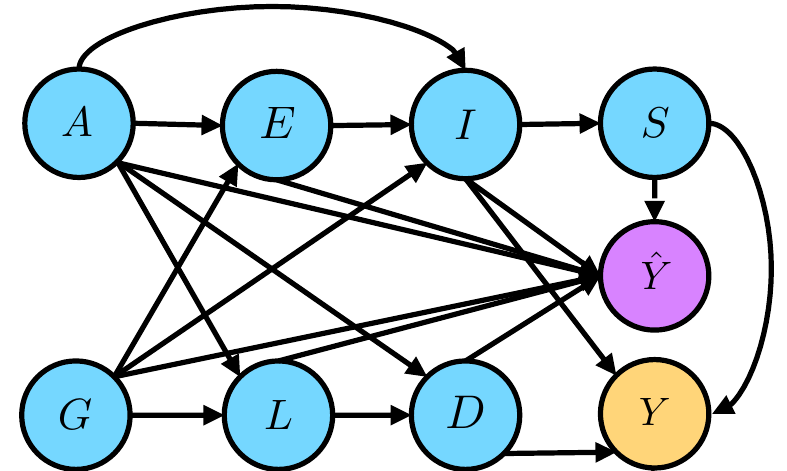}} &
\cell{l}{%
$A \sim -35 + \textnormal{Gamma}(10, \frac{1}{3.5})$\\
$Z \sim \textnormal{Bernoulli}(0.5)$\\
$E \sim -0.5 + \sigma(-1 + 0.5Z + \sigma(-0.1A)$\\
$\hspace{0.5cm}+ \textnormal{Normal}(0, 0.25))$\\
$L \sim 1 + 0.01(A-5)(5-A) + Z + \textnormal{Normal}(0, 4)$\\
$D \sim -1 + 0.01A + 2Z + L + \textnormal{Normal}(0, 9)$\\
$I \sim -4 +0.01(A+35) + 2Z + 2E + \textnormal{Normal}(0. 4)$\\
$S \sim -4 + \mathbbm{1}(I>0)(1.5I) + \textnormal{Normal}(0, 25)$\\
}
&
\cell{p{1.5in}}{Semi-synthetic data generating model adapted from \citet{karimi2020algorithmic}. DAG has more than two layers of nodes}\\
\bottomrule
\end{tabular}
}
\caption{Overview of Data Generating Processes. We provide the name, DAG, distributions, and the motivation behind each DGP. We color the nodes according to causality on the feature outcome, using blue for features that are directly causal on the true outcome, $Y$, colored in yellow. Red nodes, also denoted with a ' are not directly causal on $Y$ but are assumed to be by the model-owners, therefore used as a (known or unknown) proxy for the true causal variable. Arrows pointing into $Y$ denotes variables causally related. The predicted outcome variable $\hat{Y}$ is colored in purple; arrows pointing from a node into $\hat{Y}$ indicate which features are in the model.}
\label{Table::DAGs}
\end{table}

\clearpage




\section{Metrics}
\label{Appendix::Metrics}
Below, we provide the definitions of the metrics reported in Table \ref{Table::ExperimentOverview}.
\begin{table}[h!]
\centering
\resizebox{\textwidth}{!}{
\scriptsize
    \begin{tabular}{llp{0.5\textwidth}}
    \textheader{Metric} & \textheader{Definition} & \textheader{Description} \\ \midrule
    \cell{l}{AUC} & 
    $\int_{0}^{1}\frac{TPR}{FPR}$& 
    The average area under the receiver operating characteristic curve (AUC) for each period over all replicates
    \\ 
    \cell{l}{Gain} & 
    $\frac{\sum_{i \in \data{T}} \indic{y_i, 1}}{1000C_g||\data{T}||}$& 
    The total number of true positive individuals (x$10^3$) cumulatively approved through the last period averaged over all replicates. For comparison, we rescale to gain $C_g$ for each true positive individual. For each DGP, we normalize by $C_g$, the \textmethod{None} baseline gain for ease of comparison across mitigation methods.\\ 
    \cell{l}{Loss} & 
    $\frac{\sum_{i \in \data{T}} \indic{y_i, 0}}{1000C_l||\data{T}||}$& 
    The total number of true negative individuals (x$10^3$) approved cumulatively through the last period averaged over all replicates. For comparison, we rescale to gain $C_l$ for each true positive individual. For each DGP, we normalize by $C_l$, the inverse of the \textmethod{None} baseline loss for ease of comparison across mitigation methods.\\ 
    \cell{l}{\% Approved} & 
     $\frac{\sum_{i \in n_{eval}}\indic{\clfd{}{t}(\xb_i), 1}}{n_{eval}^{t}}$ & 
    The average \% of applicants approved per period over all replicates\\ 
    \cell{l}{$\mathbb{E}$[\# of Reapplications]} & 
    $\frac{\sum_{i \in \data{T} \cap \{\textnormal{count of reapplications}_i > 1\}} \textnormal{count of reapplications}_i}{||i \in \data{T} \cap \{\textnormal{count of reapplications}_i > 1\}||}$ & 
    The expected number of times an approved applicant needed to reapply before approval.
    \\ 
    \cell{l}{\# Never Approved} & 
    $n_{eval}^{T}$ 
    & \# of applicants who are still denied at period $T$.\\ 
    \cell{l}{Net Improvement} & 
    $\frac{\sum_{i \in n_{ret}^{t}}^{} \clfp{}(\xb_i+\ab_i) - \clfp{}(\xb_i)}{n_{ret}^{t}}$ & 
    The average difference in the likelihood of the desired true outcome $P(y=1)$ before and after a denied applicant takes a recourse action. This measures the cost of gaming in mixed causal settings.\\ 
    \cell{l}{Proportion Invalid Recourse} & 
    $\frac{\sum_{i \in n_{eval}}{\indic{\clfd{}{t}(\xb_i + \ab_i), 0}}}{n_{eval}}$ & 
    The average number of post-recourse action applicants that were still denied due to model shift during retraining. \\
    %
\end{tabular}}
\caption{Metrics used to compare systems in Table \ref{Table::ExperimentOverview}.}
\label{Table::MetricOverview}
\end{table}

\section{Full Results Table}
\label{Appendix::Full Results Table}

\renewcommand{\metriclabels}[0]{\cell{r}{%
Observed AUC\\
True AUC\\
Gain/Loss\\%
\% Approved (all)\\
\% Approved ($Z=1$)\\
$\mathbb{E}$[Reapplications]\\
\# Never Approved\\
\% Net Improvement\\
\% Invalid Recourse 
}}

\begin{figure}[htbp]
\centering
\resizebox{\linewidth}{!}{
\begin{tabular}{lrccccccccc}
 & & 
 \multicolumn{2}{c}{\textsc{Baselines}} & 
\multicolumn{2}{c}{\textsc{Exploration}} & 
\multicolumn{2}{c}{\textsc{Recourse}} & 
\multicolumn{3}{c}{\textsc{Combined}} \\ 
\cmidrule(l{3pt}r{3pt}){3-4} 
\cmidrule(l{3pt}r{3pt}){5-6} 
\cmidrule(l{3pt}r{3pt}){7-8} 
\cmidrule(l{3pt}r{3pt}){9-11}

\textbf{Data Generating Process} & 
\cell{r}{\textbf{Metrics}} & 
\textsf{None} & 
\textsf{NoCensoring} & 
\textsf{Random} & 
\textsf{IPW} & 
\textsf{Rec} & 
\textsf{Guarantee} & 
\cell{l}{\textsf{Rec} + \\\textsf{Random}} & 
\cell{l}{\textsf{Guarantee} + \\\textsf{IPW}} & 
\cell{l}{\textsf{Rec} + \\\textsf{IPW}}

\\ \toprule

\datacell{causal}{$y = x_1 + x_2 + z - 0.5$}{$\hat{y} = x_1 + x_2 + z + b$}{$\prob{Z=1} = 0.15$} & \metriclabels{} & 
\cell{r}{0.795\\0.671\\
1.0/1.0\\ 
24.2\%\\\cell{l}{\cellcolor{yellow}0.0\%}\\
-\\-\\-\\-} & 
\cell{r}{0.779\\0.777\\
1.3/1.0\\
33.8\%\\\cell{l}{\cellcolor{yellow}58.6\%}\\
-\\-\\-\\-} & 
\cell{r}{0.783\\0.738\\
1.2/1.2\\
18.6\%\\14.6\%\\
\cell{l}{\cellcolor{lime}3.59}\\447\\-\\-} & 
\cell{r}{0.79\\0.744\\
1.2/1.2\\
17.1\%\\\cell{l}{\cellcolor{lime}21.3\%}\\
3.31\\526\\-\\-} & 
\cell{r}{0.767\\0.727\\
1.6/1.3\\
49.5\%\\\cell{l}{\cellcolor{brown!65}41.6\%}\\
\cell{l}{\cellcolor{lime}1.21}\\79\\\cell{l}{\cellcolor{lime}21.7\%}\\\cell{l}{\cellcolor{lime}20.3\%}} & 
\cell{r}{0.766\\0.733\\
1.4/1.2\\
40.2\%\\38.2\%\\
1.00\\67\\26.0\%\\18.5\%} & 
\cell{r}{0.767\\0.738\\
1.6/1.3\\
54.0\%\\\cell{l}{\cellcolor{blue!25}53.0\%}\\
1.17\\76\\22.1\%\\19.2\%} &
\cell{r}{0.767\\0.736\\
1.4/1.2\\
40.0\%\\38.0\%\\
1.00\\66\\25.8\%\\18.2\%} & 
\cell{r}{0.767\\0.738\\
1.6/1.3\\
49.5\%\\48.7\%\\
1.24\\87\\21.1\%\\22.3\%} \\ 

\midrule

\datacell{causal\_blind}{$y = x_1 + x_2$}{$\hat{y} = x_1 + x_2 + z + b$}{$\prob{Z=1}=0.15$} & \metriclabels{} & 
\cell{r}{0.817\\0.729\\
1.0/1.0\\
29.6\%\\0.2\%\\
-\\-\\-\\-} & 
\cell{r}{0.789\\0.782\\
1.2/1.0\\
35.2\%\\33.5\%\\
-\\-\\-\\-} & 
\cell{r}{0.806\\0.757\\
1.1/1.2\\
20.3\%\\12.3\%\\
\cell{l}{\cellcolor{blue!25}3.56}\\428\\-\\-} & 
\cell{r}{0.816\\0.733\\
1.0/1.0\\
9.6\%\\1.2\%\\
\cell{l}{\cellcolor{blue!25}3.36}\\825\\-\\-} & 
\cell{r}{0.789\\0.767\\
1.5/1.3\\
51.9\%\\47.3\%\\
\cell{l}{\cellcolor{blue!25}1.21}\\80\\23.1\%\\19.5\%} &
\cell{r}{0.787\\0.763\\
1.3/1.2\\
40\%\\43.2\%\\
1.00\\66\\28.0\%\\15.3\%} & 
\cell{r}{0.789\\0.766\\
1.5/1.3\\
55.9\%\\51.2\%\\
\cell{l}{\cellcolor{blue!25}1.17}\\66\\24.1\%\\17.0\%} &
\cell{r}{0.787\\0.764\\
1.2/1.2\\
39.7\%\\42.8\%\\
1.00\\66\\27.8\%\\16.8\%} &
\cell{r}{0.789\\0.763\\
1.5/1.3\\
53.2\%\\48.6\%\\
\cell{l}{\cellcolor{blue!25}1.17}\\73\\23.8\%\\15.7\%} \\ 

\midrule

\datacell{causal\_linked}{$y = x_1 + x_2 + z - 0.5$}{$\hat{y} = x_1 + x_2 + z + b$}{$\prob{Z=1}=0.15$} & \metriclabels{} & 
\cell{r}{0.837\\0.766\\
1.0/1.0\\
30.8\%\\2.7\%\\
-\\-\\-\\-} & 
\cell{r}{0.826\\0.836\\
1.3/1.0\\
39.1\%\\56.7\%\\
-\\-\\-\\-} & 
\cell{r}{0.834\\0.809\\
1.1/1.2\\
21.3\%\\16.6\%\\
3.58\\405\\-\\-} & 
\cell{r}{0.841\\0.811\\
1.1/1.2\\
21.1\%\\22.6\%\\
3.36\\421\\-\\-} & 
\cell{r}{0.811\\0.780\\
1.2/1.2\\
44.8\%\\19.0\%\\
1.28\\47\\17.9\%\\24.1\%} &
\cell{r}{0.807\\0.771\\
1.0/1.1\\
28.6\%\\11.1\%\\
1.00\\40\\22.7\%\\30.0\%} &
\cell{r}{0.812\\0.801\\
1.3/1.2\\
49.6\%\\37.4\%\\
1.27\\48\\18.2\%\\24.9\%} &
\cell{r}{0.807\\0.772\\
1.0/1.1\\
28.5\%\\11.3\%\\
1.00\\40\\22.7\%\\29.7\%} &
\cell{r}{0.813\\0.778\\
1.2/1.2\\
44.7\%\\20.5\%\\
1.29\\55\\17.6\%\\25.9\%} \\ 
   
\midrule

\datacell{causal\_equal}{$y = x_1 + x_2 + z - 1.5$}{$\hat{y} = x_1 + x_2 + z + b$}{$\prob{Z=1}=0.15$} & \metriclabels{} & 
\cell{r}{0.792\\0.707\\
1.0/1.0\\
28.3\%\\0.0\%\\
-\\-\\-\\-} & 
\cell{r}{0.764\\0.774\\
1.2/1.0\\
33.9\%\\33.7\%\\
-\\-\\-\\-} & 
\cell{r}{0.783\\0.742\\
1.1/1.2\\
20\%\\12.1\%\\
3.63\\433\\-\\-} & 
\cell{r}{0.791\\0.712\\
1.0/1.0\\
9.1\%\\1.2\%\\
3.56\\839\\-\\-} & 
\cell{r}{0.761\\0.726\\
1.4/1.3\\
48.7\%\\21.3\%\\
1.21\\83\\21.4\%\\19.7\%} &
\cell{r}{0.756\\0.726\\
1.2/1.2\\
37.1\%\\20.5\%\\
1.00\\63\\26.1\%\\23.6\%} &
\cell{r}{0.768\\0.753\\
1.5/1.3\\
54.7\%\\43.3\%\\
1.17\\70\\24.0\%\\17.3\%} &
\cell{r}{0.76\\0.73\\
1.2/1.2\\
37.4\%\\24.7\%\\
1.00\\64\\26.3\%\\21.3\%} &
\cell{r}{0.763\\0.733\\
1.5/1.3\\
50.3\%\\32.7\%\\
1.19\\84\\22.2\%\\18.1\%} \\ 
   
\midrule

\datacell{mixed\_proxy}{$y = x_1 + x_2 + z - 0.5$}{$\hat{y} = x_1 + x_2^{non} + z + b$}{$\prob{Z=1}=0.15$} & \metriclabels{} & 
\cell{r}{\cell{l}{\cellcolor{pink}0.799}\\\cell{l}{\cellcolor{pink}0.679}\\
41.0/1.0\\
25.5\%\\\cell{l}{\cellcolor{blue!25}0.3\%}\\
-\\-\\-\\-} & 
\cell{r}{0.782\\0.776\\
1.3/1.0\\
35.1\%\\59.8\%\\
-\\-\\-\\-} & 
\cell{r}{0.786\\0.739\\
1.2/1.2\\
19.1\%\\\cell{l}{\cellcolor{orange}14.9\%}\\
\cell{l}{\cellcolor{blue!25}3.61}\\440\\-\\-} & 
\cell{l}{0.793\\0.746\\
1.2/1.2\\
17.8\%\\\cell{l}{\cellcolor{orange}22.9\%}\\
\cell{l}{\cellcolor{blue!25}3.27}\\511\\-\\-} & 
\cell{l}{0.767\\0.705\\
1.5/1.3\\
43.9\%\\35.5\%\\
\cell{l}{\cellcolor{blue!25}1.40}\\108\\\cell{l}{\cellcolor{lime}12.4\%}\\\cell{l}{\cellcolor{lime}36.0\%}} &
\cell{r}{0.762\\0.697\\
1.3/1.3\\
39.8\%\\32.5\%\\
1.00\\68\\17.8\%\\47.0\%} &
\cell{r}{0.77\\0.728\\
1.5/1.3
\\49\%\\\cell{l}{\cellcolor{blue!25}50.6\%}\\
\cell{l}{\cellcolor{blue!25}1.36}\\100\\12.9\%\\35.8\%} &
\cell{r}{0.762\\0.703\\
1.3/1.3\\
40\%\\34.8\%\\
1.00\\68\\17.9\%\\46.1\%} &
\cell{r}{0.768\\0.707\\
1.5/1.3\\
44.9\%\\42.9\%\\
\cell{l}{\cellcolor{blue!25}1.44}\\95\\12.3\%\\36.2\%} \\ 
   
\midrule

\datacell{mixed\_downstream}{$y = x_1 + z - 0.5$}{$\hat{y} = x_1 + x_2^{non} + z + b$}{$\prob{Z=1}=0.15$} & \metriclabels{} & 
\cell{r}{0.813\\0.697\\
1.0/1.0\\
25.9\%\\0.1\%\\
-\\-\\-\\-} & 
\cell{r}{0.799\\0.802\\
1.3/1.0\\
35.5\%\\63.1\%\\
-\\-\\-\\-} & 
\cell{r}{0.802\\0.764\\
1.2/1.2\\
19.5\%\\16\%\\
3.54\\435\\-\\-} & 
\cell{r}{0.808\\0.771\\
1.1/1.2\\
18.3\%\\24.1\%\\
3.24\\496\\-\\-} & 
\cell{r}{0.783\\0.747\\
1.5/1.3\\
46\%\\44.5\%\\
1.38\\105\\10.6\%\\33.0\%} &
\cell{r}{0.775\\0.747\\
1.3/1.3\\
41.1\%\\43.2\%\\
1.00\\70\\15.2\%\\42.1\%} &
\cell{r}{0.783\\0.757\\
1.5/1.3\\
49.5\%\\54.3\%\\
1.39\\90\\10.4\%\\36.4\%} &
\cell{r}{0.775\\0.752\\
1.2/1.3\\
40.9\%\\42.7\%\\
1.00\\69\\15.0\%\\43.2\%} &
\cell{r}{0.785\\0.747\\
1.5/1.3\\
46.6\%\\45.8\%\\
1.39\\93\\10.6\%\\34.7\%} \\ 
   
\midrule

\datacell{gaming}{$y = x_1 + z - 0.5$}{$\hat{y} = x_1^{non} + x_2^{non} + z + b$}{$\prob{Z=1}=0.15$} & \metriclabels{} & 
\cell{r}{0.805\\0.697\\
1.0/1.0\\
25.2\%\\\cell{l}{\cellcolor{blue!25}0.4\%}\\
-\\-\\-\\-} & 
\cell{r}{0.794\\0.799\\
1.3/1.0\\
35.1\%\\65.5\%\\
-\\-\\-\\-} & 
\cell{r}{0.796\\0.763\\
1.2/1.2\\
19.3\%\\17.1\%\\
3.54\\440\\-\\-} & 
\cell{r}{0.803\\0.769\\
1.2/1.2\\
18.2\%\\24.5\%\\
3.35\\502\\-\\-} & 
\cell{r}{0.761\\0.707\\
1.3/1.3\\
39.1\%\\28.4\%\\
1.68\\120\\3.2\%\\48.7\%} &
\cell{r}{0.741\\0.703\\
1.2/1.3\\
40.7\%\\34.2\%\\
1.00\\76\\5.0\%\\68.5\%} &
\cell{r}{0.76\\0.732\\
1.4/1.4\\
45\%\\\cell{l}{\cellcolor{blue!25}45.3\%}\\\
1.53\\116\\3.5\%\\45.2\%} &
\cell{r}{0.741\\0.711\\
1.2/1.3\\
41.2\%\\39.1\%\\
1.00\\78\\4.7\%\\68.9\%} &
\cell{r}{0.759\\0.716\\
1.4/1.4\\
41.1\%\\36.8\%\\
1.60\\127\\3.1\%\\44.4\%} \\ 

\midrule

\datacell{german}{$y = 0.3(-l -d + i + s)$}{\cell{l}{$\hat{y} = g + a + e + l$\\$+ d + i + s + b$}}{} & \cell{r}{%
Observed AUC\\
True AUC\\
Gain/Loss\\%
\% Approved (all)\\
\% Approved ($G=1$)\\
$\mathbb{E}$[Reapplications]\\
\# Never Approved\\
\% Net Improvement\\
\% Invalid Recourse 
} &
\cell{r}{0.74\\0.572\\
1.0/1.0\\
16.8\%\\1.4\%\\
-\\-\\-\\-} & 
\cell{r}{0.704\\0.678\\
2.0/1.1\\
44.3\%\\48.7\%\\
-\\-\\-\\-} & 
\cell{r}{0.989\\0.986\\
7.0/6.7\\
28.1\%\\27.1\%\\
3.56\\324\\-\\-} & 
\cell{r}{0.988\\0.986\\
1.9/1.0\\
23.6\%\\22.6\%\\
2.71\\466\\-\\-} & 
\cell{r}{0.726\\0.563\\
1.2/1.1\\
21.8\%\\8.4\%\\
1.47\\74\\-32.0\%\\69.4\%} &
\cell{r}{0.722\\0.572\\
1.2/1.4\\
37.7\%\\36.6\%\\
1.00\\75\\-45.0\%\\63.2\%} &
\cell{r}{0.711\\0.577\\
1.2/1.2\\
27.3\%\\15.7\%\\
1.40\\67\\-33.1\%\\72.3\%} &
\cell{r}{0.725\\0.571\\
1.2/81.4\\
38.2\%\\37.2\%\\
1.00\\77\\-44.0\%\\59.9\%} &
\cell{r}{0.724\\0.561\\
1.2/1.2\\
22.3\%\\9.0\%\\
1.49\\79\\-31.1\%\\69.7\%}
\end{tabular}
}
\caption{Overview of censoring and mitigation strategies across data generating processes. We highlight results referenced in our remarks: {\color{yellow}Censoring Does Not Resolve Itself Without Intervention}, {\color{pink}On the Difficulties of Detection}, {\color{brown!65}All Mitigation Methods Recover from Censoring}, {\color{lime}Recourse Works Best in Causal Settings}, {\color{orange}IPW for Greater Censored Group Uncertainty}, and {\color{blue!25}Combination Methods: A Solution For Unknown Causal Settings}. These phenomena arise across DGPs. We use $b$ to specify the intercept.}
\label{App::Table::ExperimentOverview}
\end{figure}

\clearpage

\section{Uncertainty Measurement}
\label{Appendix::Uncertainty Measurement}
\begin{figure}[!htbp]
\centering
\resizebox{\linewidth}{!}{
\cell{c}{\includegraphics[trim={0 6.5in 0 1.3in}, clip, width=0.3\linewidth]{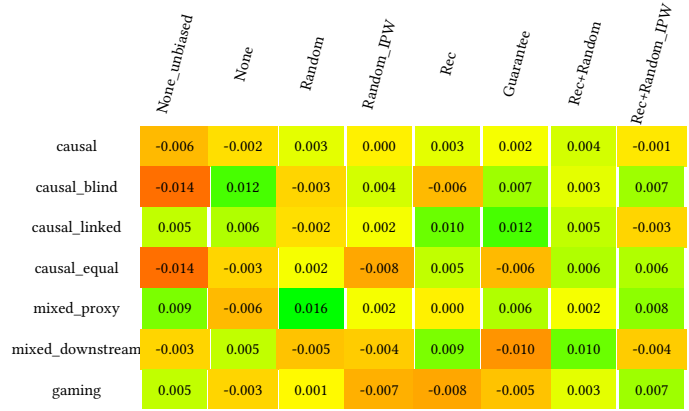}}
}
\caption{Heatmap of the Difference in the Average Miscalibration Area (MA) across Training and New Datasets. Miscalibration area is a means of quantifying how much a model's normalized residuals do not fit the Gaussian curve. In this case, we divide each dataset (training, new) into training, validation, and test sets to train a uncertainty model and produce the sharpness score. We calculate the MA across all of the points in the training dataset and the new points (never predicted upon) dataset at each period and then average across all periods to report the average MA for each experiment. We then subtract the MA for the new datasets from the training dataset and report the value here (train - true). A positive value indicates a higher training MA whereas a negative value indicates a higher true MA. In general, we observe very small differences between training and new (unbiased dataset sample) points, indicating a training MA better than that of deployment. This is due to selective labeling, showing that we may still be slightly overestimating the performance metrics of the system. We find that the estimated deployment MA is lower than than it in setups with less causal features (more orange towards the lower half of the plot), meaning that the safeguards don't perform as well with fewer causal features, although they are positive with the \textmethod{Rec+IPW} \textds{gaming} setup.}
\label{Fig::Heatmap of Miscalibration Area}
\end{figure}

\begin{figure}[!htbp]
\centering
\resizebox{\linewidth}{!}{
\cell{c}{\includegraphics[trim={0 6.5in 0 1.3in}, clip, width=0.3\linewidth]{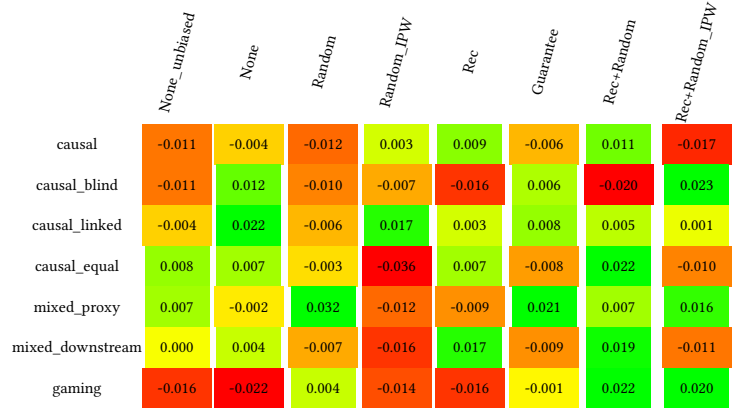}}
}
\caption{Heatmap of the Difference in the Average Sharpness Scores across Training and New Datasets. Sharpness score is defined by ~\citet[][]{kuleshov2018accurate} as the average variance of the uncertainty estimates on the test set. In this case, we split each dataset (training, new) into training, validation, and test sets to fit a uncertainty model and compute the sharpness score. We calculate the sharpness score over all of the points in the training dataset and the new points (never predicted upon) dataset at each period and then average across all periods to report the average MA for each experiment. We then subtract the sharpness score for the new datasets from the training dataset and report the value here (train - true). A positive value indicates a higher training sharpness score whereas a negative value indicates a higher true sharpness score. We desire a low sharpness score, similar to wanting to have a low ML-predicted standard deviation. We aim to minimize the difference between the training and true sharpness scores. Generally we do not observe large differences across \textmethod{NoCensoring} and \textmethod{None}, further highlighting the difficulty in detecting censoring through uncertainty or other performance metrics. 
}
\label{Fig::Heatmap of Sharpness Score}
\end{figure}

\begin{figure}[htbp]
\centering
\resizebox{\linewidth}{!}{
\begin{tabular}{p{0.55in}p{2in}p{2in}p{2in}}
  \toprule
    \textbf{DGP} & Train (Observed) & True & Difference (Train - True)\\
  \toprule
  \cell{c}{\textds{causal}} & \cell{c}{\includegraphics[trim={0 0 0 1.5cm}, clip, width=\linewidth]{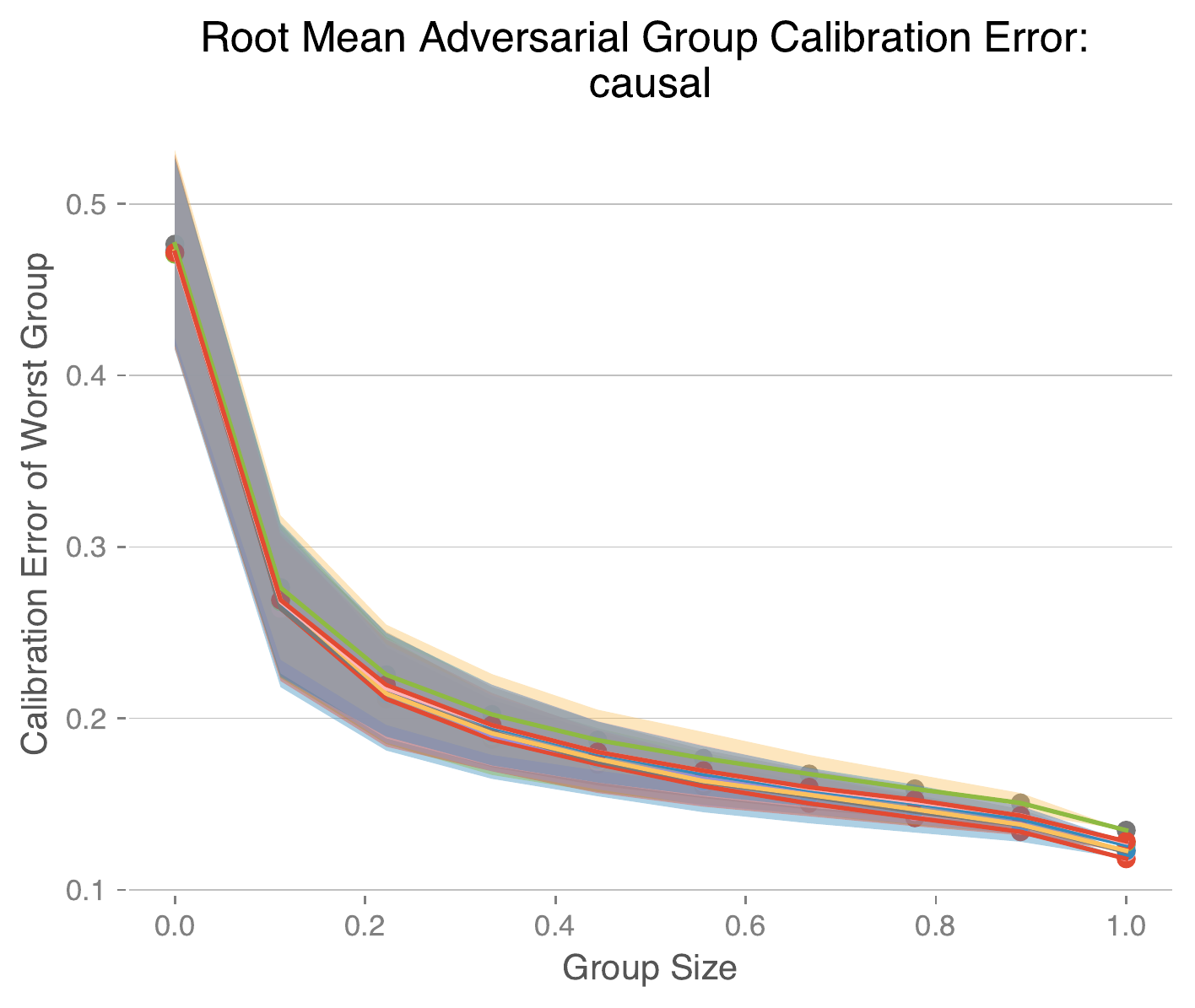}} &
  \cell{c}{\includegraphics[trim={0 0 0 1.5cm}, clip, width=\linewidth]{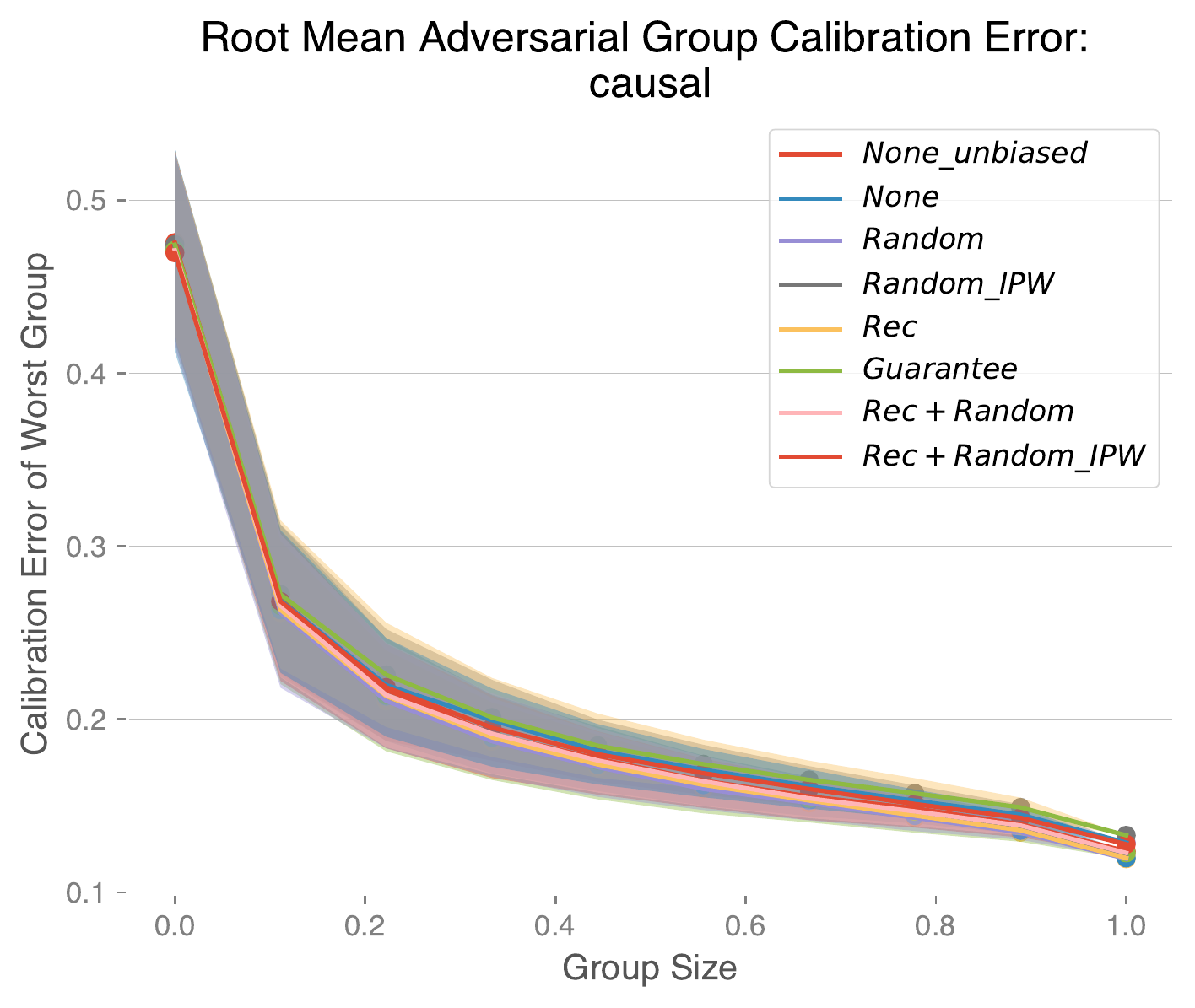}} &
  \cell{c}{\includegraphics[trim={0 0 0 1.5cm}, clip, width=\linewidth]{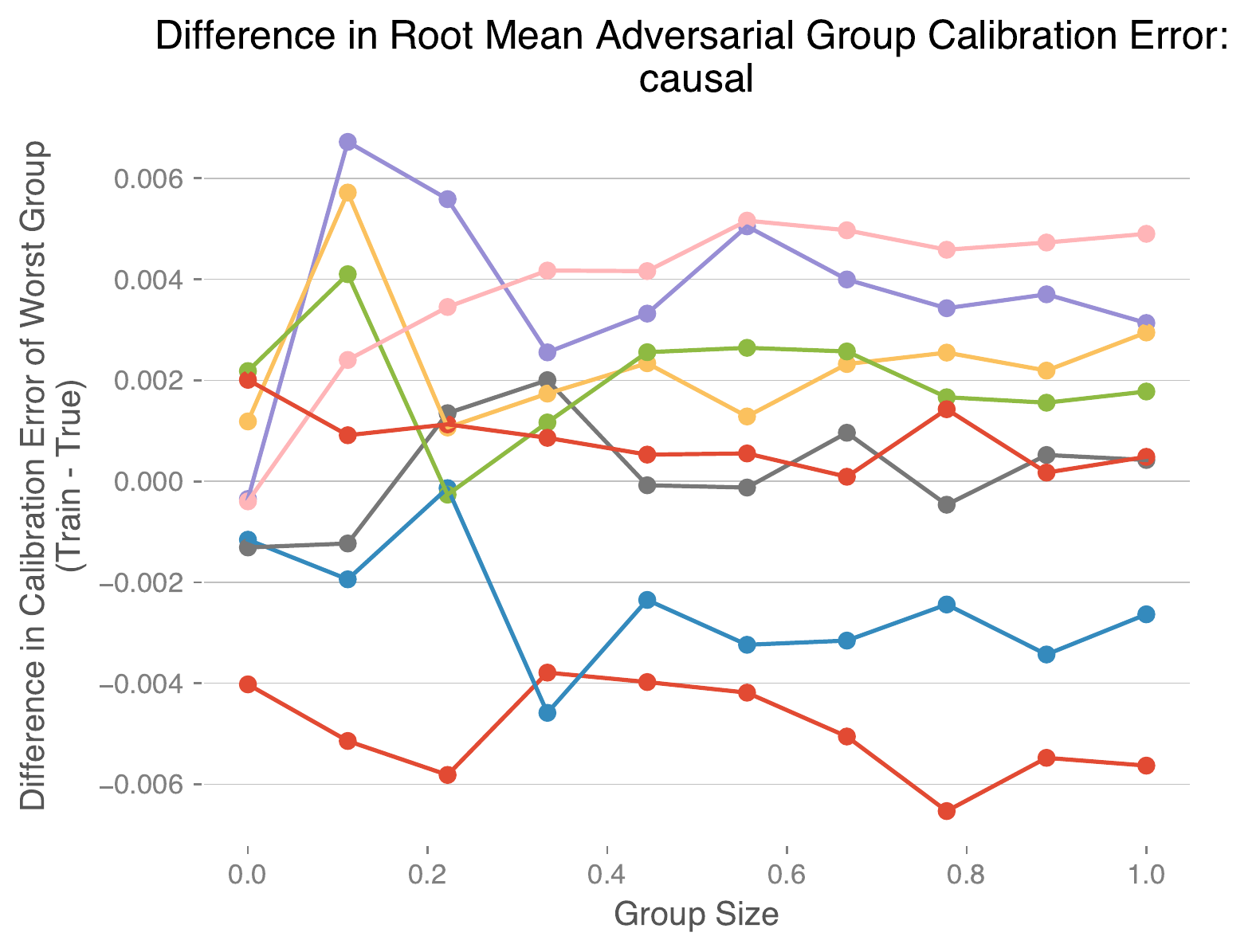}}
  \\
  \midrule
  \cell{c}{\textds{causal\_blind}} & \cell{c}{\includegraphics[trim={0 0 0 1.5cm}, clip, width=\linewidth]{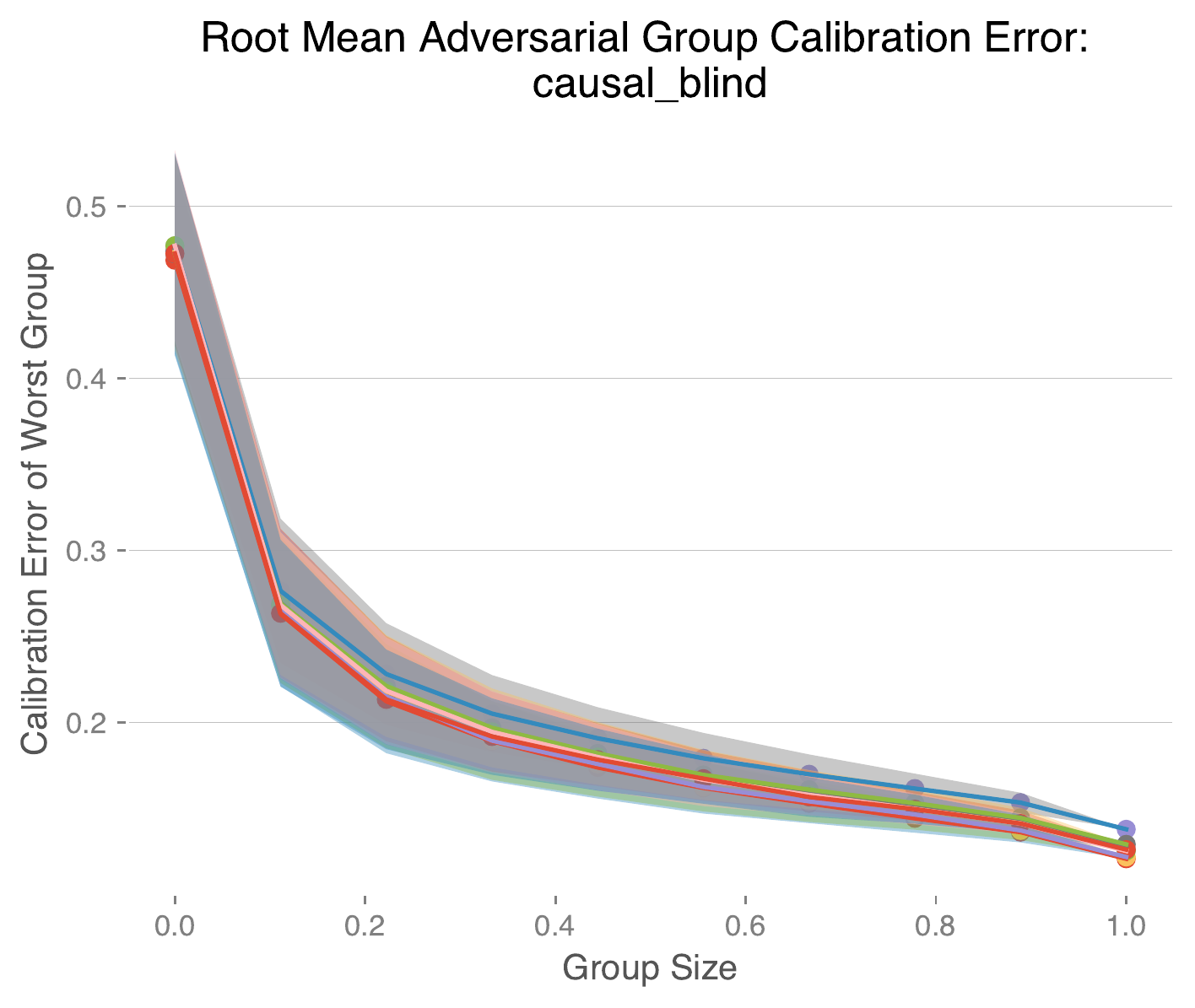}} &
  \cell{c}{\includegraphics[trim={0 0 0 1.5cm}, clip, width=\linewidth]{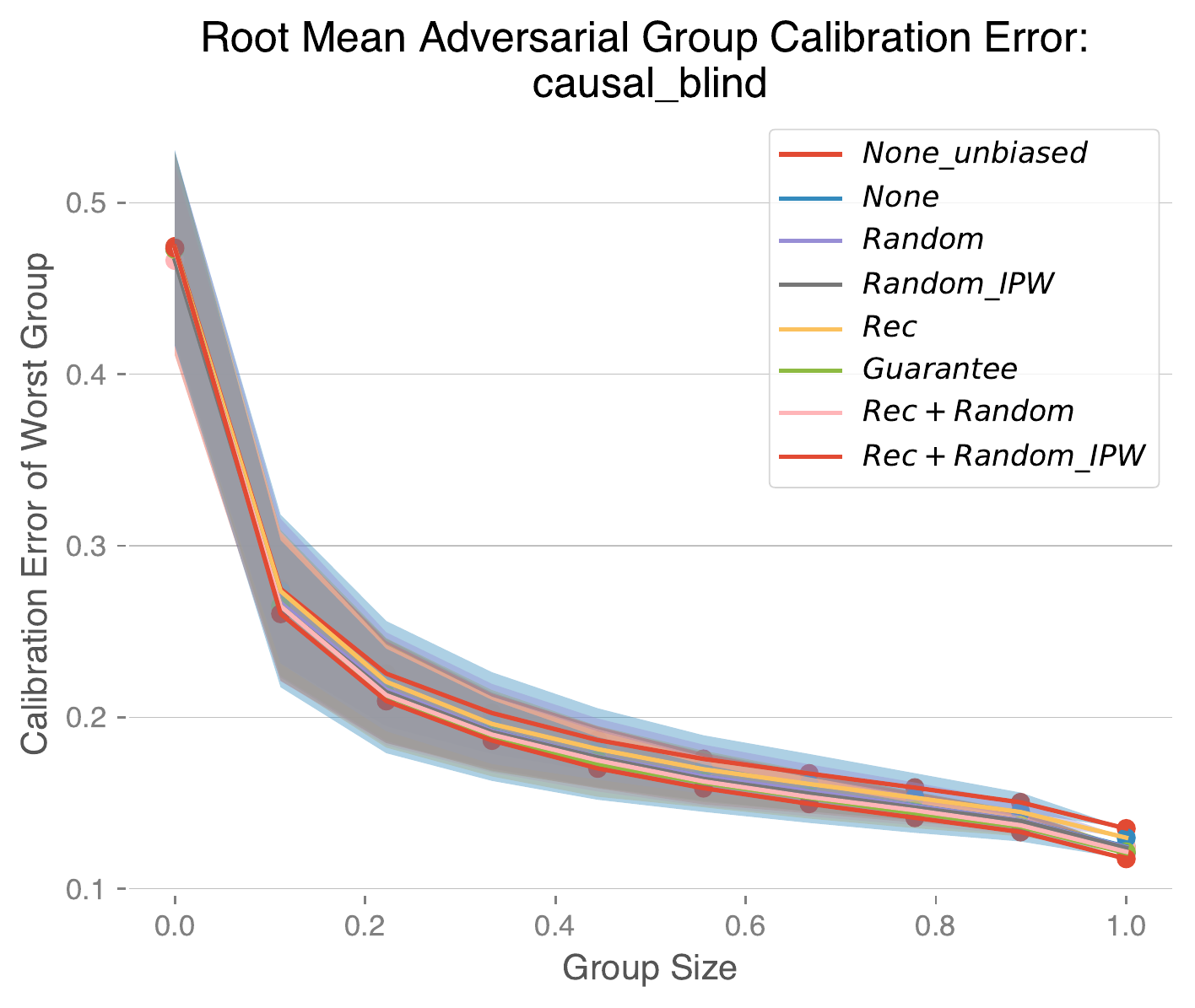}} &
  \cell{c}{\includegraphics[trim={0 0 0 1.5cm}, clip, width=\linewidth]{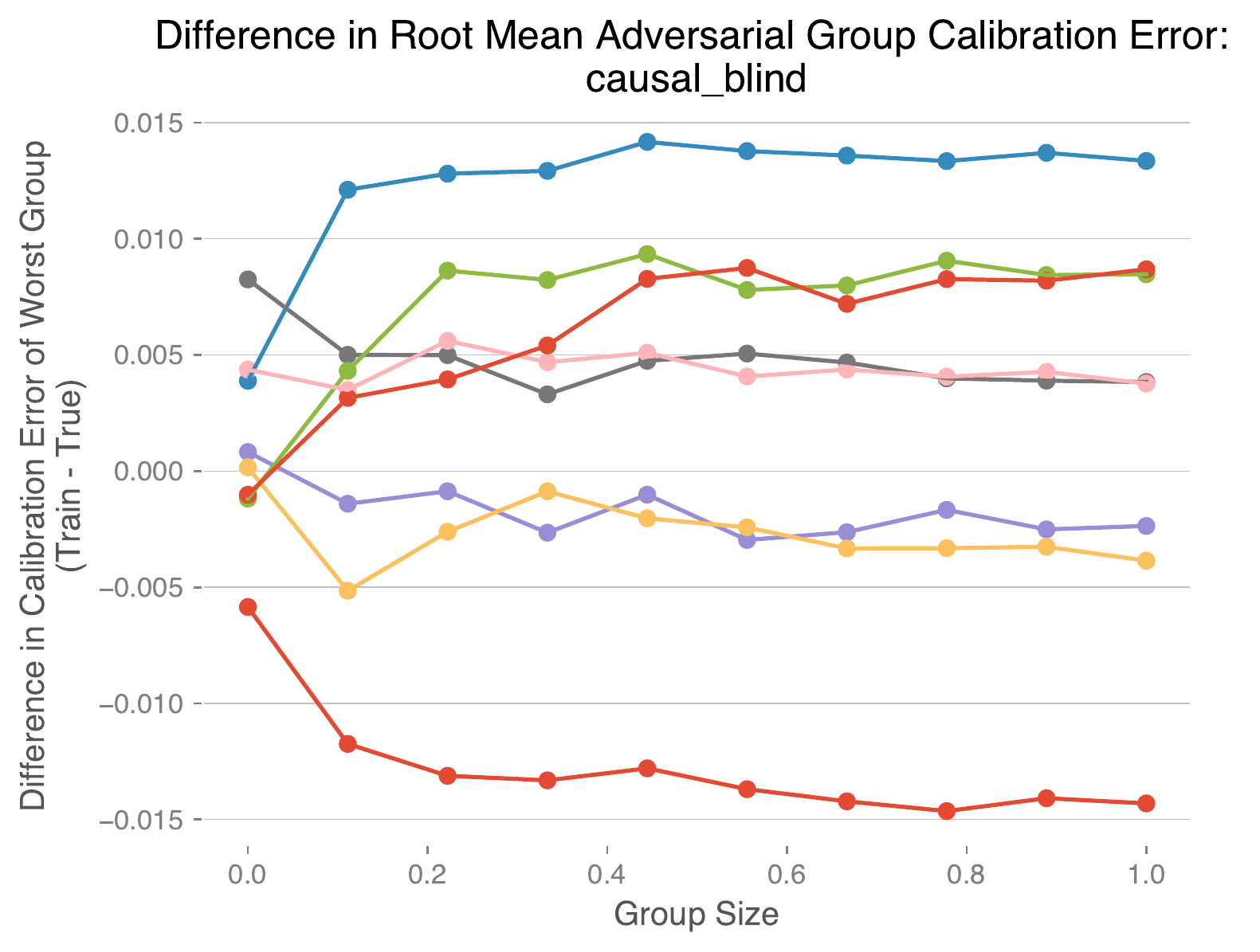}}
  \\
  \midrule
  \cell{c}{\textds{causal\_linked}} & \cell{c}{\includegraphics[trim={0 0 0 1.5cm}, clip, width=\linewidth]{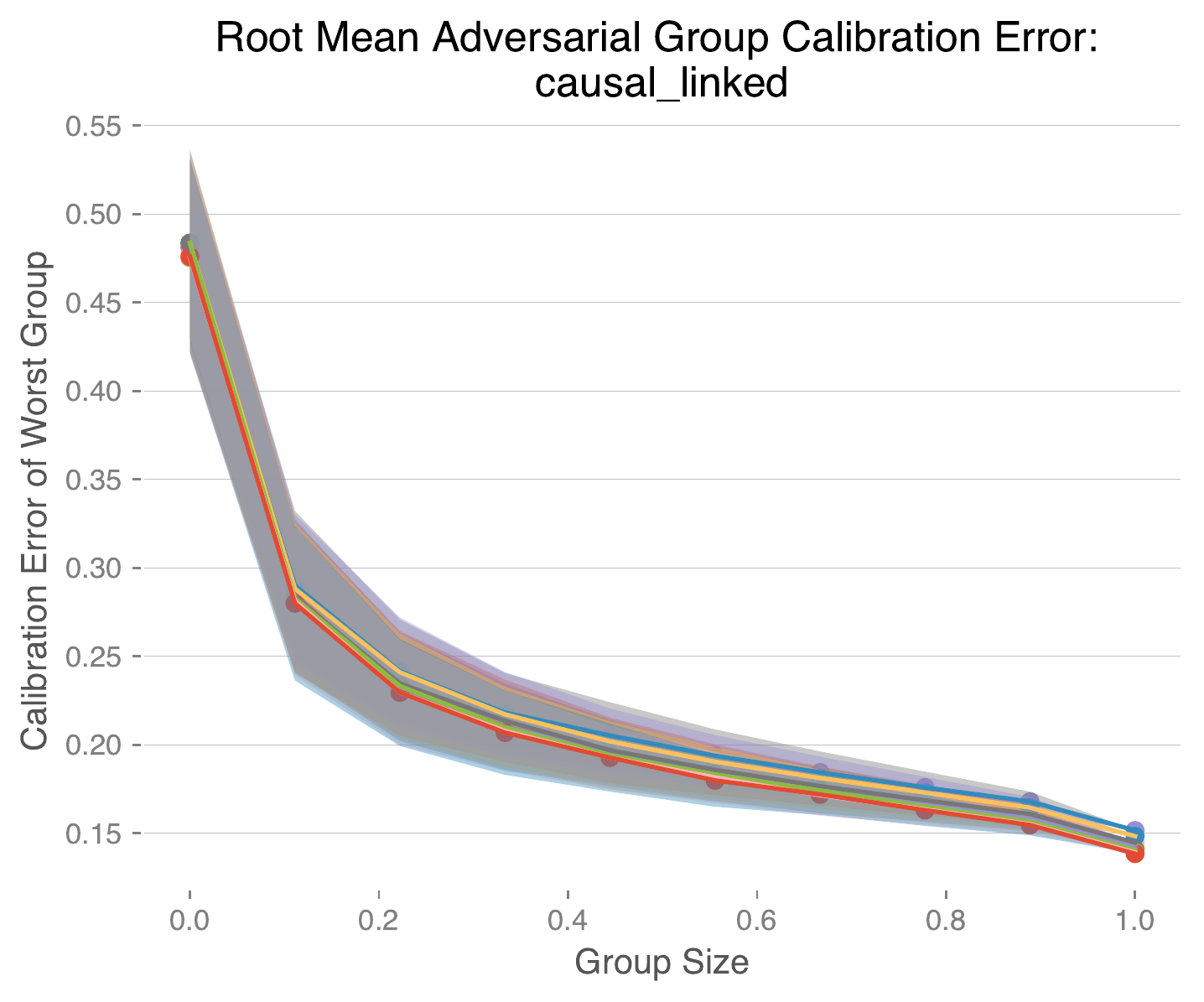}} &
  \cell{c}{\includegraphics[trim={0 0 0 1.5cm}, clip, width=\linewidth]{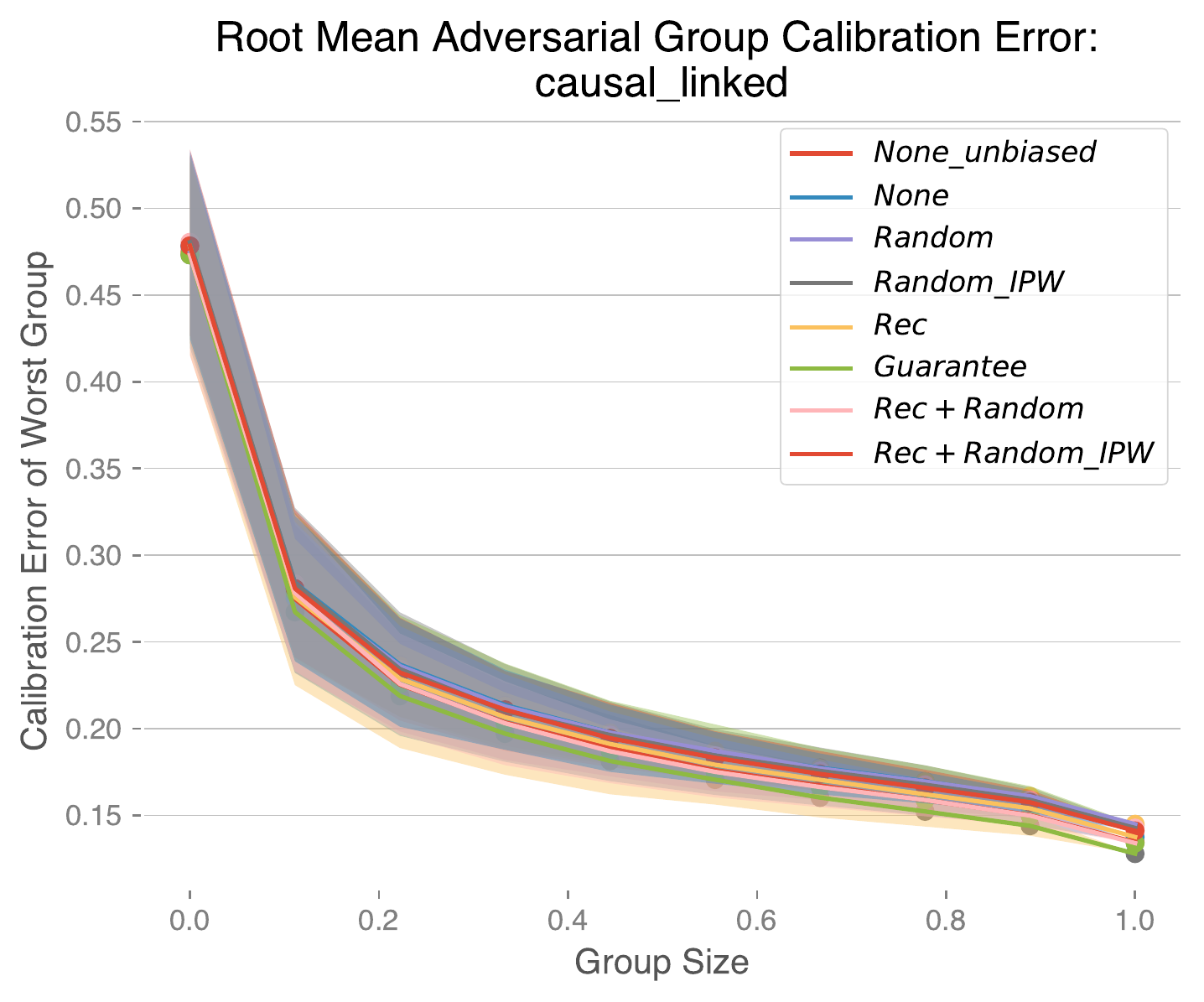}}  &
  \cell{c}{\includegraphics[trim={0 0 0 1.5cm}, clip, width=\linewidth]{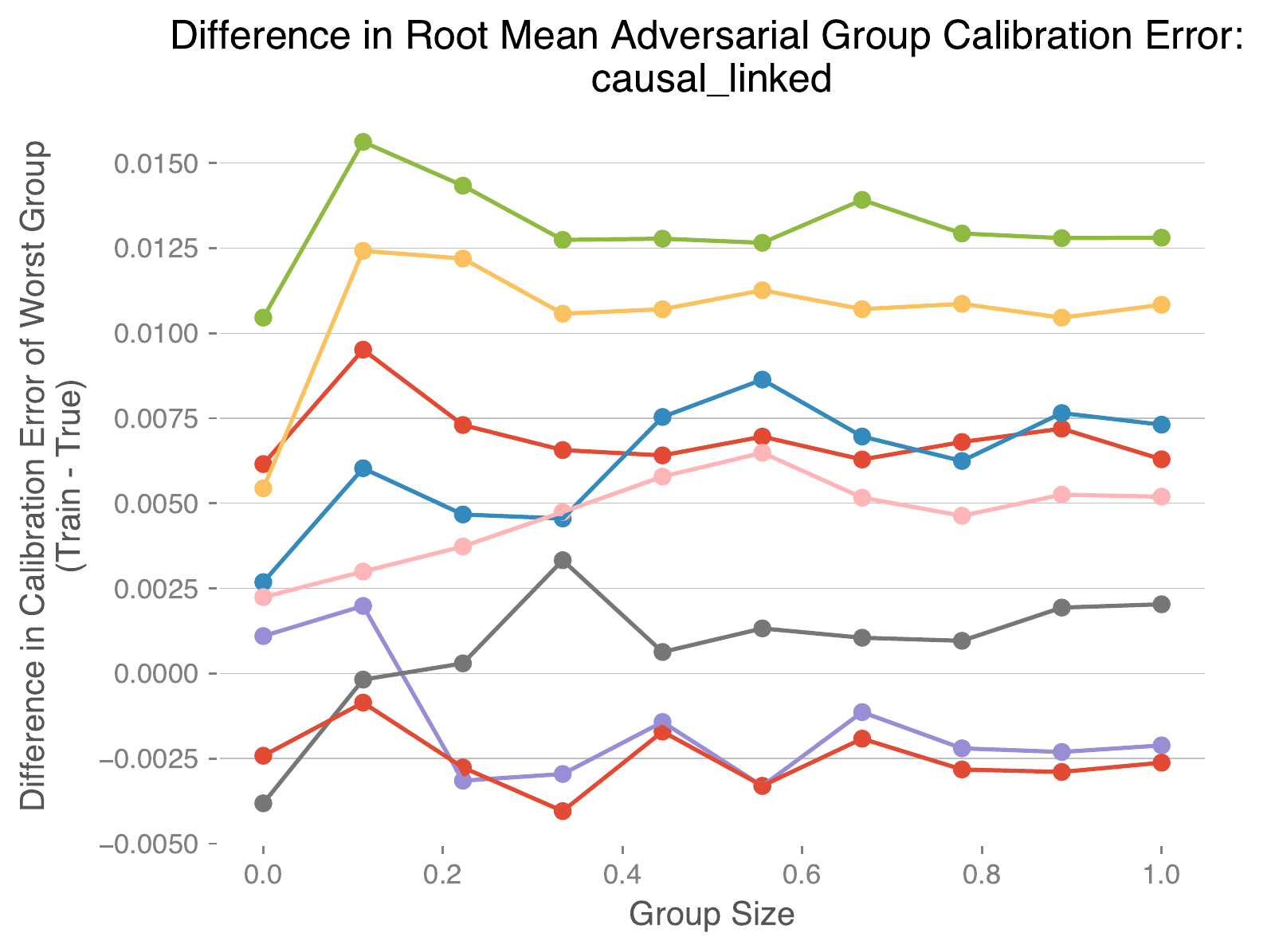}}\\
  \midrule
  \cell{c}{\textds{causal\_equal}} & \cell{c}{\includegraphics[trim={0 0 0 1.5cm}, clip, width=\linewidth]{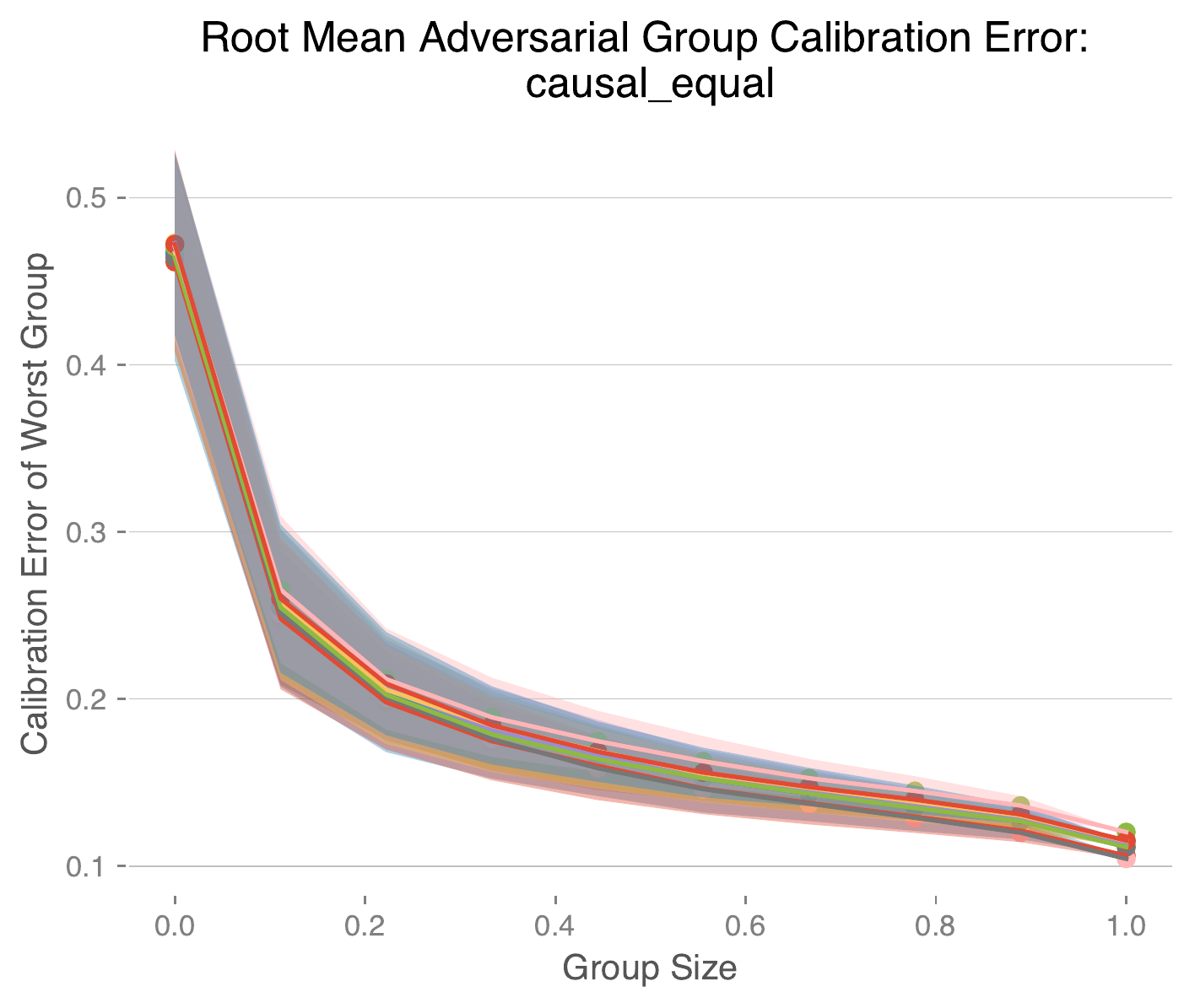}} &
  \cell{c}{\includegraphics[trim={0 0 0 1.5cm}, clip, width=\linewidth]{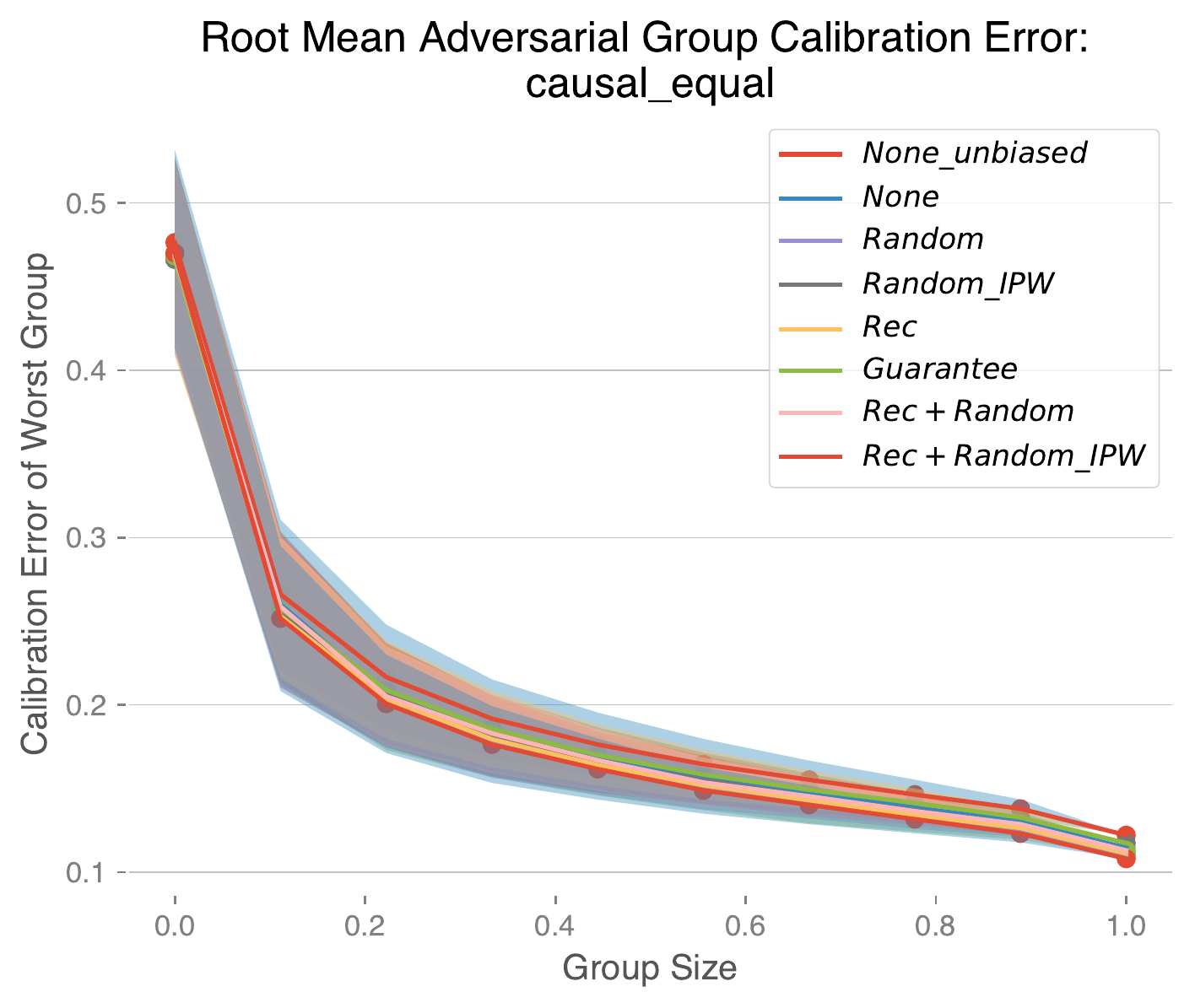}}  &
  \cell{c}{\includegraphics[trim={0 0 0 1.5cm}, clip, width=\linewidth]{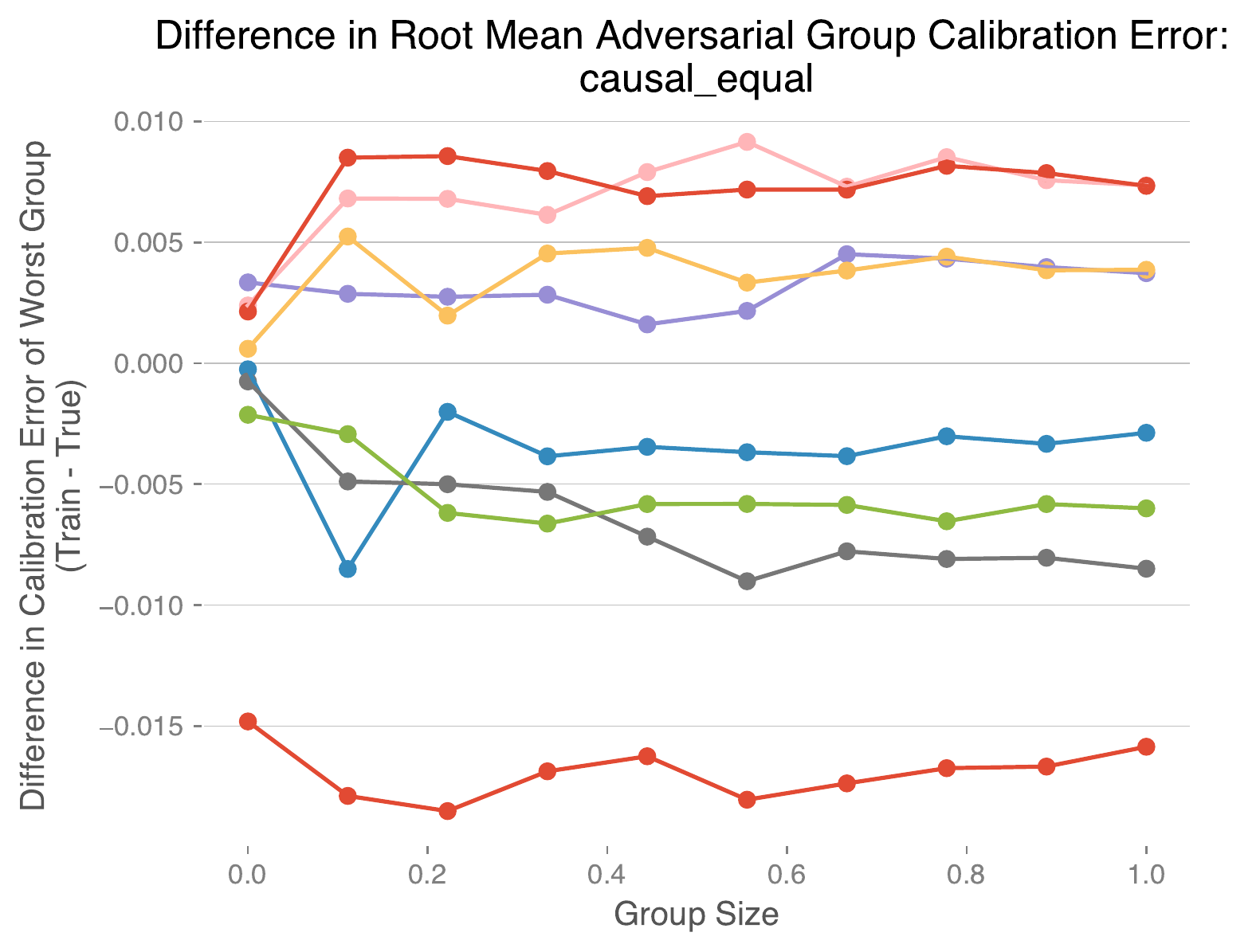}}\\
  \bottomrule
\end{tabular}
}
\caption{Scatterplots of Root Mean Squared Adversarial Group Calibration Error. To further demonstrate difficulty in detection, we show the adversarial group calibration averaged across all replicates for a given period and then averaged across all periods. In addition we plot one standard deviation above and below. Group size refers to the proportion of test dataset size. For 'Training', we calculate this metric on the training accumulate at each period. For 'True', we calculate this metric on a unbiased sample drawn from the entire population. For ease of comparison, we plot the difference between the training and true metric. Overall, we observe little difference in the root mean squared adversarial group error, highlight the difficulty in detecting censoring.}
\label{Fig::RMSAG1}
\end{figure}

\begin{figure}[htbp]
\centering
\resizebox{\linewidth}{!}{
\begin{tabular}{p{0.55in}p{2in}p{2in}p{2in}}
  \toprule
    \textbf{DGP} & Train (Observed) & True & Difference (Train - True)\\
  \toprule
\cell{c}{\textds{mixed\_proxy}} & \cell{c}{\includegraphics[trim={0 0 0 1.5cm}, clip, width=\linewidth]{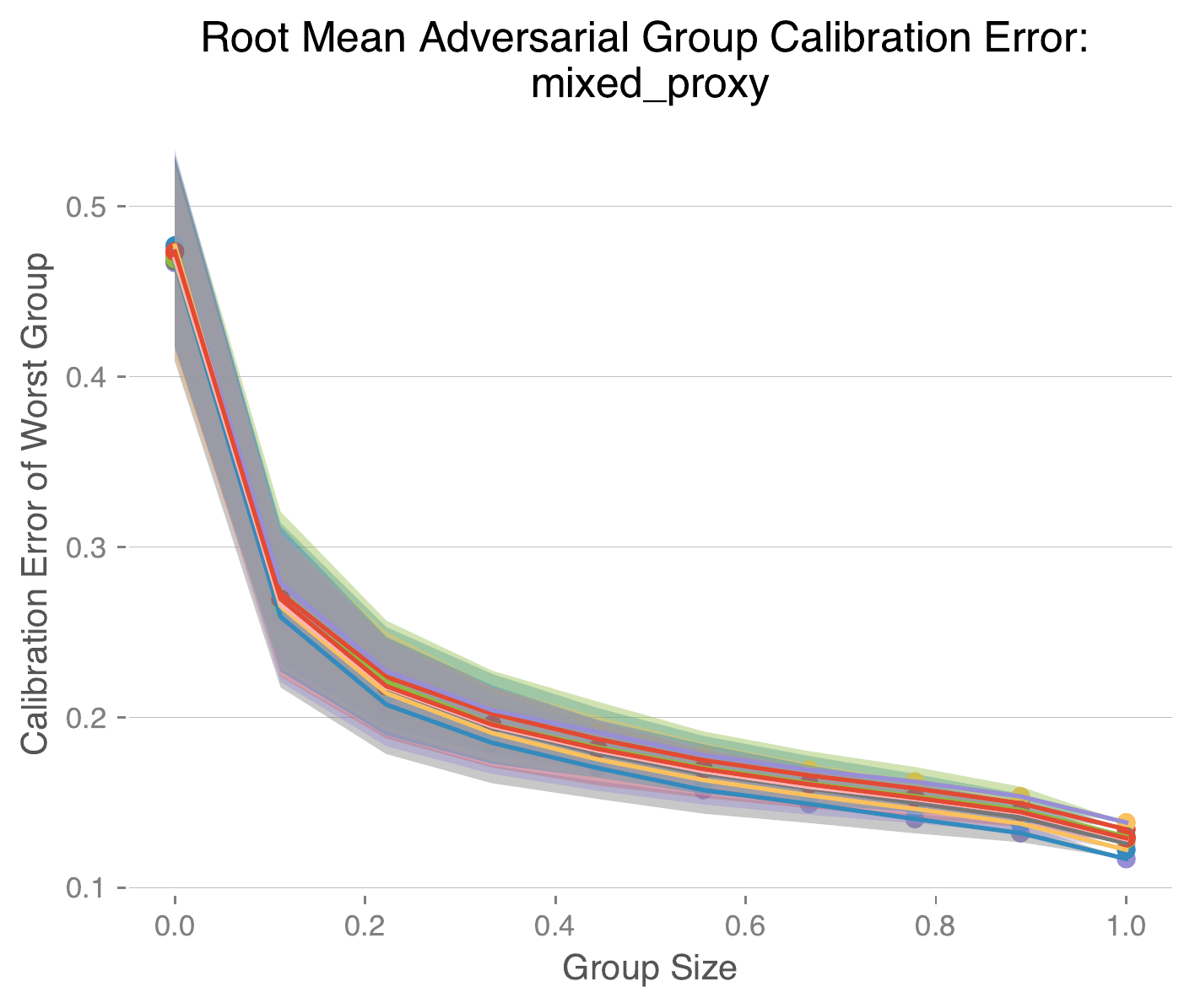}} &
  \cell{c}{\includegraphics[trim={0 0 0 1.5cm}, clip, width=\linewidth]{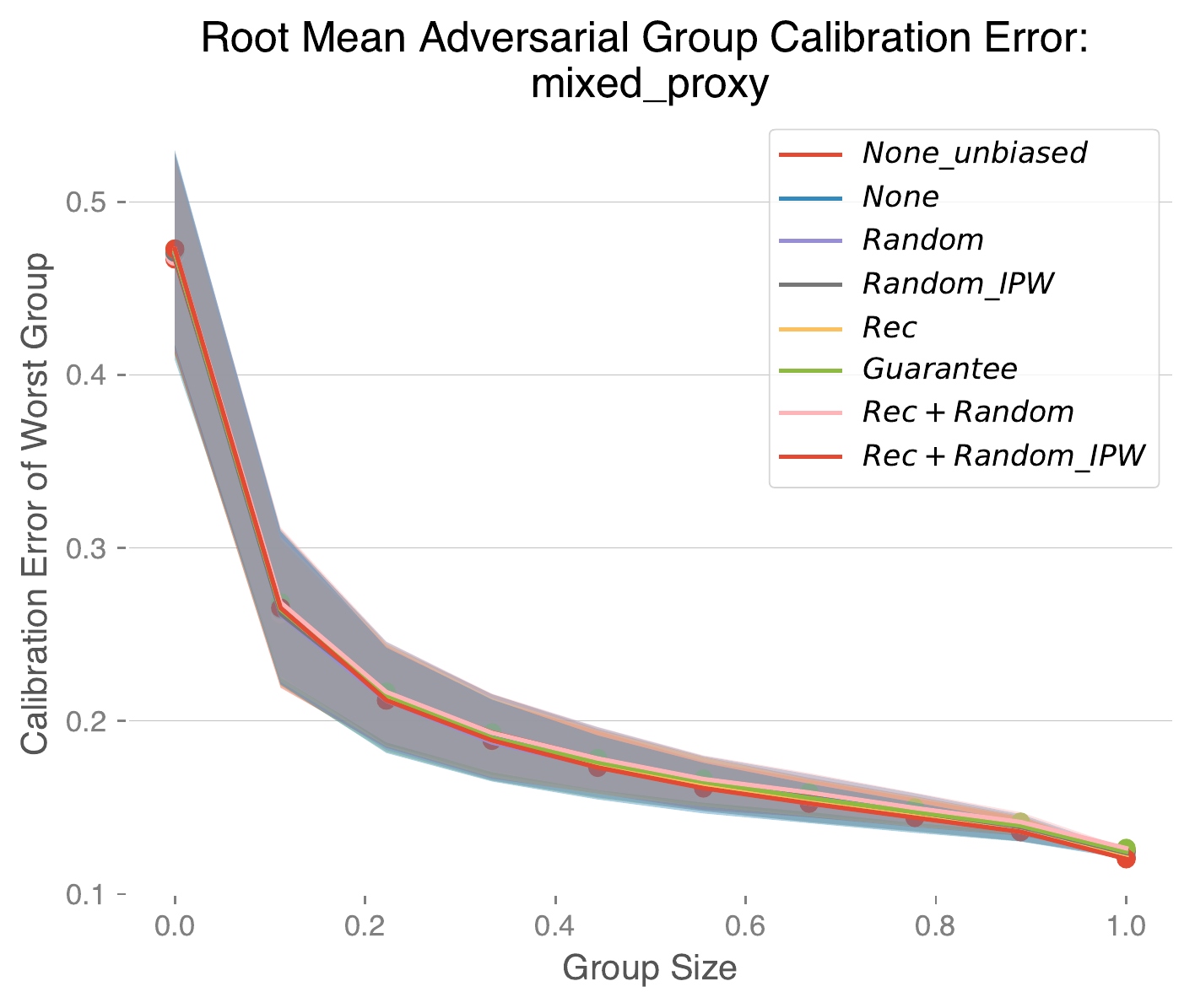}}  &
  \cell{c}{\includegraphics[trim={0 0 0 1.5cm}, clip, width=\linewidth]{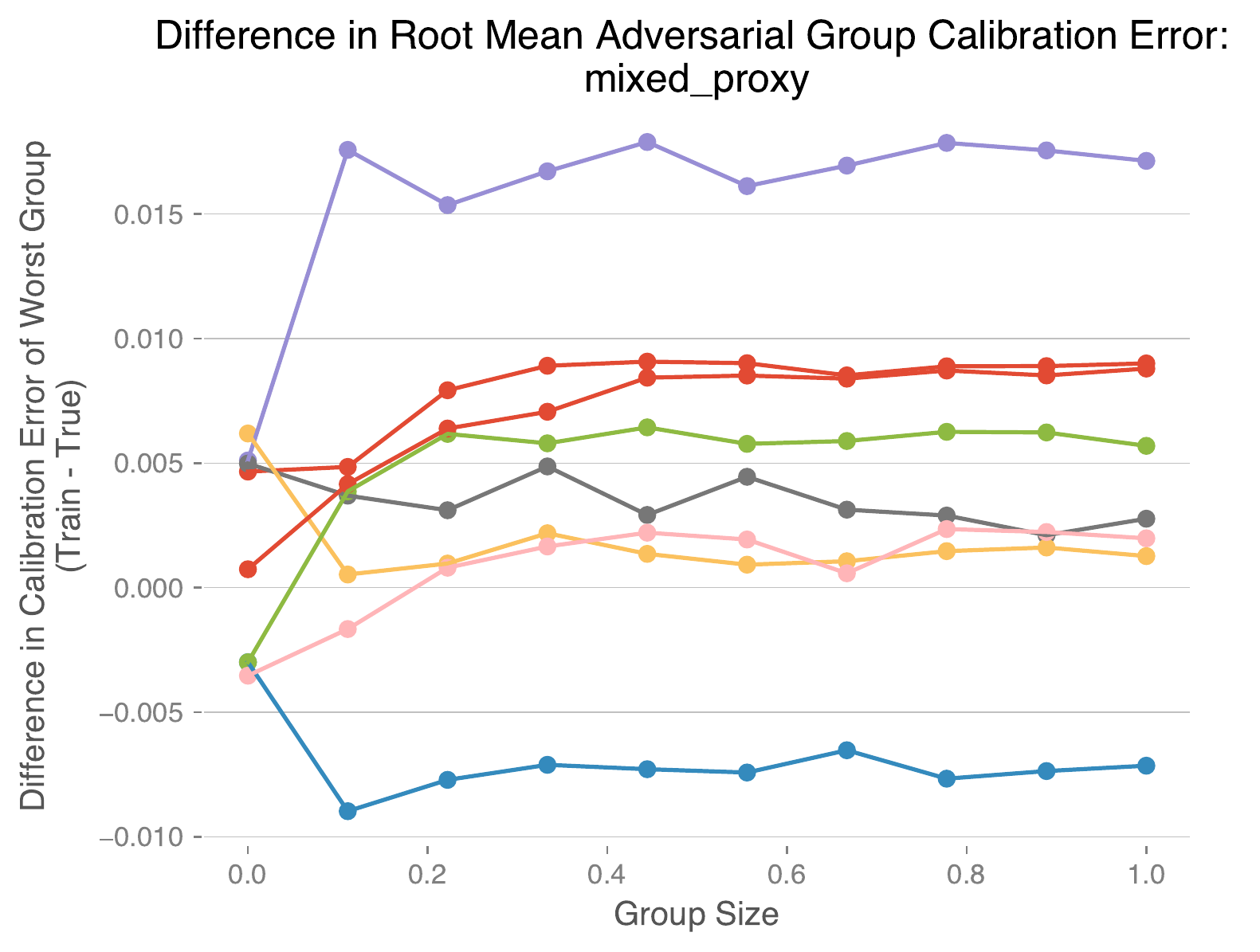}}\\
  \midrule
  \cell{c}{\textds{mixed\_downstream}} & \cell{c}{\includegraphics[trim={0 0 0 1.5cm}, clip, width=\linewidth]{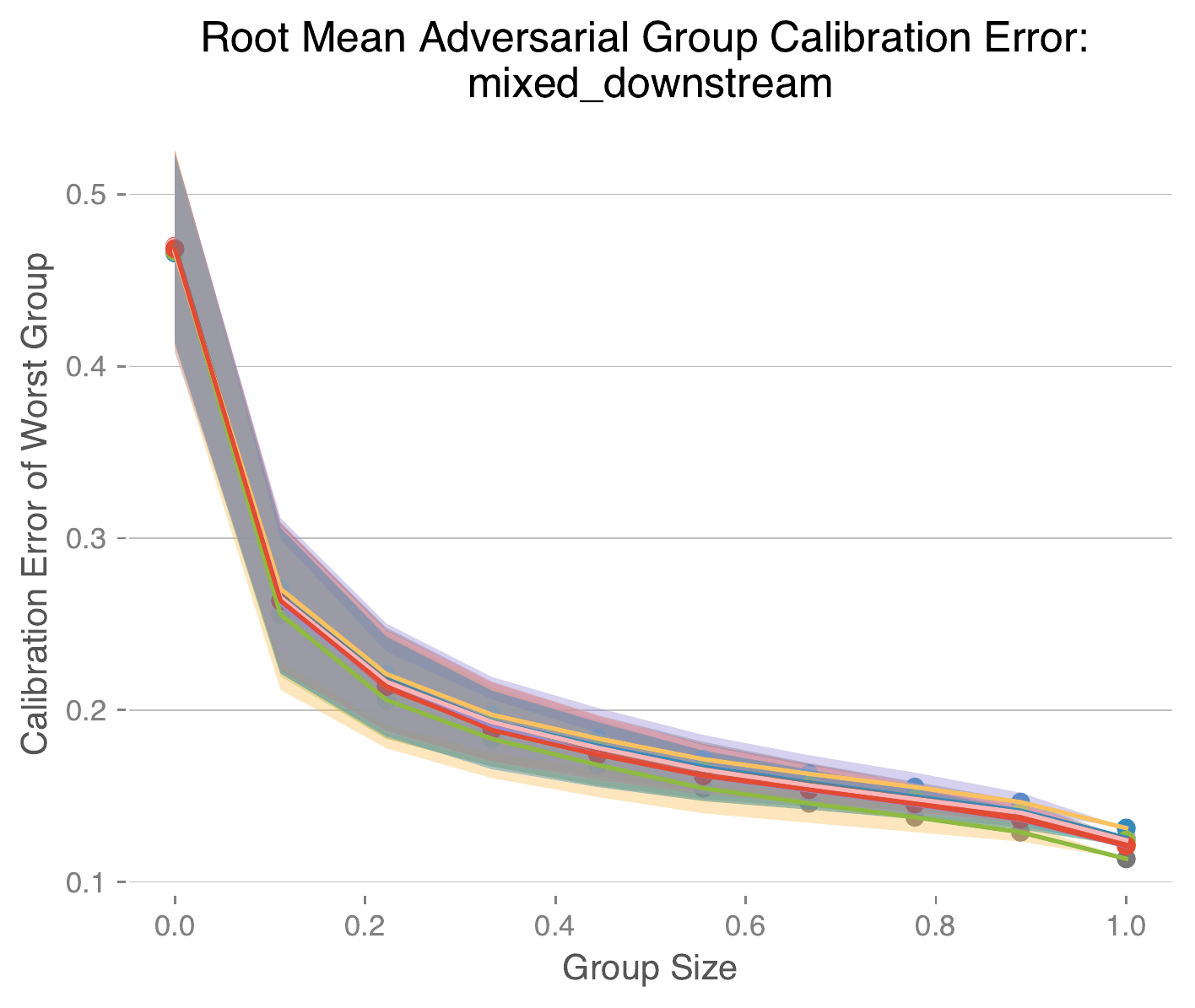}} &
  \cell{c}{\includegraphics[trim={0 0 0 1.5cm}, clip, width=\linewidth]{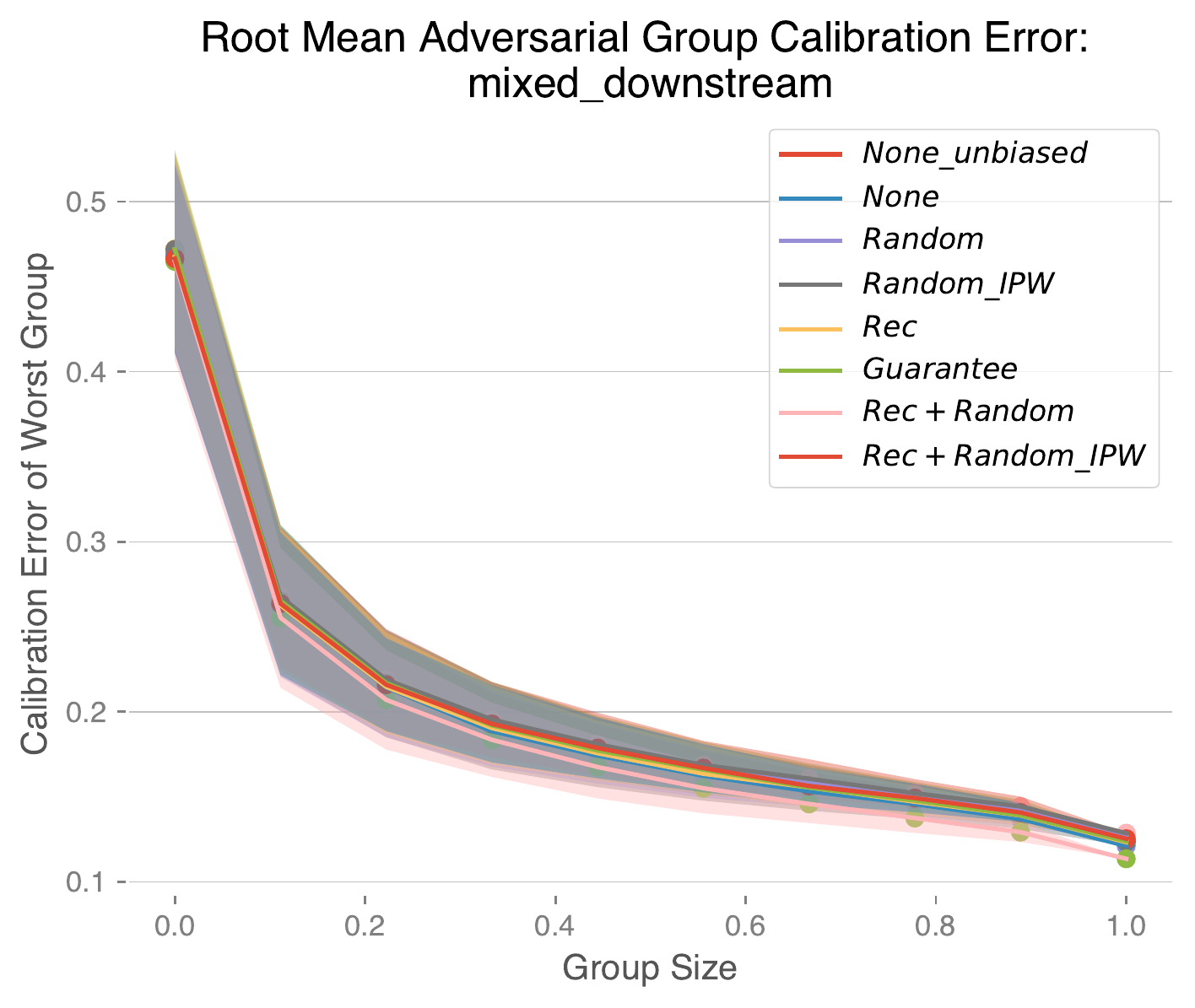}}  &
  \cell{c}{\includegraphics[trim={0 0 0 1.5cm}, clip, width=\linewidth]{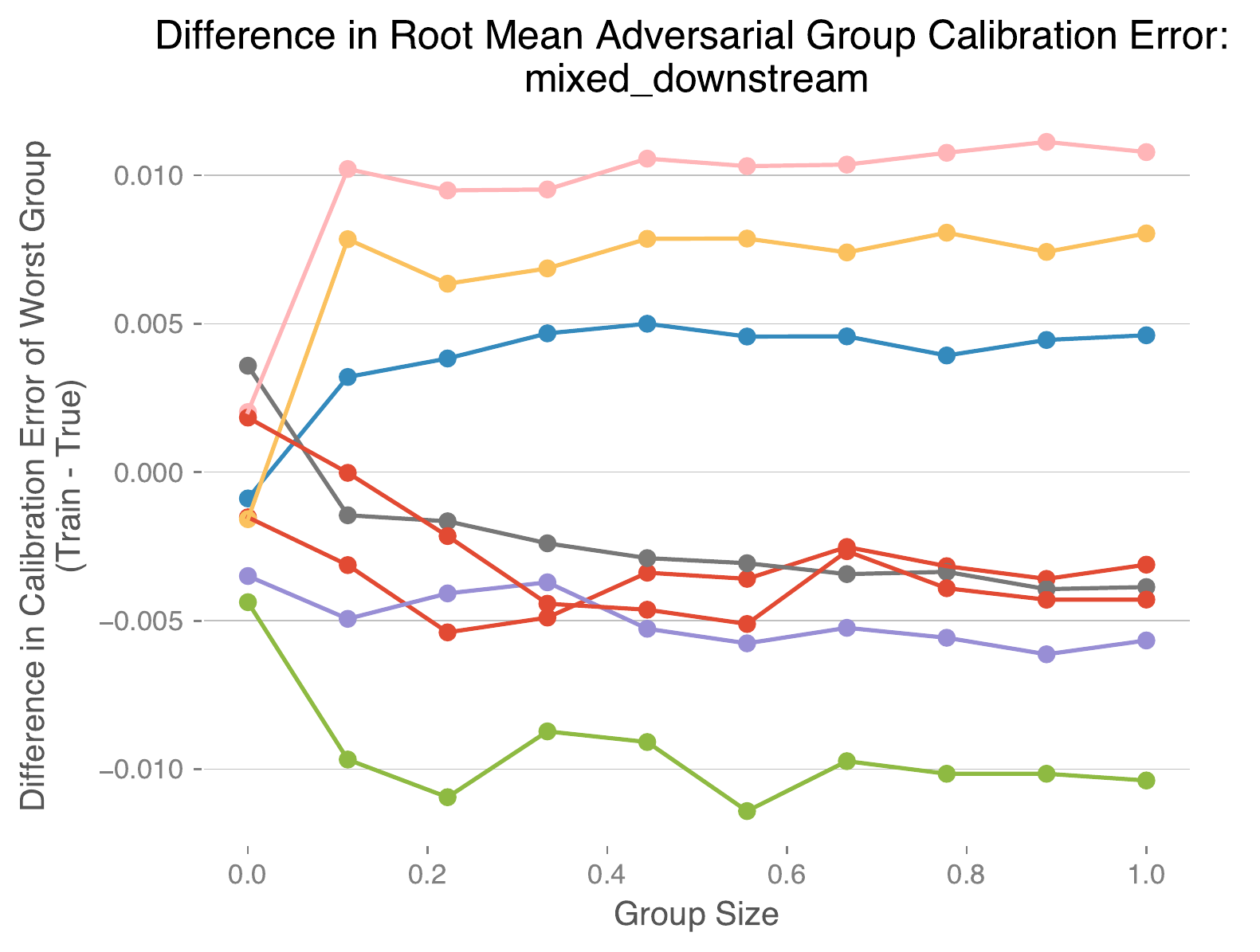}}\\
  \midrule
  \cell{c}{\textds{gaming}} & \cell{c}{\includegraphics[trim={0 0 0 1.5cm}, clip, width=\linewidth]{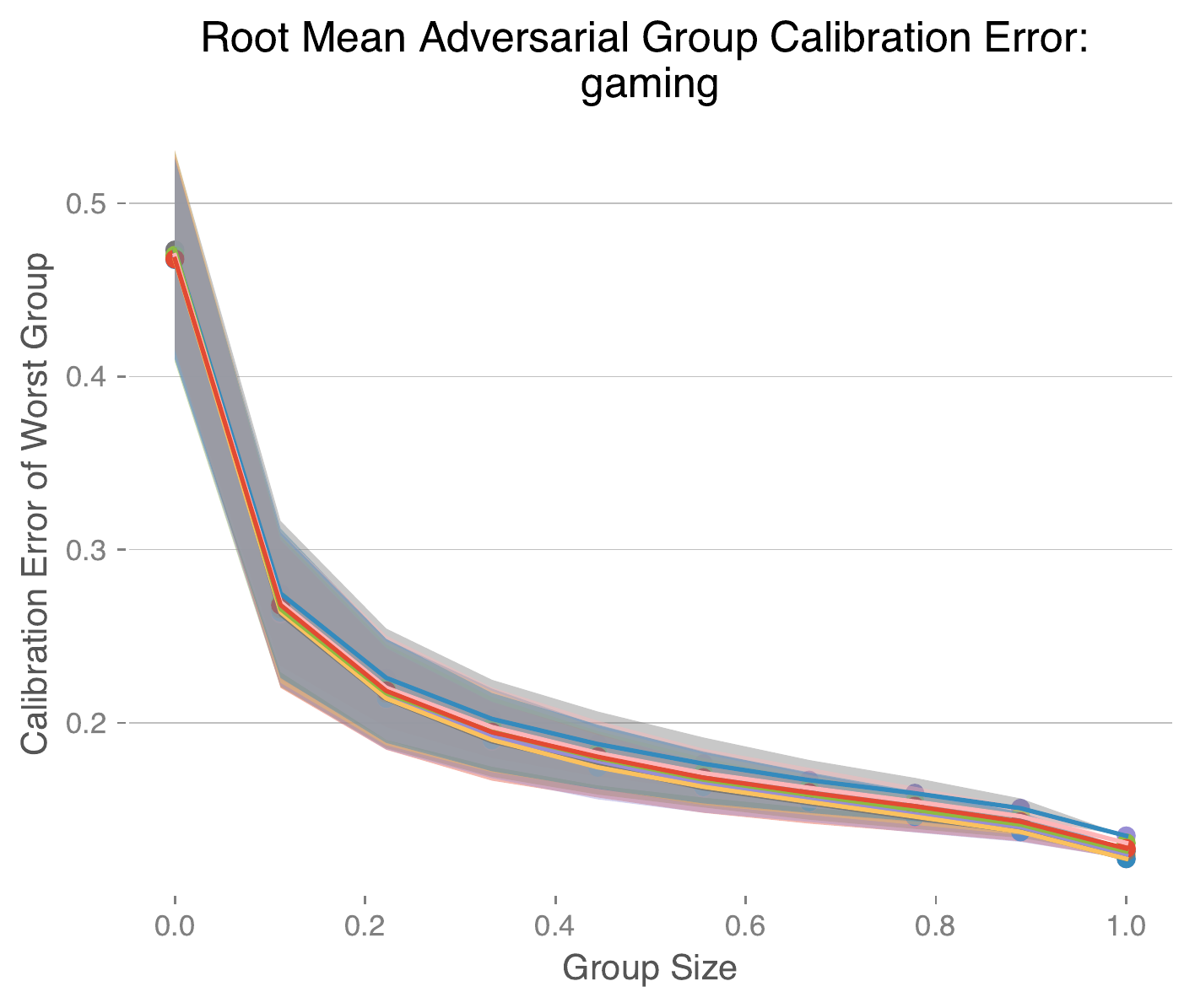}} &
  \cell{c}{\includegraphics[trim={0 0 0 1.5cm}, clip, width=\linewidth]{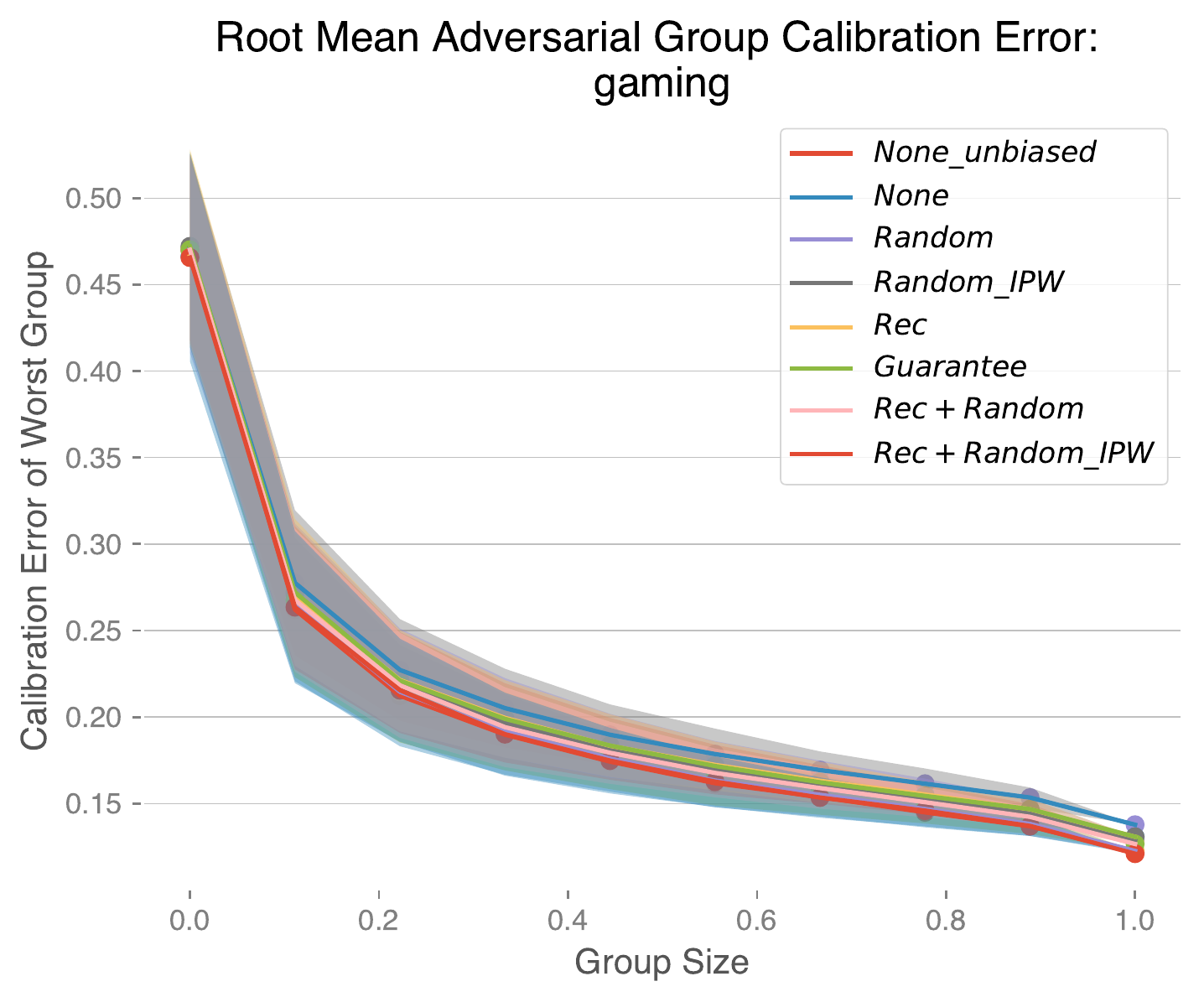}}  &
  \cell{c}{\includegraphics[trim={0 0 0 1.5cm}, clip, width=\linewidth]{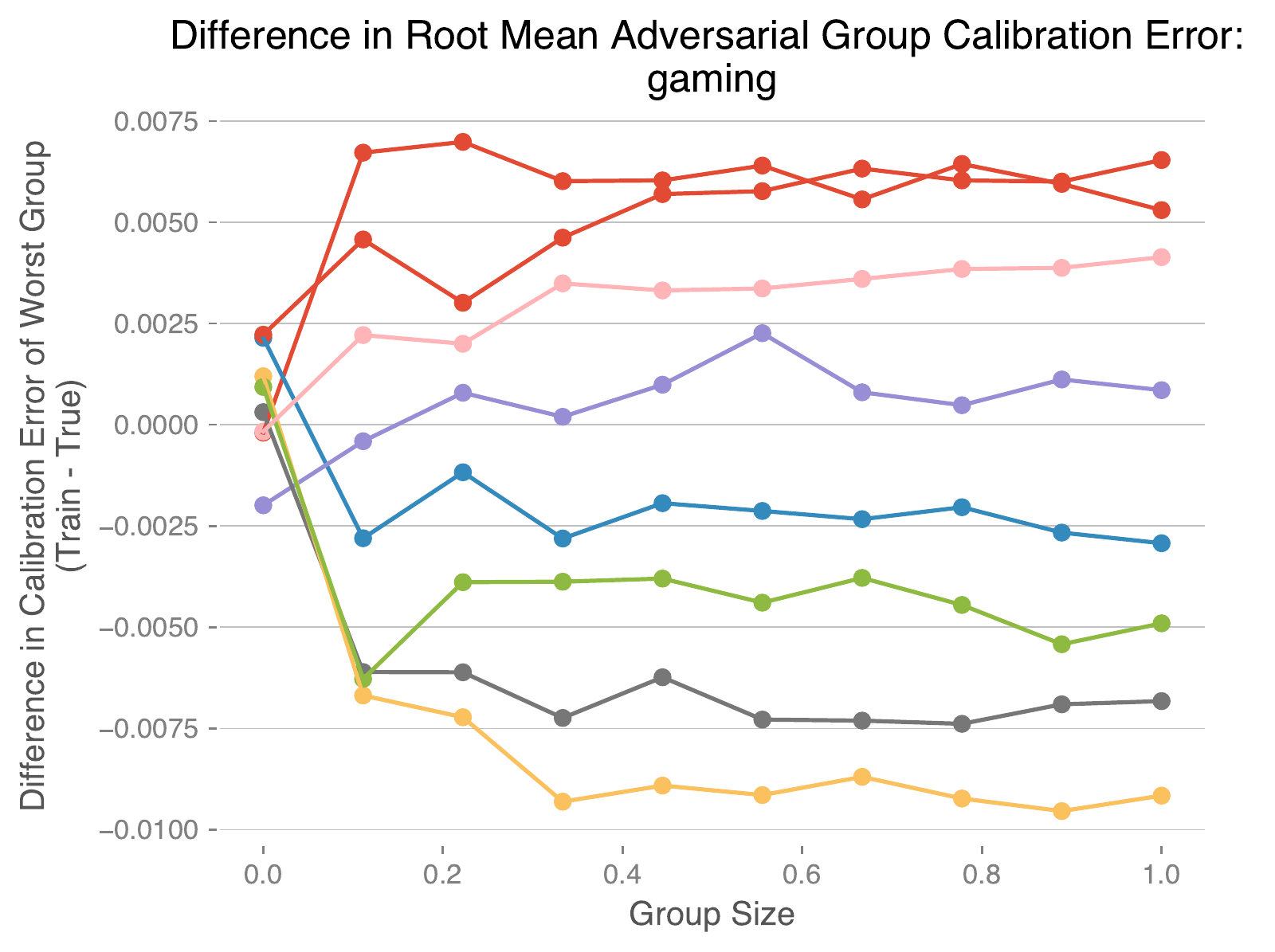}}\\
  \midrule
  \cell{c}{\textds{german}} & \cell{c}{\includegraphics[trim={0 0 0 1.5cm}, clip, width=\linewidth]{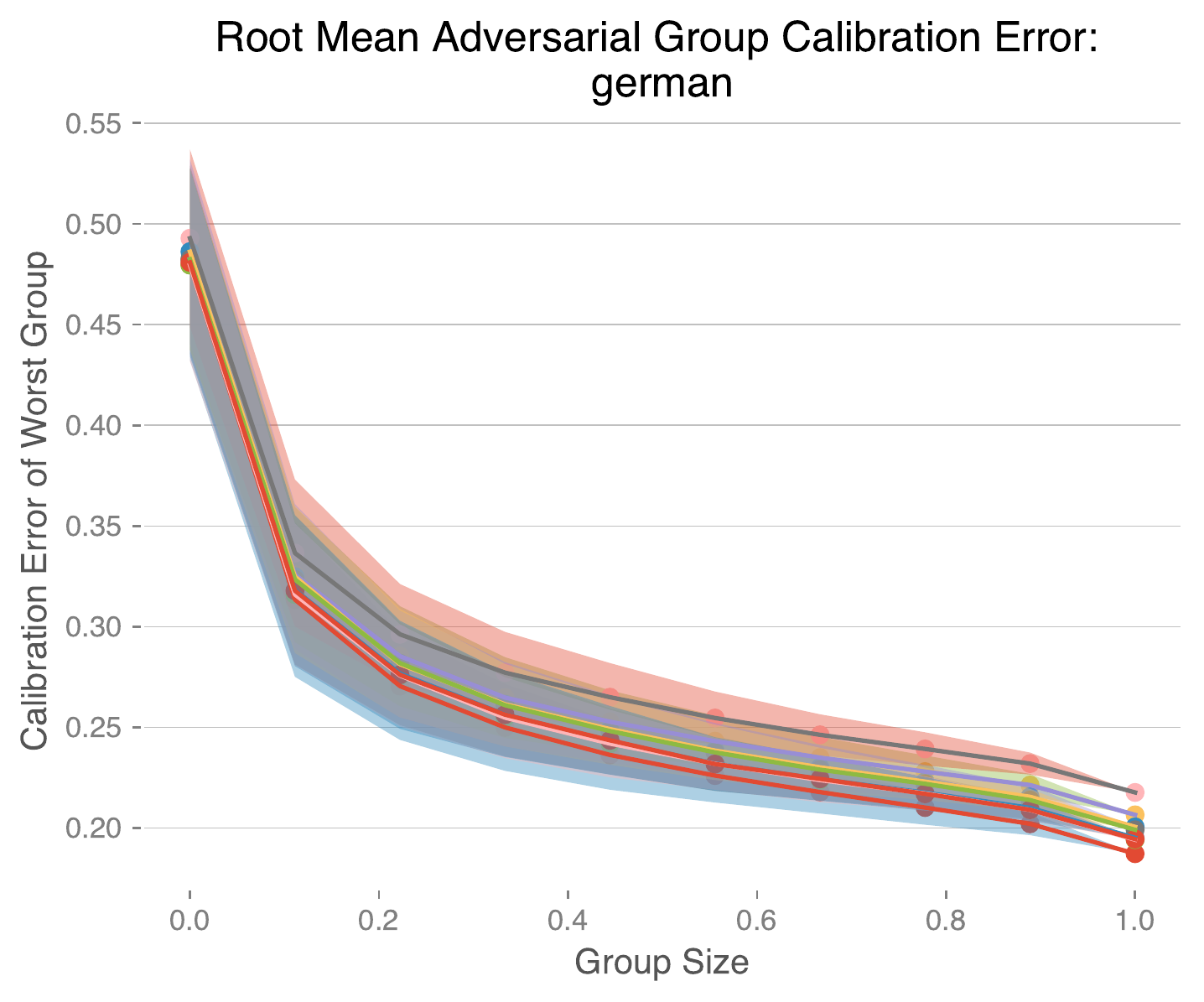}} &
  \cell{c}{\includegraphics[trim={0 0 0 1.5cm}, clip, width=\linewidth]{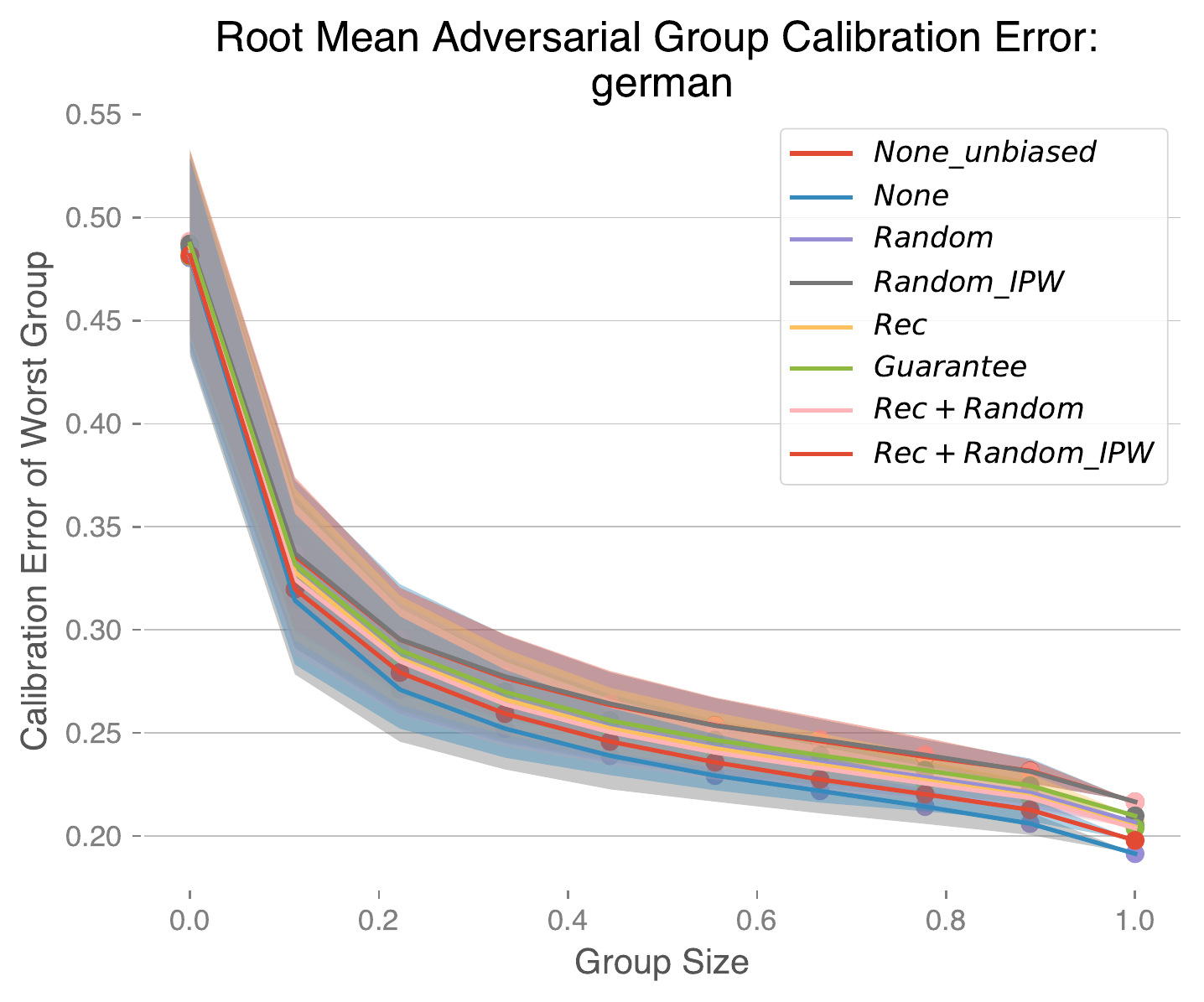}}  &
  \cell{c}{\includegraphics[trim={0 0 0 1.5cm}, clip, width=\linewidth]{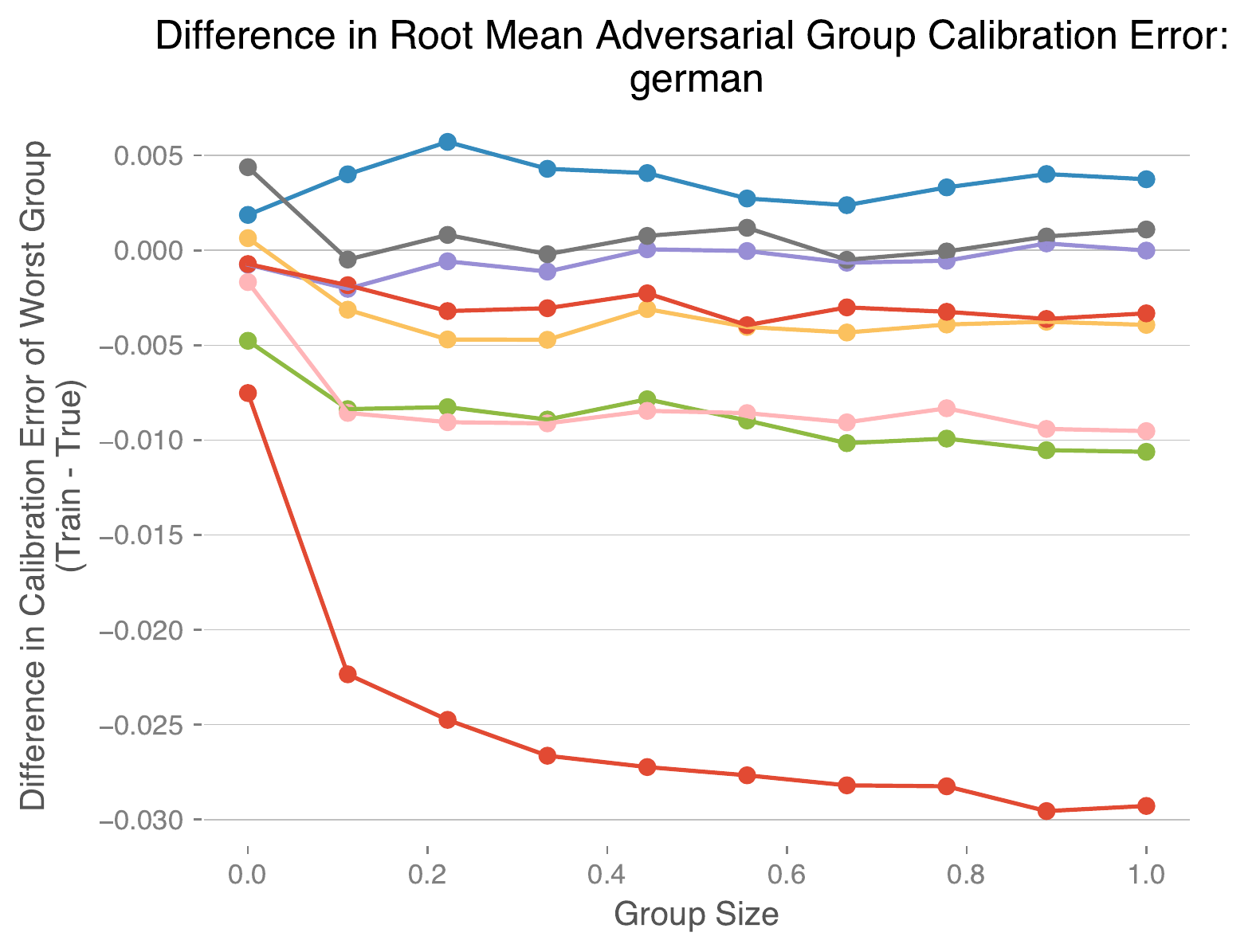}}\\
  \bottomrule
\end{tabular}
}
\caption{Scatterplots of Root Mean Squared Adversarial Group Calibration Error (contd)}
\label{Fig::RMSAG2}
\end{figure}

In order to further demonstrate the difficulty in detection, we report metrics defined and implemented by ~\citep[][]{tran2020methods, chung2021uncertainty} for uncertainty quantification. We demonstrate that censoring is not easily detected by uncertainty metrics such as miscalibration area (\cref{Fig::Heatmap of Miscalibration Area}), sharpness score (\cref{Fig::Heatmap of Sharpness Score}), and root mean squared group calibration error (\cref{Fig::RMSAG1}, \cref{Fig::RMSAG2}). 

\clearpage

\clearpage

\end{document}